\documentclass[runningheads]{llncs}
\usepackage[mobile]{eccv}
\usepackage{eccvabbrv}
\usepackage{graphicx}
\usepackage{booktabs}
\usepackage{algorithm}
\usepackage{algorithmic}
\usepackage{graphicx}
\usepackage{tabularx}
\usepackage{booktabs}
\usepackage{caption}
\usepackage{makecell}
\usepackage{multirow}
\usepackage{multicol}
\usepackage{color}
\usepackage{colortbl}
\usepackage{amssymb}
\usepackage{bm}
\usepackage{gensymb}
\usepackage{textcomp}
\usepackage{stfloats}
\usepackage{url}
\usepackage{verbatim}
\usepackage{graphicx}
\usepackage{graphbox}
\usepackage{pgfplots}
\usepackage{pgfplotstable}  
\usetikzlibrary{matrix}
\usepgfplotslibrary{groupplots}
\pgfplotsset{compat=newest}

\usepackage{adjustbox}

\usepackage{caption}
\usepackage[thinlines]{easytable}

\usepackage{pifont}
\newcommand{\cmark}{\ding{51}}
\newcommand{\xmark}{\ding{55}}
\definecolor{ForestGreen}{RGB}{34,139,34}
\definecolor{tabfirst}{rgb}{1, 0.7, 0.7} 
\definecolor{tabsecond}{rgb}{1, 0.85, 0.7} 
\definecolor{tabthird}{rgb}{1, 1, 0.7} 
\newcommand{\markgood}[1]{{\color{ForestGreen}#1}}

\newcommand{\figref}[1]{Fig. \ref{#1}}
\newcommand{\tabref}[1]{Tab. \ref{#1}}

\newcommand{\secref}[1]{Sec. \ref{#1}}

\newcommand{\paragrapht}[1]{\vspace{2pt}\noindent\textbf{#1}}

\usepackage[accsupp]{axessibility}
\usepackage{hyperref}
\usepackage{orcidlink}

\begin{document}

\title{Talk3D: High-Fidelity Talking Portrait Synthesis via Personalized 3D Generative Prior} 

\titlerunning{Talk3D}

\author{Jaehoon Ko\inst{1} \and
Kyusun Cho\inst{1} \and
Joungbin Lee\inst{1} \and
Heeji Yoon\inst{1}\and
Sangmin Lee\inst{1}\and
Sangjun Ahn\inst{2}\and
Seungryong Kim\inst{1}}

\authorrunning{Ko et al.}

\institute{Korea University, Republic of Korea \and
NCSOFT, Republic of Korea
}

\maketitle
\vspace{-19pt}
\begin{center}
\url{https://ku-cvlab.github.io/Talk3D}
\end{center}

\begin{figure}[h]
\vspace{-5pt}
\newcolumntype{M}[1]{>{\centering\arraybackslash}m{#1}}
\setlength{\tabcolsep}{1pt}
\renewcommand{\arraystretch}{0.75}
\centering
\normalsize
\resizebox{1\linewidth}{!}{
\begin{tabular}{M{0.16\linewidth}M{0.16\linewidth}M{0.16\linewidth}  @{\hskip 0.02\linewidth} M{0.16\linewidth}M{0.16\linewidth}M{0.16\linewidth}}

\includegraphics[width=\linewidth]{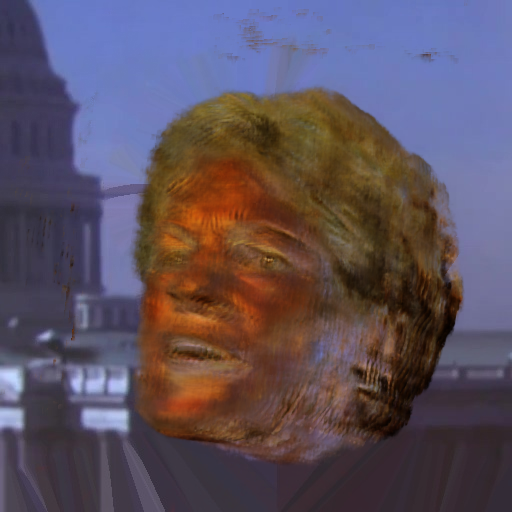}\hfill &
\includegraphics[width=\linewidth]{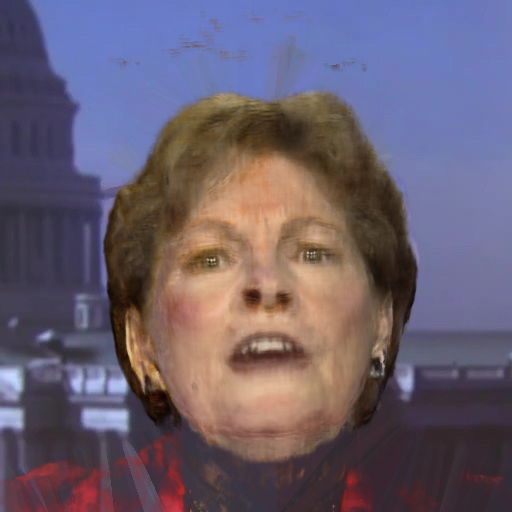}\hfill &
\includegraphics[width=\linewidth]{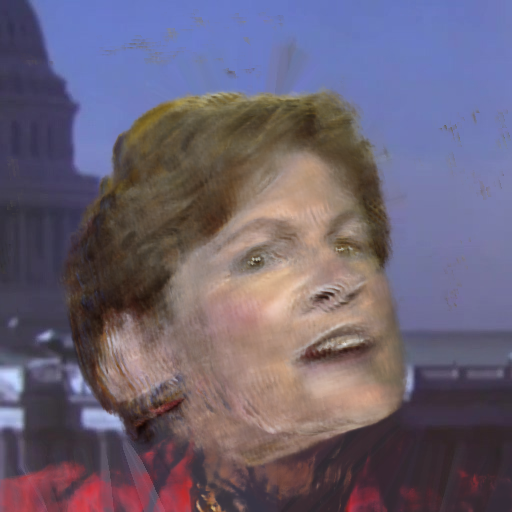}\hfill &

\includegraphics[width=\linewidth]{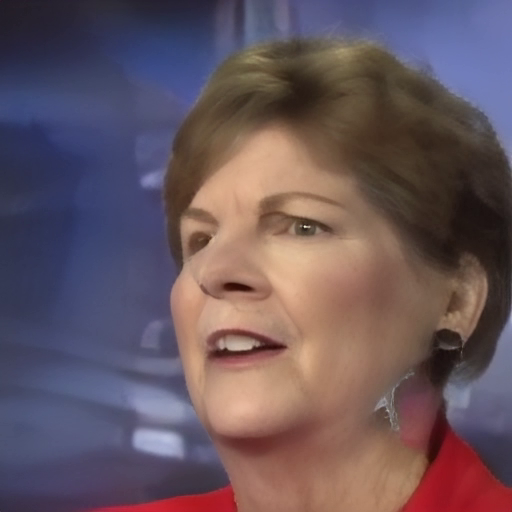}\hfill &
\includegraphics[width=\linewidth]{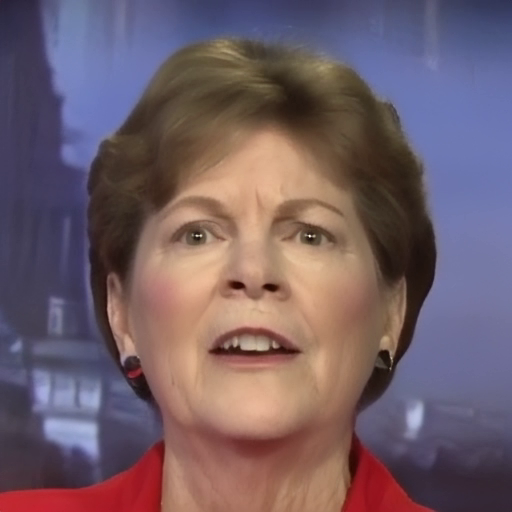}\hfill &
\includegraphics[width=\linewidth]{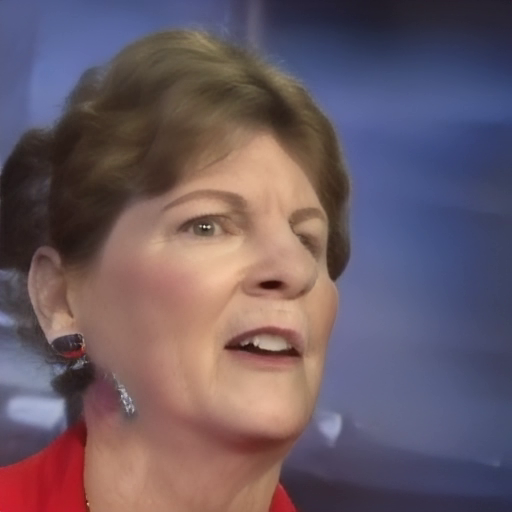}\hfill \\

\includegraphics[width=\linewidth]{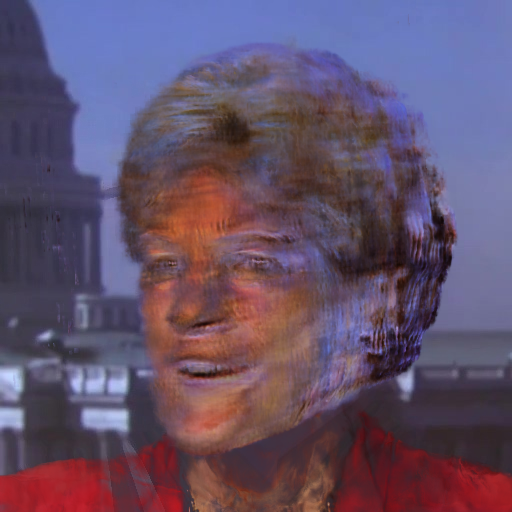}\hfill &
\includegraphics[width=\linewidth]{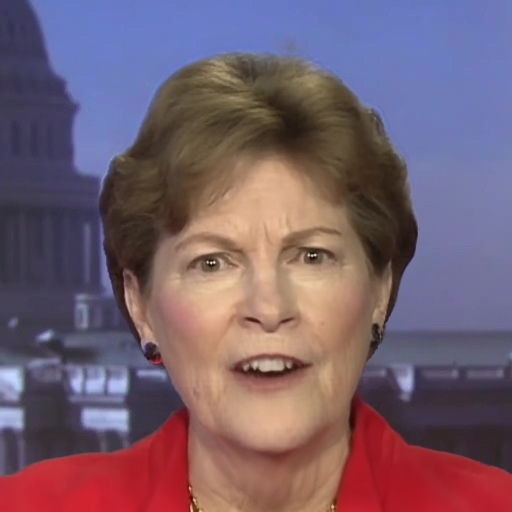}\hfill &
\includegraphics[width=\linewidth]{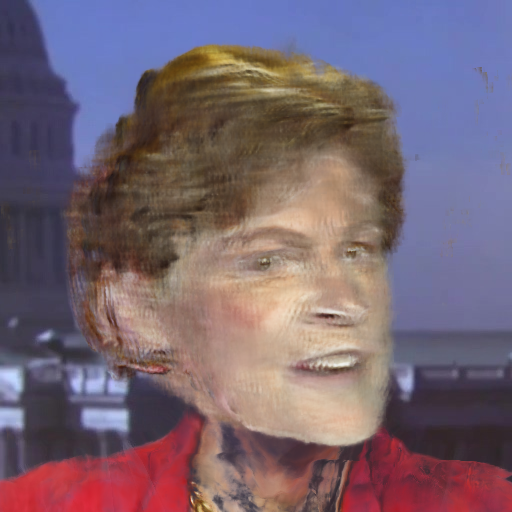}\hfill &

\includegraphics[width=\linewidth]{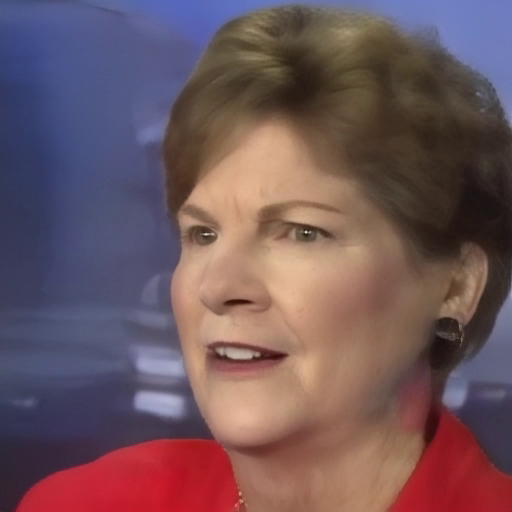}\hfill &
\includegraphics[width=\linewidth]{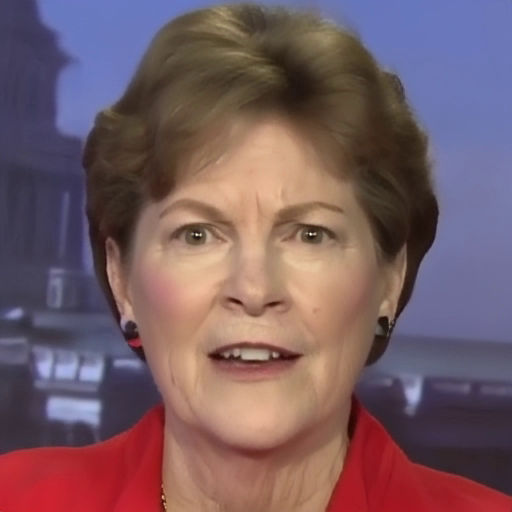}\hfill &
\includegraphics[width=\linewidth]{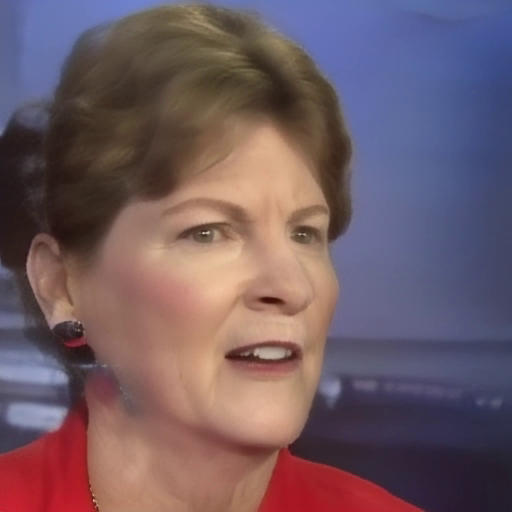}\hfill \\

\includegraphics[width=\linewidth]{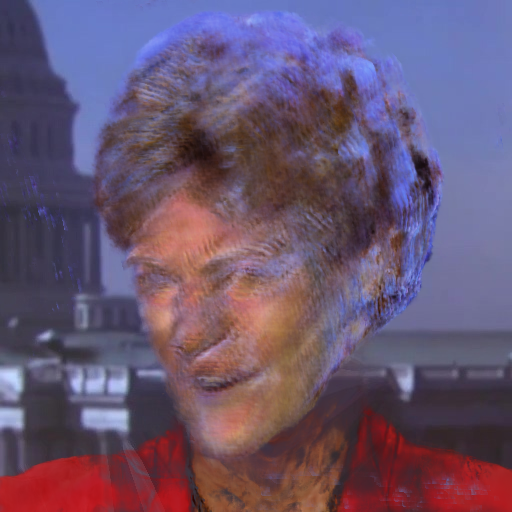}\hfill &
\includegraphics[width=\linewidth]{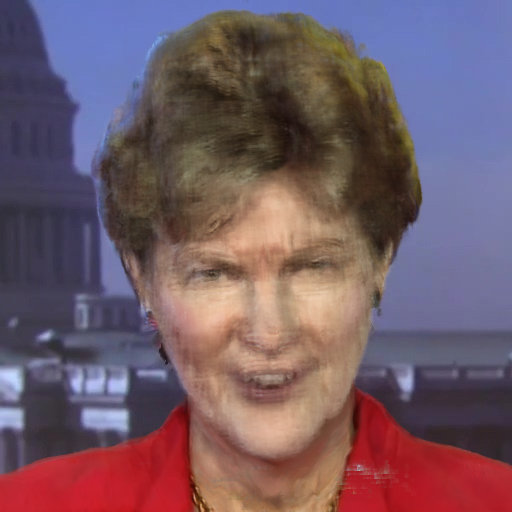}\hfill &
\includegraphics[width=\linewidth]{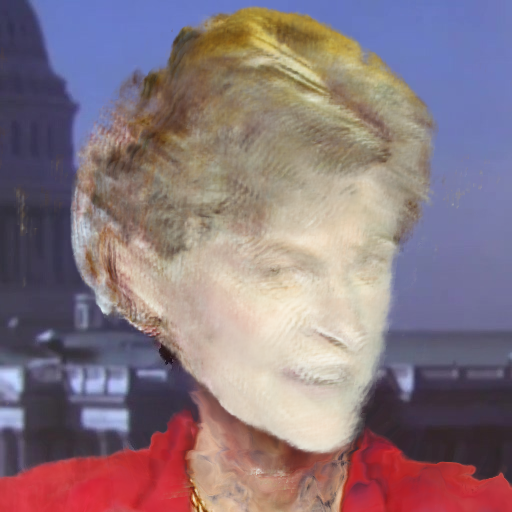}\hfill &

\includegraphics[width=\linewidth]{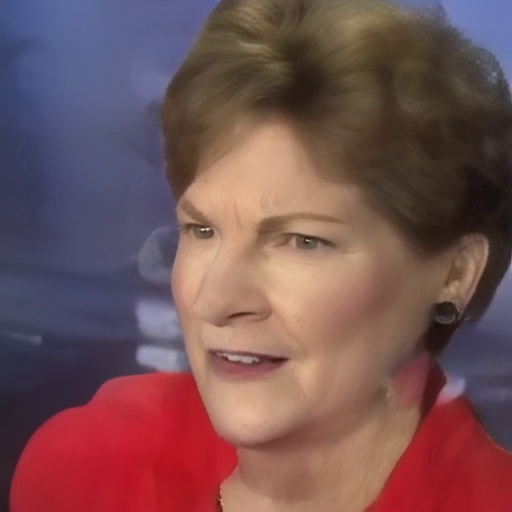}\hfill &
\includegraphics[width=\linewidth]{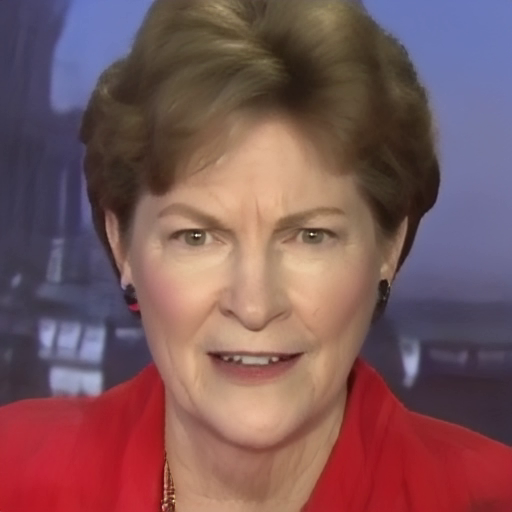}\hfill &
\includegraphics[width=\linewidth]{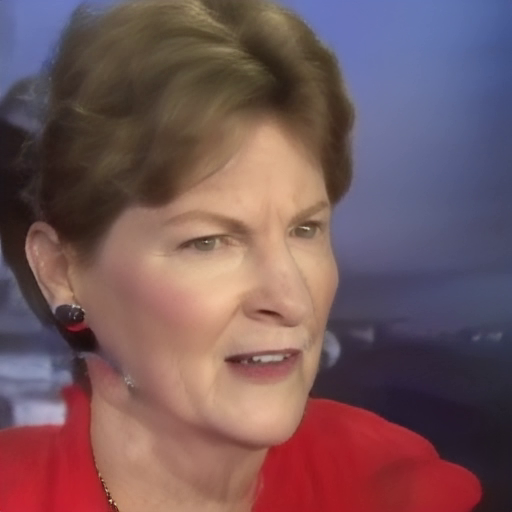}\hfill \\[3pt]

 \multicolumn{3}{c}{ER-NeRF~\cite{li2023ernerf}}  &  \multicolumn{3}{c}{\textbf{Talk3D (Ours)}} \\

\end{tabular}} \vspace{-5pt}

\caption{\textbf{Visualizations of generated talking heads by state-of-the-art NeRF-based method, ER-NeRF~\cite{li2023ernerf}, and \textbf{Talk3D} rendered at extreme novel camera poses.} Our Talk3D shows the robustness in generating high-fidelity realistic geometry of talking heads even at unseen poses during training. }
\vspace{-25pt}
\label{fig:teaser}
\end{figure}

\vspace{-5pt}
\begin{abstract}
Recent methods for audio-driven talking head synthesis often optimize neural radiance fields (NeRF) on a monocular talking portrait video, leveraging its capability to render high-fidelity and 3D-consistent novel-view frames. However, they often struggle to reconstruct complete face geometry due to the absence of comprehensive 3D information in the input monocular videos. In this paper, we introduce a novel audio-driven talking head synthesis framework, called $\mathbf{Talk3D}$, that can faithfully reconstruct its plausible facial geometries by effectively adopting the pre-trained 3D-aware generative prior. 
Given the personalized 3D generative model, we present a novel audio-guided attention U-Net architecture that predicts the dynamic face variations in the NeRF space driven by audio. Furthermore, our model is further modulated by audio-unrelated conditioning tokens which effectively disentangle variations unrelated to audio features. Compared to existing methods, our method excels in generating realistic facial geometries even under extreme head poses. We also conduct extensive experiments showing our approach surpasses state-of-the-art benchmarks in terms of both quantitative and qualitative evaluations. 

\end{abstract}

\section{Introduction}

Audio-driven talking portrait synthesis aims to synthesize a facial video clip featuring a human portrait with lip movements synchronized to the input audio stream~\cite{wiles2018x2face, chen2019hierarchical, jamaludin2019you, prajwal2020wav2lip, zhang2023sadtalker,suwajanakorn2017synthesizing,thies2020nvp, song2022everybody}.
This task poses various challenges including accurately capturing phonemes, generating realistic movements in facial dynamics, and achieving high-fidelity facial image synthesis. 
Early approaches have tackled this task using 2D generative models, focusing on image-based reconstruction of lip motion~\cite{wiles2018x2face, chen2019hierarchical, jamaludin2019you, prajwal2020wav2lip}, but these methods often exhibit limitation on head pose control.  

To address these challenges, recent studies have integrated neural radiance fields (NeRF)~\cite{mildenhall2020nerf} in talking head synthesis to leverage its 
capability in rendering realistic and multi-view consistent images. These approaches either directly condition NeRF on audio features~\cite{guo2021adnerf, liu2022sspnerf, yao2022dfanerf, shen2022dfrf, tang2022radnerf, li2023ernerf} or utilize intermediate representations like facial landmarks~\cite{ye2022geneface, ye2023geneface++}. 
Despite notable advancements, the task of directly constructing dynamic facial NeRF from monocular videos remains challenging. This difficulty stems from the deficiency of diverse head poses and essential 3D information on the monocular videos, resulting in a lack of visual quality when rendered from viewpoints unseen during training.
Even the state-of-the-art methods struggle to generate high-fidelity images from extreme camera poses, with its corresponding depth information showing implausible holes and artifacts (see \figref{fig:teaser}).

On the other hand, 3D-aware generative models~\cite{an2023panohead, chan2022efficient, gu2021stylenerf, sun2023next3d} 
recently gained popularity for the ability 
to generate high-fidelity 3D-aware images through the integration of 3D spatial representation.
In response, several works have explored single-view 3D reconstruction on facial images by fusing these 3D generative models and GAN inversion methods~\cite{ko20233d, yin20233d, bhattarai2024triplanenet, trevithick2023real, yuan2023make, fruhstuck2023vive3d, xu2023innout3d}. By inverting the input image into the latent space of a pretrained 3D-aware GAN, they faithfully synthesize novel-view images from a single input image.

In this paper, we argue that this capability of 3D-aware GANs can be seamlessly extended to 3D talking head generation, offering the advantage of rendering realistic talking portraits from unseen viewpoints. To implement such a model, one of the intuitive strategies would be directly predicting the latent vector within the GAN latent space conditioned by audio. Notably, previous works have employed this strategy, either conducting audio-driven editing using GAN latent space~\cite{bai2023hfagp} or predicting latent through 3DMM parameters~\cite{ma2023otavatar}. However, these approaches encounter challenges due to the high dimensionality of the GAN latent space, which is a complex feature space for the model to learn the elaborate lip movement within the NeRF space. 
Moreover, the objective of audio-driven NeRF editing tasks is to specifically modify localized regions
such as the lip region or chin, but GAN latent space influences the entire scene, posing a disadvantage to model training.

To overcome these, we introduce a novel audio-driven talking head generation framework leveraging a 3D-aware generative prior, dubbed \textbf{Talk3D}, allowing accurate lip movement synchronization with its realistic geometry estimation. Specifically, we integrate two crucial strategies: 1) leveraging 3D-aware GAN prior for realistic geometry and 2) executing direct NeRF space editing beyond the GAN latent space. Specifically, our U-Net-based architecture takes a fixed personalized triplane to yield the triplane offset, namely deltaplane. Modulated by audio features, our model learns to predict a deltaplane that represents precise lip movement within the NeRF space, tailored to the corresponding audio.
Additionally, our U-Net architecture employs an attention-based module and accommodates extra features as conditioning tokens for disentanglement. Notably, this architectural choice enables our model to disentangle the local facial variations within the portrait image such as the torso, background, and eye movements. This disentanglement also improves lip-sync accuracy by ensuring the successful mapping of audio features to lip movements.

To summarize, our proposed talking head synthesis method contains contributions as follows:
\begin{itemize}
    \item We present to adopt 3D generative priors for synthesizing a 3D-aware talking head avatar with realistic geometry.
    \item We propose an audio-driven deltaplane prediction strategy to modify the NeRF space of the employed 3D generative model conditioned by audio.
    \item Our audio-guided attention U-Net architecture successfully disentangles the local variations within frame, such as background, torso, and eye movement.
    \item Our model shows state-of-the-art talking head generation results, proved on extensive experimental results.
\end{itemize}

\section{Related Work}
\vspace{-5pt}
\paragrapht{Audio-driven talking portrait synthesis. } 
Talking portrait synthesis is a challenging task, as the generated lips must be synced to the given audio, while also producing a consistent identity with realistic facial movements. 
Early deep-learning based methods~\cite{prajwal2020wav2lip,  das2020speech, christos2020headgan} employed 2D generative adversarial networks (GANs)~\cite{goodfellow2014generative} to synthesize audio-synchronized lip motions while maintaining realistic facial structure, but lacked control over head poses.
Although subsequent works attempted pose control using facial landmarks and 3D facial models~\cite{thies2020nvp, wang2020mead, lu2021lsp, zhang2023sadtalker}, the reliance on intermediate representations inadvertently led to information loss during the training process~\cite{shen2022learning}.

\begin{table}[t!]
\caption{\textbf{Comparison of our model with existing methods.} }
\vspace{-25pt}
\begin{center}

\resizebox{\textwidth}{!}{
\begin{tabular}{l|cccccccc|c}
\toprule
 & Wav2Lip &   AD-NeRF&   SSP-NeRF & DFA-NeRF &  RAD-NeRF  &    ER-NeRF  & HFA-GP  & OTAvatar      & \textbf{Talk3D} \\
& \cite{prajwal2020wav2lip} & \cite{guo2021adnerf}&    \cite{liu2022sspnerf}  &    \cite{yao2022dfanerf}   &  \cite{tang2022radnerf}    & \cite{li2023ernerf} &   \cite{bai2023hfagp}    & \cite{ma2023otavatar}     &\textbf{(Ours)}

\\ \midrule
Audio-driven?       & \markgood{\cmark}  & \markgood{\cmark} &  \markgood{\cmark}   &  \markgood{\cmark}     & \markgood{\cmark}  & \markgood{\cmark}  & \markgood{\cmark} & \xmark & \markgood{\cmark}\\ 

3D-aware?      &\xmark  & \markgood{\cmark}  & \markgood{\cmark} &  \markgood{\cmark}  &  \markgood{\cmark}   &  \markgood{\cmark}     & \markgood{\cmark}  & \markgood{\cmark} & \markgood{\cmark}\\ 

Generalizability to unseen poses?      & \xmark  & \xmark & \xmark  & \xmark & \xmark & \xmark  & \markgood{\cmark} & \markgood{\cmark}& \markgood{\cmark}\\

Disentangled facial attributes?  & \xmark  & \xmark & \xmark  &  \markgood{\cmark}  &  \markgood{\cmark}  & \markgood{\cmark}  & \xmark & NA   & \markgood{\cmark}\\

Region-aware condition mapping?     & \xmark  & \xmark & \markgood{\cmark} & \xmark  & \xmark   & \markgood{\cmark}  & \xmark & \xmark   & \markgood{\cmark}\\

Volumetric torso? &  \xmark & \markgood{\cmark} & \markgood{\cmark}  & \markgood{\cmark} & \xmark  & \xmark  & \markgood{\cmark}    &  \markgood{\cmark} & \markgood{\cmark} \\ 

Joint volume of head and torso? & \xmark & \xmark  &  \markgood{\cmark}  &  \xmark  &  \xmark  & \xmark &  \markgood{\cmark}    &  \markgood{\cmark}  & \markgood{\cmark}  \\ \bottomrule

\end{tabular}}
\end{center}
 \label{tab:related_work_comparison}
 \vspace{-10pt}

\end{table}

AD-NeRF~\cite{guo2021adnerf} first incorporated neural radiance fields (NeRF)~\cite{mildenhall2020nerf} to address the challenges of 3D head structure in audio-driven talking portrait synthesis.
SSP-NeRF~\cite{liu2022sspnerf} incorporated a semantic sampling strategy to predict the differential impact of audio on facial areas. 
RAD-NeRF~\cite{tang2022radnerf} and ER-NeRF~\cite{li2023ernerf} achieved substantial improvements in visual quality and efficiency leveraging Instant-NGP~\cite{muller2022instant}.
Nevertheless, these NeRF-based methods~\cite{guo2021adnerf, liu2022sspnerf, tang2022radnerf, li2023ernerf, shen2022dfrf, ye2022geneface, ye2023geneface++, yao2022dfanerf} struggle to achieve multi-view consistency and realistic geometries, because they learn facial representations only from a single monocular video.
Detailed comparison of relevant existing works with our Talk3D is summarized in~\tabref{tab:related_work_comparison}. 

\paragrapht{NeRF-based 3D-aware GANs. } 
Some works have extended NeRF for generating images preserving multi-view consistency, by conditioning NeRF on sampled random vector~\cite{chan2021pi} or semantic codes~\cite{schwarz2020graf, niemeyer2021giraffe}. 
Subsequent works enabled high-resolution image synthesis by using a super-resolution module~\cite{gu2021stylenerf, or2022stylesdf, chan2022efficient} or its hybrid 3D representations~\cite{chan2022efficient, schwarz2022voxgraf, devries2021unconstrained} that integrate NeRF representations with the merits of explicit representations.
Notably, EG3D~\cite{chan2022efficient}, which we use as our base representation, has achieved state-of-the-art image generation quality while maintaining 3D consistency through the design of an efficient triplane hybrid 3D representation.

\paragrapht{Facial reconstruction with generative priors. }
Subsequent works took advantage of EG3D's efficient representation and diverse latent space for usage in 3D facial reconstruction. Some works~\cite{ko20233d, lin20223dganinversion, xie2023high, bhattarai2024triplanenet, yuan2023make, yin20233d, jiang2023meta, trevithick2023real} perform GAN inversion on 3D-aware GANs to reconstruct and edit a 3D facial avatar given a single image. Extensive works explored their various approaches such as direct optimization strategies~\cite{ko20233d, yin20233d}, encoder-based inversion~\cite{bhattarai2024triplanenet, trevithick2023real, yuan2023make, jiang2023meta}, or extending method for inverting consecutive video frames~\cite{fruhstuck2023vive3d, xu2023innout3d}. Recent works~\cite{xu2023omniavatar, sun2023next3d, bai2023hfagp, ma2023otavatar} enabled facial animation from a given triplane representation, either by presenting a deformation field to morph the reconstructed 3D model, or by sequentially estimating the latent codes that correspond to the given condition.  

\section{Methodology}

\subsection{Preliminary: NeRF-based 3D-aware GANs}

While conventional neural radiance field (NeRF)~\cite{mildenhall2020nerf, chen2022tensorf, muller2022instant, yu2021plenoxels} aims to be optimized for a single static scene, NeRF-based 3D-aware GANs~\cite{schwarz2020graf, niemeyer2021giraffe, chan2021pi, xue2022giraffehd, chan2022efficient, deng2022gram, schwarz2022voxgraf, xiang2022gramhd} achieved explicitly pose-controllable image generation by conditioning their NeRF space with random-sampled latent code $\mathbf{w}$.
Among these works, EG3D~\cite{chan2022efficient} demonstrates its superior performance using three stages. 
First, EG3D employs a plane generator $\mathcal{G}(\cdot;\theta_{\mathcal{G}})$ parametrized by $\theta_{\mathcal{G}}$ that efficiently synthesizes low resolution feature plane $\mathbf{P}$ such that $\mathbf{P} = \mathcal{G}(\mathbf{w};\theta_{\mathcal{G}})$. This feature plane is reshaped to three orthogonal feature planes, $\{ \mathbf{P}^\mathrm{xy}, \mathbf{P}^\mathrm{yz},\mathbf{P}^\mathrm{zx} \}$. 
EG3D then utilizes an MLP that takes features aggregated from the orthogonal planes and maps it to volume density $\sigma$ and feature $\mathbf{f}$. This feature field is rendered to a low resolution 2D feature map $\mathbf{F}$, 
Finally, the produced feature map $\mathbf{F}$ undergoes processing in a 2D super-resolution module comprised of several convolutional layers to generate the final image $I$. We denote $\mathcal{R}(\cdot;\theta_\mathcal{R})$ as this sequential process involving volume rendering and super-resolution module. Given $\theta_\mathcal{R}$ as the learnable parameters, the final synthesized image can be formulated as: $I=\mathcal{R}(\mathbf{P}, \pi;\theta_\mathcal{R})$.

\begin{figure*}[!t]
    \centering
    \includegraphics[width=1\linewidth]{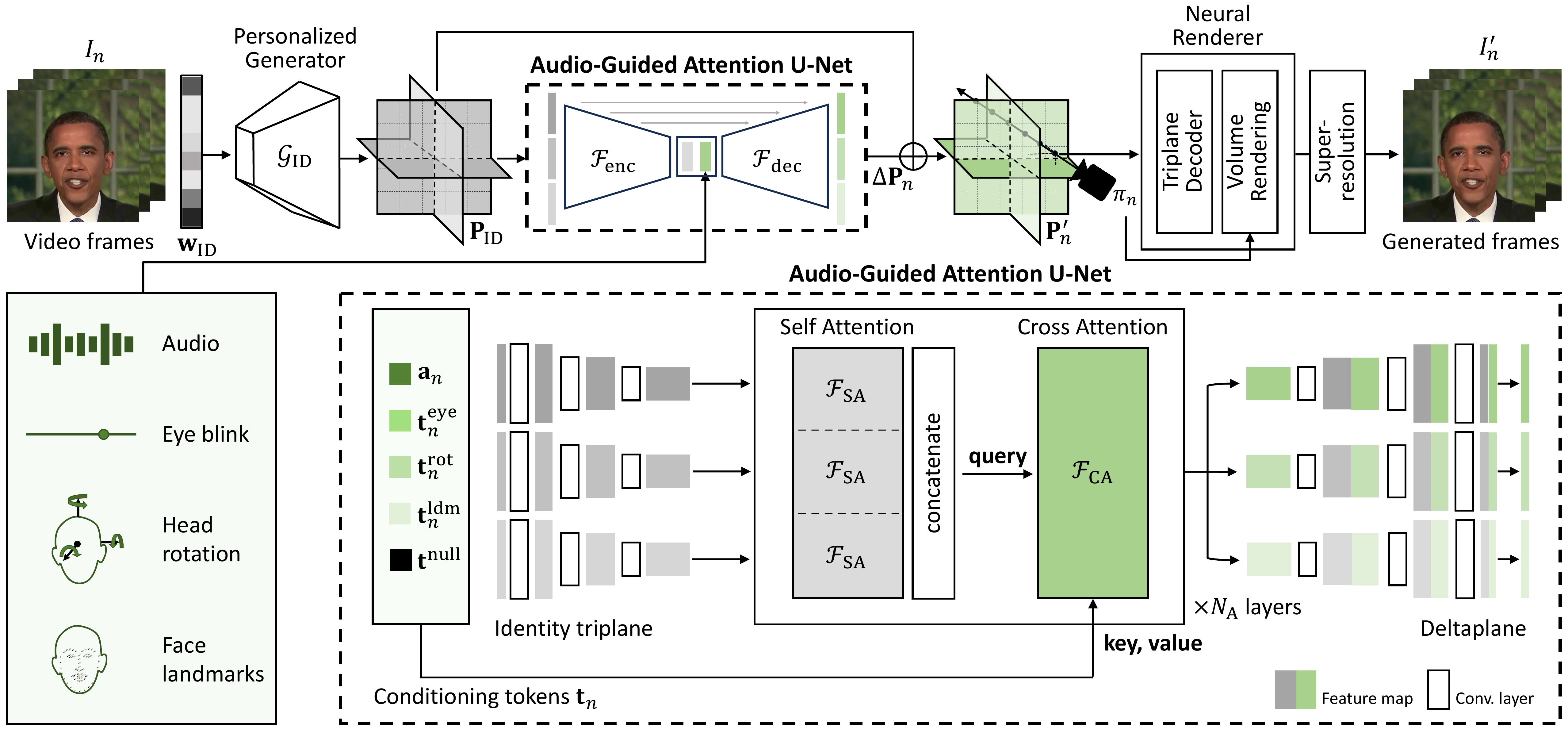}
    \caption{\textbf{Overview of our Talk3D framework.} Our model mainly utilizes a personalized generator. Given identity triplane $\mathbf{P}_\mathrm{ID}$ and $n$-th frame's conditioning tokens $\mathrm{t}_n$, audio-guided attention U-Net predicts the deltaplane $\Delta \mathbf{P}_n$ which represents the dynamic residual scene variation of the corresponding ground-truth image $I_n$. 
    This is further combined with $\mathbf{P}_\mathrm{ID}$ through summation, forming $\mathbf{P}'_n$, and fed to the neural renderer with given camera viewpoint $\pi_n$ to generate final output image $I'_n$.}
    \label{fig:main_figure}
    \vspace{-10pt}
\end{figure*}
\subsection{Problem Formulation and Overview}
In this section, we describe the main components of our method, $\mathbf{Talk3D}$, which enables pose-controllable audio-driven high-fidelity talking portrait synthesis. 
Given $N$ number of video frames for a specific identity, $\mathcal{V}=\{I_n\}$, our model takes $n$-th frame image $I_n$ with corresponding audio feature $\mathbf{a}_n$, and camera parameter $\pi_n$. We then formulate the audio-driven rendering process as:
\begin{equation}\label{eq:audio-driven_basic}
\begin{split}
    &\mathbf{P} = \mathcal{G}(\mathbf{w};\theta_\mathcal{G}),\\
    &I'_n = \mathcal{R}(\mathbf{P}, \pi_n,\mathbf{a}_n;\theta_\mathcal{R}).
\end{split}
\end{equation} 
To attain the rendered portrait image $I'_n$ that best replicates the lip movement of the frame $I_n$,
our model aims to find the optimal EG3D~\cite{chan2022efficient} parameters denoted as $\{\theta^*_\mathcal{G}$, $\theta^*_\mathcal{R}\}$, and the optimal triplane $\mathbf{P}^*$ which encapsulates the appropriate scene encodings. At inference, given new audio $\mathbf{a}^\mathrm{novel}_n$, we reformulate \eqref{eq:audio-driven_basic} as:
\begin{equation}
\begin{split}
    &\mathbf{P}_\mathrm{ID} = \mathcal{G}(\mathbf{w}_\mathrm{ID};\theta^*_\mathcal{G}),\\
    &I^\mathrm{novel}_n = \mathcal{R}(\mathbf{P}_\mathrm{ID}, \pi_n,\mathbf{a}^\mathrm{novel}_n;\theta^*_\mathcal{R}),
\end{split}
\end{equation} 
where $\mathbf{w}_\mathrm{ID}$ denotes an identity latent code that corresponds to a specific person's facial identity. Then, such a personalized generator generates $\mathbf{P}_\mathrm{ID}$, namely identity triplane.

To formulate this, we first train a personalized generator that gives $\mathbf{w}_\mathrm{ID}$ and
$\{\theta^*_{\mathcal{G}}, \theta^*_{\mathcal{R}}\}$. In the renderer $\mathcal{R}(\cdot)$, to condition $\mathbf{a}^\mathrm{novel}_n$, we propose a deltaplane generator that generates a new plane $\Delta \mathbf{P}^\mathrm{novel}_n$ from $\mathbf{a}^\mathrm{novel}_n$ to manipulate the identity plane $\mathbf{P}_\mathrm{ID}$ such that $\mathbf{P}' = \mathbf{P}_\mathrm{ID} + \Delta \mathbf{P}^\mathrm{novel}_n$. Then our final renderer is defined as follows:
\begin{equation}
    I^\mathrm{novel}_n = \mathcal{R}(\mathbf{P}', \pi_n;\theta^*_\mathcal{R}) = \mathcal{R}(\mathbf{P}_\mathrm{ID} + \Delta \mathbf{P}^\mathrm{novel}_n, \pi_n;\theta^*_\mathcal{R}).
\end{equation}

In the following, 
we first explain our generator fine-tuning strategy (\secref{ID-plane}) 
and our audio-conditioned deltaplane prediction method (\secref{motion-plane}). 
Finally, we describe the loss functions employed in our proposed framework (\secref{loss_function}). An overview of our proposed framework is depicted in~\figref{fig:main_figure}.

\subsection{Personalized Generator for Identity Triplane Generation} \label{ID-plane}
3D-aware GANs are usually trained on an extensive dataset of facial images such as FFHQ~\cite{karras2019style}, allowing for the generation of a wide range of personal identities. This nature of 3D-aware GANs may not be an optimal choice for our specific problem setting, which involves capturing the speech of a single person recorded by a monocular video. In this work, we adopt VIVE3D~\cite{fruhstuck2023vive3d}, a fine-tuning strategy for 3D-aware GAN to produce single-identity images. This personalizing step aims to enhance the model's editability and visual fidelity.

The personalization strategy involves a variant of pivotal tuning~\cite{roich2021pivotal}, which inverts a few selected frames to find the optimal latent vector in $\mathbf{w}$-space, and then jointly fine-tunes the generator $\mathcal{G}$ and $\mathcal{R}$. They first select $M$ number of frames $I_m$ and simultaneously optimize the latent vectors $\mathbf{w}_\mathrm{ID} + \mathbf{o}_m$ of each frame, where
$\mathbf{o}_m$ is additional offset vectors that aim to capture the local variants such as facial expression or lip movement. Consequently, they jointly fine-tune the weight of the generator on $I_m$, while keeping the optimal latent vectors $\mathbf{w}_\mathrm{ID} + \mathbf{o}_m$ fixed. Finally, for the $N$ number of the total target frames $I_n$, they conduct frame-by-frame video inversion on the fine-tuned generator $\mathcal{G}_\mathrm{ID}$ to predict a stack of offsets $\mathbf{o}_n$ and camera parameters $\pi_n$.

\subsection{Audio-Guided Attention U-Net for Deltaplane Generation}\label{motion-plane}
Throughout the aforementioned inversion process, we achieve the personalized generator $\mathcal{G}_\mathrm{ID}$, the global identity $\mathbf{w}_\mathrm{ID}$, and the camera parameters for each frame $\pi_n$. We further derive the identity triplane $\mathbf{P}_\mathrm{ID} = \mathcal{G}_\mathrm{ID}(\mathbf{w}_\mathrm{ID};\theta^*_\mathcal{G})$ to integrate into our training framework. 
Our ultimate goal is to modulate the generator with an audio feature vector, thus conditioning the NeRF space and enabling image manipulation. While the most straightforward approach~\cite{bai2023hfagp} involves predicting the latent vector within the generator's learned manifold, we experimentally found that it may not necessarily be the most optimal choice. 
Alternatively, we introduce a training method that focuses on the direct prediction of a triplane grid rather than the $\mathbf{w}$-space latent vector. 
In the following, we will explain how to manipulate the triplane with given condition $\mathbf{a}_n$.

\paragrapht{Audio-guided attention U-Net architecture.}
As depicted in \figref{fig:main_figure}, U-Net based architecture $\mathcal{F}$ is employed, where identity triplane serves as an input, yielding offset triplane grid $\Delta \mathbf{P}_n$ such that: $\Delta \mathbf{P}_n = \mathcal{F}(\mathbf{P}_\mathrm{ID}, \mathbf{a}_n;\theta)$, where $\mathbf{a}$ denotes given audio feature. This offset grid $\Delta \mathbf{P}_n$, which we call deltaplane is further combined with $\mathbf{P}_\mathrm{ID}$.
This training strategy offers several distinct advantages compared to GAN latent prediction. 
First of all, this GAN latent prediction approach struggles to represent the disentangled lip movement due to the high-dimensionality of GAN latent space. This obstacle leads to undesired prediction such as flickering in background and torso area.
Furthermore, the triplane grid directly represents the 3D grid structure of the NeRF space which guides the model to understand and manipulate the spatial relationships within the scene.
Lastly, the triplane grid is basically a 2D feature map returned from convolutional networks, which offers the convenience of leveraging existing 2D-based network architectures.

\paragrapht{Attention design.}
In an ideal setting, the deltaplane should seamlessly amalgamate temporal motion signals with the identity triplane, ensuring that the signals are appropriately synchronized with the relevant facial segments. This becomes imperative for audio, as their impact on the entirety of the facial movements is not uniform. We incorporate cross-attention at the deepest hidden layer of U-Net architecture to effectively capture localized facial dynamics during the generation of the deltaplane.
Specifically, the U-Net encoder $\mathcal{F}_\mathrm{enc}$ encodes $\mathbf{P}$ into a low-resolution feature map as $\mathbf{E}=\mathcal{F}_\mathrm{enc}(\mathbf{P})$.
Consequently, this feature map is passed through ${N}_{\mathrm{A}}$ number of attention layers, each comprised of self-attention ($\mathrm{SA}$) and cross-attention ($\mathrm{CA}$) layer, which we denote as: $\mathcal{F}_\mathrm{SA}$ and $\mathcal{F}_\mathrm{CA}$. Specifically, $\mathrm{SA}$ and $\mathrm{CA}$ can be defined as:
\begin{equation}
\begin{split}
    &\mathbf{e} = \mathcal{F}_{\mathrm{SA}}(\mathrm{flatten}(\mathbf{E} + \mathbf{E}^\mathrm{pos})).\\ 
    &\mathbf{E}^\mathrm{out}_n = \mathcal{F}_{\mathrm{CA}}(\mathbf{e}, \mathbf{a}_n),
\label{eq:CA}
\end{split}
\end{equation}
where $\mathbf{E}^\mathrm{pos}$ denotes 3D positional encoding. 
Finally, the U-Net decoder generates the deltaplane by $\Delta \mathbf{P}_{n}=\mathcal{F}_\mathrm{dec}(\mathbf{E}^\mathrm{out}_{n})$.

\paragrapht{Split-convolution.}
The original EG3D~\cite{chan2022efficient} employs a single convolution network to generate the triplane, where each plane, $\mathbf{P}^\mathrm{xy}$, $\mathbf{P}^\mathrm{yz}$, and $\mathbf{P}^\mathrm{zx}$, is channel-wise concatenated. However, we observed a performance decline when utilizing the $\mathcal{F}_\mathrm{enc}$ structure as a singular model. This degradation stems from the orthogonality of each plane within the NeRF space, and the channel-wise concatenation hinders the 3D-awareness of the triplane. To address this issue, our architecture processes each plane independently to maintain the individual plane's characteristics. Nevertheless, since each plane's features equally contribute to the query sampled points by concatenation, the aforementioned split convolution structure may impede the learning of the correlation between each plane. Therefore, we incorporate the roll-out method~\cite{wang2023rodin, chen2023mimic3d} to appropriately blend features from each plane.
\label{paragraph:split_conv}

\paragrapht{Augmenting condition.}
Following previous NeRF-based works~\cite{guo2021adnerf, tang2022radnerf, li2023ernerf, liu2022sspnerf},
our dataset settings closely align with theirs, except for variations in image crop regions. 
This disparity is caused by the image cropping process in the utilization of the EG3D~\cite{chan2022efficient}.
Consequently, a specific challenge arises, wherein alterations to the crop area may give the appearance of unnecessary movement between the background and the torso's position, which interferes with the learning of audio features.
To mitigate this challenge, we encode additional signals with causal relationships to the torso and background movements. Features capturing independent actions, such as background motion (inferred from facial landmarks coordinate in the original video), and torso dynamics (correlated with head rotation), are tokenized as $\mathbf{t}^\mathrm{rot}$ and $\mathbf{t}^\mathrm{ldm}$ and then incorporated through cross-attention layers. The intuition here lies in the effectiveness of our model's cross-attention layer, allowing diverse tokens to be efficiently learned for local editing within the triplane. As utilized in prior work~\cite{li2023ernerf}, we employ the AU45~\cite{Ekman1978FacialAC} features to describe eye movements, which also be tokenized into $\mathbf{t}^\mathrm{eye}$. Additionally, a single null-token $\mathbf{t}^\mathrm{null}$ incorporated uniformly across all frames, serving the purpose of global features across video frames. Again,~\eqref{eq:CA} can be reformulated as:
\begin{equation}
    \mathbf{E}^\mathrm{out}_{n} = \mathcal{F}_{\mathrm{CA}}(\mathbf{e}, \mathbf{t}_n), ~~ \mathbf{t}_n = \{ \mathbf{a}_{n}, \mathbf{t}^\mathrm{eye}_n, \mathbf{t}^\mathrm{rot}_n, \mathbf{t}^\mathrm{ldm}_n, 
    \mathbf{t}^\mathrm{null}\},
\label{eq:CA_reform}
\end{equation}
where $\mathbf{t}_\mathrm{n}$ denotes the concatenation of all tokens.

\subsection{Loss Functions} 
To train the model, we mainly adopt $L$1 loss $\mathcal{L}_{\mathrm{L1}}$ and LPIPS loss $\mathcal{L}_{\mathrm{lpips}}$~\cite{zhang2018lpips} to reconstruct given input frame $I$. Let $\mathcal{L}_\mathrm{rec}$ denotes the combination of the above reconstruction loss as: $\mathcal{L}_\mathrm{rec}=\mathcal{L}_{\mathrm{L1}} + \lambda_\mathrm{lpips} \mathcal{L}_{\mathrm{lpips}}$. Similarly, we obtain a semantic segmentation of facial region with BiSeNet~\cite{yu2018bisenet, yu2021bisenet}, to enhance the reconstruction loss on the local image area. We give additional reconstruction loss on lip segment $S_\mathrm{lip}(I)$. Moreover, we adopt ID similarity loss $\mathcal{L}_{\mathrm{id}}$ and syncnet loss $\mathcal{L}_{\mathrm{sync}}$~\cite{chung2017syncnet1, prajwal2020wav2lip} to further optimize the generation results. In summary, our total loss function $\mathcal{L}$ can be formulated as:
\begin{equation}
     \mathcal{L} = \mathcal{L}_\mathrm{rec}(I, I') + \lambda_\mathrm{lip} \mathcal{L}_\mathrm{lip} ({S}_\mathrm{lip}(I), {S}_\mathrm{lip}(I')) 
     + \lambda_\mathrm{id} \mathcal{L}_\mathrm{id}(I, I') + \lambda_\mathrm{sync} \mathcal{L}_{\mathrm{sync}}(I, I').
\end{equation}

We additionally take a few epochs to update the super-resolution module.
This additional fine-tuning step aims to boost rendering quality to generate sharper image results. During the fine-tuning stage, only the reconstruction loss $\mathcal{L}_\mathrm{rec}$ is applied, resulting in the following loss function: $\mathcal{L}_\mathrm{tune} = \mathcal{L}_\mathrm{rec}(I, I')$.
\label{loss_function}
\begin{table*}[t]

\caption{\textbf{Quantitative comparison under the \emph{novel view synthesis} setting.} We measure the image fidelity and lip synchronization accuracy of the generated talking portraits using different NeRF-based methods rendered at different head poses. The head poses are selected with $15^{\circ}$ yaw intervals and $10^{\circ}$ pitch intervals. The top, second-best, and third-best results are shown in \textcolor{red}{red}, \textcolor{orange}{orange}, and  \textcolor{yellow}{yellow}, respectively.}
\vspace{-15pt}
\begin{center}
\setlength{\tabcolsep}{0pt}
\definecolor{tabfirst}{rgb}{1, 0.7, 0.7} 
\definecolor{tabsecond}{rgb}{1, 0.85, 0.7} 
\definecolor{tabthird}{rgb}{1, 1, 0.7} 
\scriptsize
\renewcommand{\arraystretch}{1.3}
\newcolumntype{L}[1]{>{\raggedright\arraybackslash}p{#1}}
\newcolumntype{M}[1]{>{\centering\arraybackslash}m{#1}}
\scalebox{0.8}{
\begin{tabular}{ L{0.17\linewidth}  M{0.08\linewidth} M{0.08\linewidth} M{0.08\linewidth} @{\hskip 0.015\linewidth}  M{0.08\linewidth} M{0.08\linewidth} M{0.08\linewidth} @{\hskip 0.015\linewidth}  M{0.08\linewidth} M{0.08\linewidth} M{0.08\linewidth} @{\hskip 0.015\linewidth} M{0.08\linewidth} M{0.08\linewidth} M{0.08\linewidth} @{\hskip 0.015\linewidth}  M{0.08\linewidth} M{0.08\linewidth} M{0.08\linewidth} @{\hskip 0.015\linewidth} M{0.08\linewidth} M{0.08\linewidth} M{0.08\linewidth}} 
\toprule
\multirow{2}{*}{\makecell[c]{\vspace{-6pt} \\ Head angle \\ (yaw, pitch)}}
   & \multicolumn{3}{c}{ $(-30^{\circ}, -20^{\circ})$}    & \multicolumn{3}{c}{ $(-15^{\circ}, -10^{\circ})$}  & \multicolumn{3}{c}{ $(+15^{\circ}, +10^{\circ})$}  & \multicolumn{3}{c}{ $(+30^{\circ}, +20^{\circ})$}   \\ \cmidrule(lr){2-4} \cmidrule(lr){5-7} \cmidrule(lr){8-10} \cmidrule(lr){11-13}
 & Sync$\uparrow$ & FID$\downarrow$ & IDSIM$\uparrow$ & Sync$\uparrow$ & FID$\downarrow$ & IDSIM$\uparrow$ & Sync$\uparrow$ & FID$\downarrow$ & IDSIM$\uparrow$  & Sync$\uparrow$ & FID$\downarrow$ & IDSIM$\uparrow$   \\ \midrule  
AD-NeRF~\cite{guo2021adnerf}               & $2.236$   & $212.845$ & $0.068$   & $3.474$   & $175.978$ & $0.280$     & $3.821$   & $152.018$ & $0.481$     & $2.523$   & $193.343$ & \cellcolor{tabthird} {$0.034$} \\
RAD-NeRF~\cite{tang2022radnerf}        & \cellcolor{tabsecond} {$4.938$}   & \cellcolor{tabsecond} {$167.834$} & \cellcolor{tabthird} {$0.186$}   & \cellcolor{tabthird} {$5.543$}   & \cellcolor{tabthird} {$123.924$} & \cellcolor{tabthird} {$0.378$}     & \cellcolor{tabsecond} {$6.831$}   & \cellcolor{tabthird} {$94.674$} & \cellcolor{tabsecond} {$0.607$}     & \cellcolor{tabsecond} {$5.447$}   & \cellcolor{tabthird} {$185.718$} & \cellcolor{tabsecond} {$0.283$} \\
ER-NeRF~\cite{li2023ernerf}            & \cellcolor{tabthird} {$4.774$} & \cellcolor{tabthird} {$198.291$}  &  \cellcolor{tabsecond} {$0.226$}        & \cellcolor{tabsecond} {$7.335$}   & \cellcolor{tabsecond} {$87.594$} & \cellcolor{tabsecond} {$0.575$}         & \cellcolor{tabthird} {$6.652$}   & \cellcolor{tabsecond} {$80.562$} & \cellcolor{tabthird} {$0.503$}         & \cellcolor{tabthird} {$2.702$}  & \cellcolor{tabsecond} {$141.625$} & $0.022$\\  \arrayrulecolor{black!50}\specialrule{0.15ex}{0.1ex}{0.1ex}
\textbf{Talk3D (Ours) }      & \cellcolor{tabfirst} $7.201$ & \cellcolor{tabfirst}$81.113$  & \cellcolor{tabfirst}$0.611$     & \cellcolor{tabfirst}$7.932$   & \cellcolor{tabfirst}$37.774$ &\cellcolor{tabfirst} $0.766$      & \cellcolor{tabfirst}$8.144$  & \cellcolor{tabfirst}$39.971$ & \cellcolor{tabfirst}$0.797$      & \cellcolor{tabfirst}$7.766$  & \cellcolor{tabfirst}$68.680$ & \cellcolor{tabfirst}$0.643$\\
\arrayrulecolor{black!100}\bottomrule
\end{tabular}
}
\end{center}

\label{tab:ood_pose}
\vspace{-15pt}

\end{table*}

\section{Experiments}

\definecolor{tabfirst}{rgb}{1, 0.7, 0.7} 
\definecolor{tabsecond}{rgb}{1, 0.85, 0.7} 
\definecolor{tabthird}{rgb}{1, 1, 0.7} 

\begin{table*}[t]
\vspace{5pt}
\caption{\textbf{The quantitative results of the \emph{self-driven} setting}.  \textcolor{red}{Red}, \textcolor{orange}{orange}, and \textcolor{yellow}{yellow} highlights indicate the 1st, 2nd, and 3rd-best performing technique
for each metric. 
Note that PSNR, SSIM, and LPIPS are not valid for Wav2Lip~\cite{prajwal2020wav2lip}, as it can see the ground-truth during the self-driven evaluation. }
\vspace{-5pt}
\centering
\scriptsize
\renewcommand{\arraystretch}{1.3}
\setlength{\tabcolsep}{0pt}
\newcolumntype{M}[1]{>{\centering\arraybackslash}m{#1}}
\scalebox{0.8}{
\begin{tabular}{ m{0.19\linewidth}  M{0.1\linewidth} M{0.1\linewidth} M{0.1\linewidth} M{0.1\linewidth} M{0.1\linewidth} M{0.1\linewidth} M{0.1\linewidth} M{0.1\linewidth} } 
\toprule
Methods & PSNR $\uparrow$ & SSIM $\uparrow$  & LPIPS $\downarrow$ & FID $\downarrow$ & LMD $\downarrow$ & AUE $\downarrow$ & Sync $\uparrow$ & IDSIM $\uparrow$ \\
\midrule
Ground Truth  & N/A            &  N/A               & $0$              & $0$              & $0$              & $0$         & $9.077$   & $1$            \\ \midrule
Wav2Lip~\cite{prajwal2020wav2lip}      & 28.678  & 0.862            & 0.053          & $33.074$        & $4.658$  & $3.040$ & \cellcolor{tabfirst}{$10.096$}   & $0.893$  \\
PC-AVS~\cite{zhou2021pcavs}       & $20.729$          & $0.638$        &  $0.112$       & $42.646$   & \cellcolor{tabthird} {$3.419$}      & $2.497$        & \cellcolor{tabsecond}{$8.945$}          & $0.520$     \\ \arrayrulecolor{black!50}\specialrule{0.1ex}{0.2ex}{0.2ex}
AD-NeRF~\cite{guo2021adnerf}      & $27.611$          & $0.877$        & $0.049$          & $20.243$         & $5.692$    & $2.331$     & $5.692$       & $0.904$     \\
RAD-NeRF~\cite{tang2022radnerf}      & \cellcolor{tabthird}{$28.797$}    & \cellcolor{tabthird} {$0.886$}          & \cellcolor{tabthird} {$0.038$}          & \cellcolor{tabthird} {$14.218$}         & $3.467$       & \cellcolor{tabthird}{$2.163$}         & $6.316$  & \cellcolor{tabthird}{$0.921$}   \\     
ER-NeRF~\cite{li2023ernerf}  & \cellcolor{tabsecond}{$29.284$}          & \cellcolor{tabsecond}{$0.891$} & \cellcolor{tabsecond}$0.032$ & \cellcolor{tabsecond} {$11.860$} & \cellcolor{tabsecond}{$3.417$}    & \cellcolor{tabsecond}{$2.025$}          & $6.724$  & \cellcolor{tabsecond}{$0.940$}   \\    
\arrayrulecolor{black!50}\specialrule{0.1ex}{0.2ex}{0.2ex}
\textbf{Talk3D (Ours)} & \cellcolor{tabfirst}{$30.185$}       & \cellcolor{tabfirst}{$0.895$} & \cellcolor{tabfirst}{$0.027$} & \cellcolor{tabfirst}{$8.626$} & \cellcolor{tabfirst}{$2.932$}    & \cellcolor{tabfirst}{$1.920$}          & \cellcolor{tabthird}{$7.383$}  & \cellcolor{tabfirst}{$0.944$}         \\ \arrayrulecolor{black!100} \bottomrule 
\end{tabular}
}
\vspace{5pt}
\label{tab:selfdriven}

\end{table*}

\begin{table}[t]
\caption{\textbf{Quantitative comparison under the \emph{cross-driven} setting.} 
We extract two audio clips from the demo of SynObama~\cite{suwajanakorn2017synthesizing} to drive each method and compare the audio-lips synchronization and lips movement consistency.
}
\scriptsize
\renewcommand{\arraystretch}{1.2} 
\vspace{-13pt}
\begin{center}
\setlength{\tabcolsep}{5pt}
\scalebox{.8}{
\begin{tabular}{l c c c c c c} 
\toprule

     & \multicolumn{3}{c}{Testset A}  & \multicolumn{3}{c}{Testset B}  \\ \cmidrule(lr){2-4} \cmidrule(lr){5-7} 
 Methods & Sync$\uparrow$ & LMD$\downarrow$ & AUE$\downarrow$ & Sync$\uparrow$ & LMD$\downarrow$ & AUE$\downarrow$     \\ \midrule 

Ground Truth                                 & $7.850$ & $0$    & $0  $  & $6.976$ & $0$ & $0 $  \\
\arrayrulecolor{black!50}\specialrule{0.1ex}{0.2ex}{0.2ex}
Wav2Lip~\cite{prajwal2020wav2lip}            & \cellcolor{tabsecond} {$8.272$} & $7.039$  &\cellcolor{tabfirst} {$4.154$} & \cellcolor{tabfirst} {$7.907$}  & $5.561$ & $3.967$  \\
PC-AVS~\cite{zhou2021pcavs}  & \cellcolor{tabfirst} {$8.408$} & $7.754$  & $6.278$ & \cellcolor{tabsecond} {$7.533$}  & $6.560$ & $4.518$\\
\arrayrulecolor{black!50}\specialrule{0.1ex}{0.2ex}{0.2ex}
AD-NeRF~\cite{guo2021adnerf}              & $5.670$ & $7.378$  & $4.736$ & $5.076$  & \cellcolor{tabthird} {$5.542$} & $3.711$ \\
RAD-NeRF~\cite{tang2022radnerf}        & $6.532$ & \cellcolor{tabsecond} {$5.848$}  & $4.717$ & $5.472$  & $5.599$ & \cellcolor{tabthird} {$3.666$}\\
ER-NeRF~\cite{li2023ernerf}             & $6.507$ & \cellcolor{tabthird} {$6.181$}  & \cellcolor{tabsecond} {$4.489$} & $5.160$  & \cellcolor{tabsecond} {$5.374$} & \cellcolor{tabsecond} {$3.519$} \\
\arrayrulecolor{black!50}\specialrule{0.1ex}{0.2ex}{0.2ex}
\textbf{Talk3D (Ours) }          & \cellcolor{tabthird} {$6.827$} & \cellcolor{tabfirst} {$5.352$}  & \cellcolor{tabthird} {$4.693$} & \cellcolor{tabthird} {$5.780$}  & \cellcolor{tabfirst} {$4.814$} & \cellcolor{tabfirst} {$3.132$} \\
\arrayrulecolor{black!100}\bottomrule
\end{tabular}
}
\end{center}
\label{tab:crossdriven}
\vspace{-5pt}

\end{table}

\subsection{Experimental Settings}\label{sec:setting}
\paragrapht{Dataset. } 
To perform audio-driven talking head synthesis, we require a few minutes of speaking portrait video paired with an audio track. Specifically, in order to compare with the state-of-the-art method, we directly employ datasets from \cite{lu2021lsp, guo2021adnerf}, comprising person-centric videos averaging 6,000 frames at 25 fps. Following the training methodology of previous NeRF-based works~\cite{guo2021adnerf, tang2022radnerf, li2023ernerf}, we split the video into training and testing sets.
Furthermore, we utilize a pre-trained Wav2Vec model~\cite{baevski2020wav2vec} to extract audio features from each speech audio.

\paragrapht{Comparison baselines. }
We compare our method with 2D talking head research, such as Wav2Lip \cite{prajwal2020wav2lip} and PC-AVS \cite{zhou2021pcavs}. In addition, we also compare our method with several NeRF-based models: AD-NeRF \cite{guo2021adnerf}, RAD-NeRF~\cite{tang2022radnerf}, and ER-NeRF~\cite{li2023ernerf}. Furthermore, we evaluate our method directly on the ground-truth to provide a clearer comparison.

In the supplementary material (Sec. 2), we extend our comparisons to include GeneFace~\cite{ye2022geneface} and HFA-GP~\cite{bai2023hfagp}. is a powerful NeRF-based talking portrait synthesis model but is designed for different settings compared to our paper. HFA-GP is also a noteworthy related work, but the comparison is conducted separately due to the instability of its source code.

\subsection{Quantitative Evaluation}
\paragrapht{Comparison settings and metrics. }
In quantitative evaluation, we compare the synthesized quality of the head and synchronized lips by constructing three distinct settings: 1) the \emph{novel-view synthesis} experiment, 2) the \emph{self-driven} experiment, and 3) the \emph{cross-driven} experiment.
The first  \emph{novel-view synthesis} setting involves assessing the robustness of viewpoint editing by rendering from diverse novel viewpoints. Specifically, we alter a certain amount of angle degree (yaw: $-30^{\circ}$$\sim$$30^{\circ}$, pitch: $-20^{\circ}$$\sim$$20^{\circ}$) from the canonical viewpoint.
The \emph{self-driven} shares the same training settings as \emph{novel-view synthesis} differing only in the rendering camera viewpoints, which are extracted from the ground-truth video.
Finally, for the \emph{cross-driven} setting, the model is driven by entirely unrelated audio tracks to measure lip synchronization performance using two extracted audio clips from demos of SynObama~\cite{suwajanakorn2017synthesizing}. 

In the \emph{self-driven} setting, we utilize evaluation metrics commonly employed in talking portrait synthesis research, including peak signal-to-noise ratio (\textbf{PSNR}), structural similarity index measure (\textbf{SSIM}), learned perceptual image patch similarity (\textbf{LPIPS})~\cite{zhang2018lpips} to evaluate image reconstruction quality. 
Given the absence of ground-truth images for the same identity in the other two settings, we adhere to the methodology of previous works~\cite{guo2021adnerf, tang2022radnerf, li2023ernerf} and introduce identity-agnostic metrics such as the Fréchet inception distance (\textbf{FID})~\cite{heusel2017fid}, landmark distance (\textbf{LMD})~\cite{chen2018lmd}, SyncNet confidence score (\textbf{Sync})~\cite{chung2017syncnet1, chung2017syncnet2} for lip synchronization, and action units error (\textbf{AUE})~\cite{baltrusaitis2018openface, baltruvsaitis2015openface2} to evaluate face motion accuracy.
We also use an identity similarity metric (\textbf{ID-SIM}) calculated using an off-the-shelf identity detection network~\cite{huang2020curricularface} to measure the similarity of facial identities.
For a fair comparison, each generated result is cropped and rescaled to the facial area into the same cropping region. Furthermore, since the NeRF-based methods utilize a pre-defined background extracted from the video, we exclusively measure reconstruction metrics on the facial region of the generated result. 
\begin{figure}[!t]
\newcolumntype{M}[1]{>{\centering\arraybackslash}m{#1}}
\setlength{\tabcolsep}{0.2pt}
\renewcommand{\arraystretch}{0.1}
\centering

\scriptsize
\begin{tabular}{ M{0.09\linewidth} M{0.11\linewidth} M{0.11\linewidth} M{0.11\linewidth} M{0.11\linewidth}   @{\hskip 0.01\linewidth} M{0.11\linewidth} M{0.11\linewidth} M{0.11\linewidth} M{0.11\linewidth}  }

$\mathbf{y}$, $\mathbf{p}$
& $\texttt{+}30^{\circ},\texttt{+}20^{\circ}$ & $\texttt{+}15^{\circ}, \texttt{+}10^{\circ}$ & $\texttt{-}15^{\circ}, \texttt{-}10^{\circ}$ & $\texttt{-}30^{\circ}, \texttt{-}20^{\circ}$ & $\texttt{+}30^{\circ}, \texttt{+}20^{\circ}$ & $\texttt{+}15^{\circ}, \texttt{+}10^{\circ}$ & $\texttt{-}15^{\circ}, \texttt{-}10^{\circ}$ & $\texttt{-}30^{\circ}, \texttt{-}20^{\circ}$  \\ \\ \\

AD-NeRF &
 
\includegraphics[width=\linewidth]{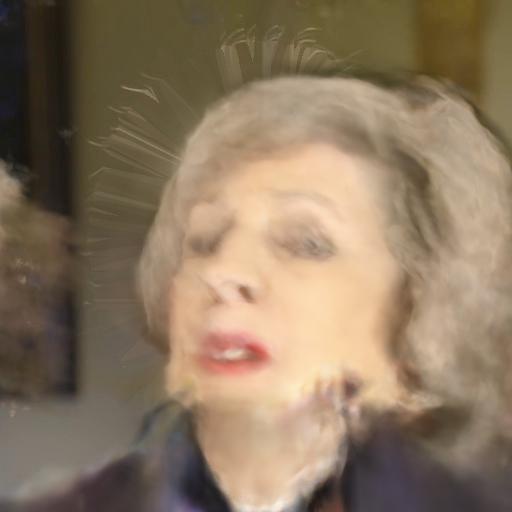}\hfill &
\includegraphics[width=\linewidth]{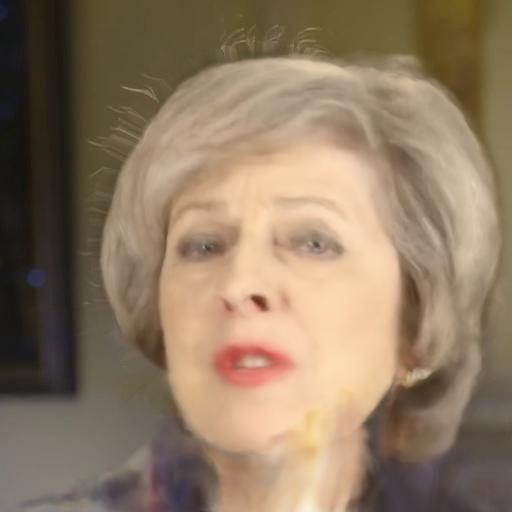}\hfill &
\includegraphics[width=\linewidth]{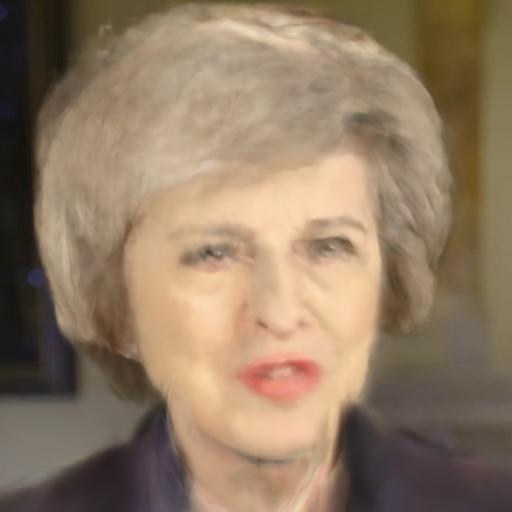}\hfill &
\includegraphics[width=\linewidth]{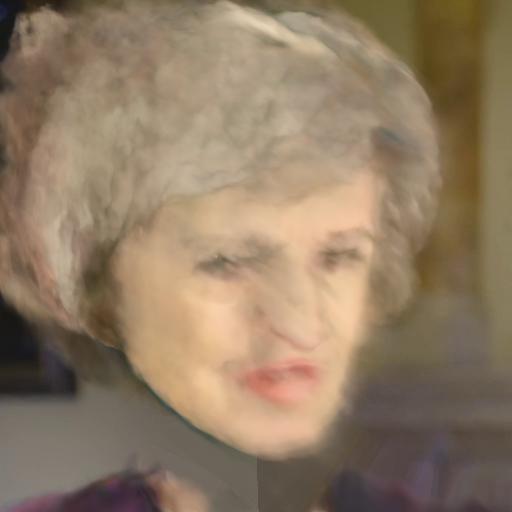}\hfill &
\includegraphics[width=\linewidth]{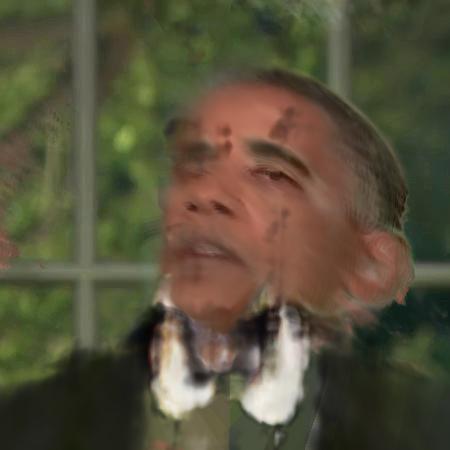}\hfill &
\includegraphics[width=\linewidth]{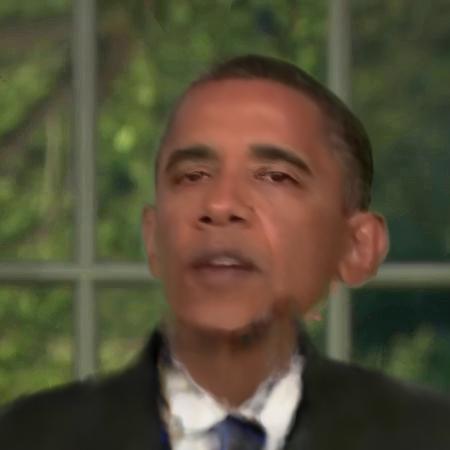}\hfill &
\includegraphics[width=\linewidth]{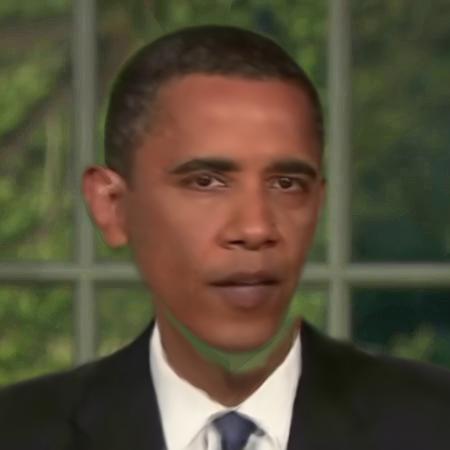}\hfill &
\includegraphics[width=\linewidth]{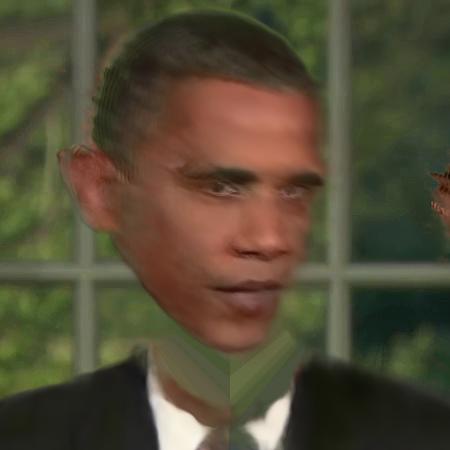}\hfill \\

RAD-NeRF &

\includegraphics[width=\linewidth]{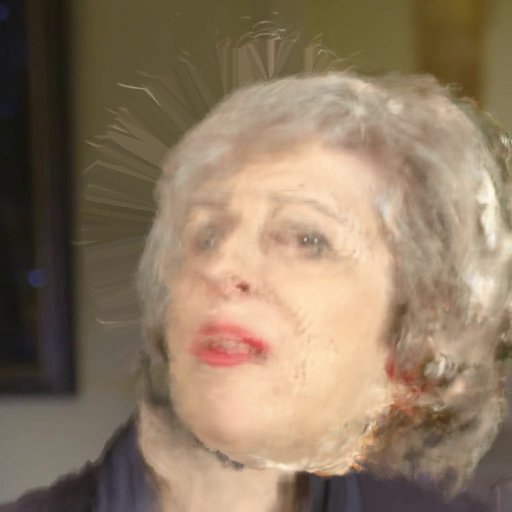}\hfill &
\includegraphics[width=\linewidth]{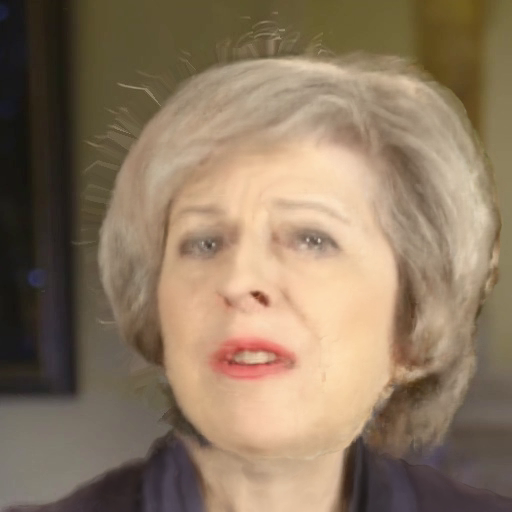}\hfill &
\includegraphics[width=\linewidth]{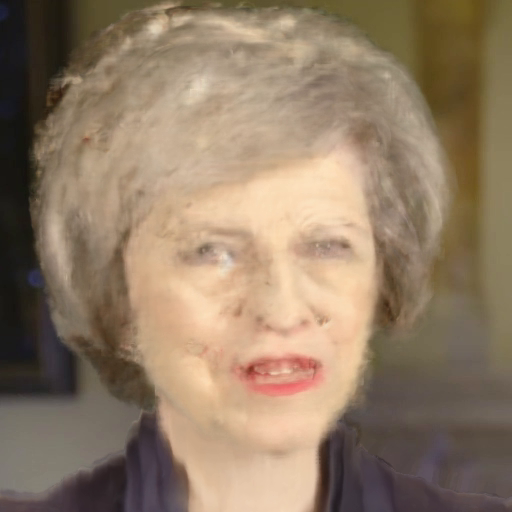}\hfill &
\includegraphics[width=\linewidth]{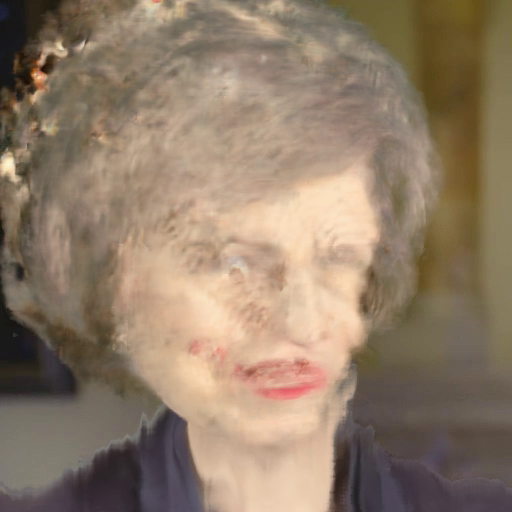}\hfill &
\includegraphics[width=\linewidth]{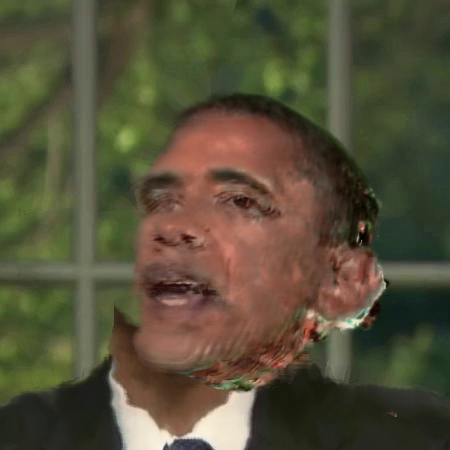}\hfill &
\includegraphics[width=\linewidth]{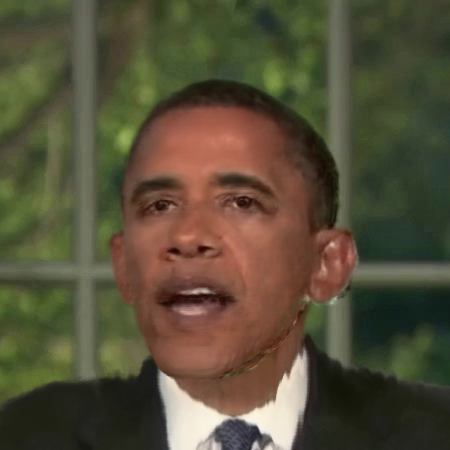}\hfill &
\includegraphics[width=\linewidth]{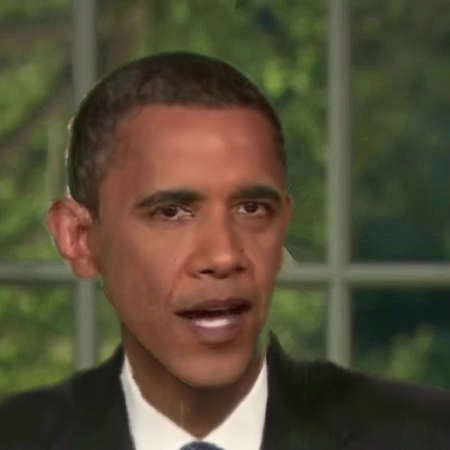}\hfill &
\includegraphics[width=\linewidth]{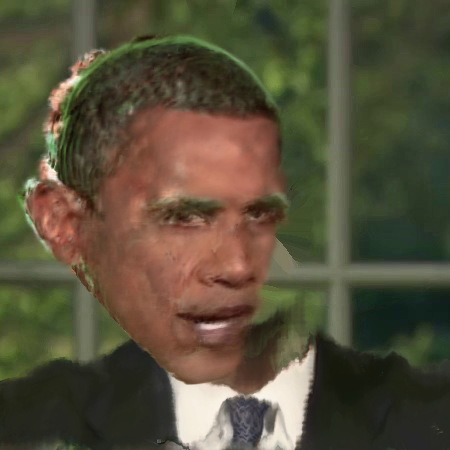}\hfill\\

ER-NeRF &

\includegraphics[width=\linewidth]{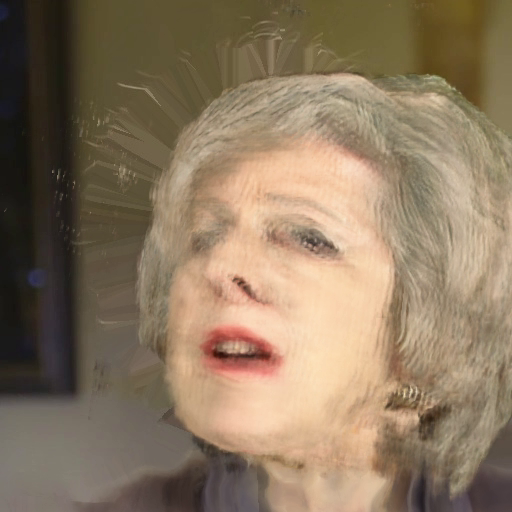}\hfill &
\includegraphics[width=\linewidth]{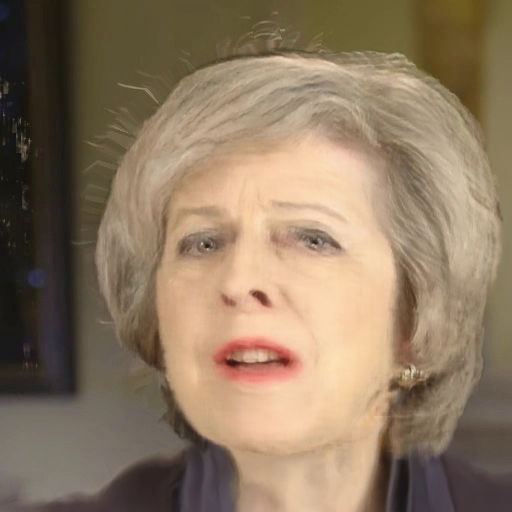}\hfill &
\includegraphics[width=\linewidth]{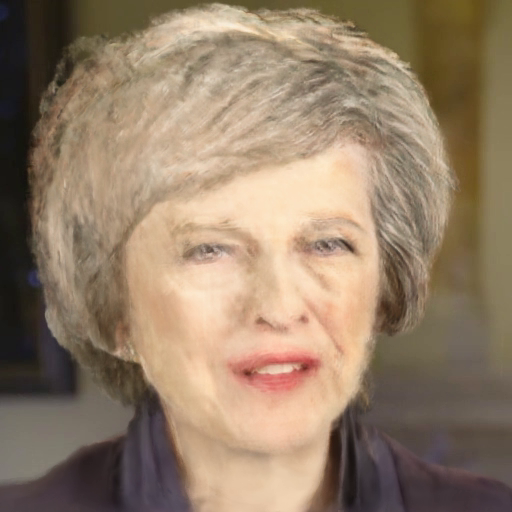}\hfill &
\includegraphics[width=\linewidth]{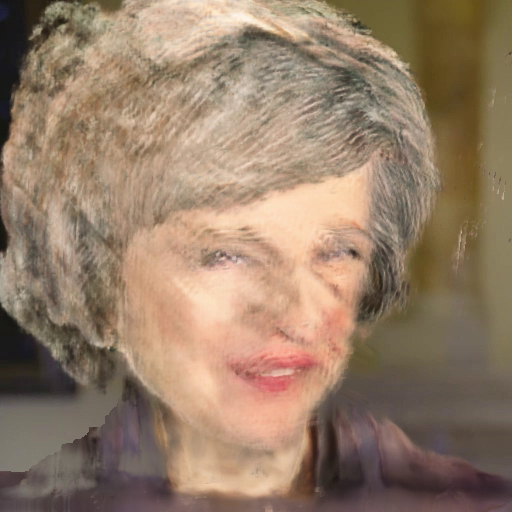}\hfill &
\includegraphics[width=\linewidth]{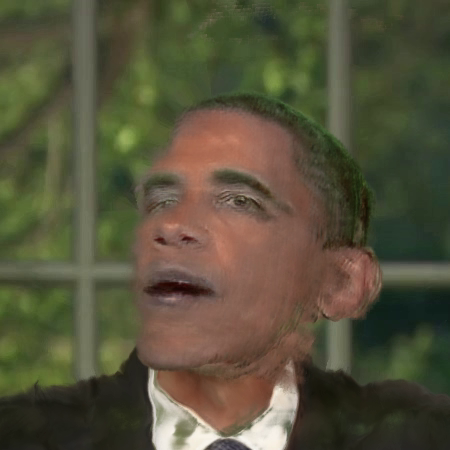}\hfill &
\includegraphics[width=\linewidth]{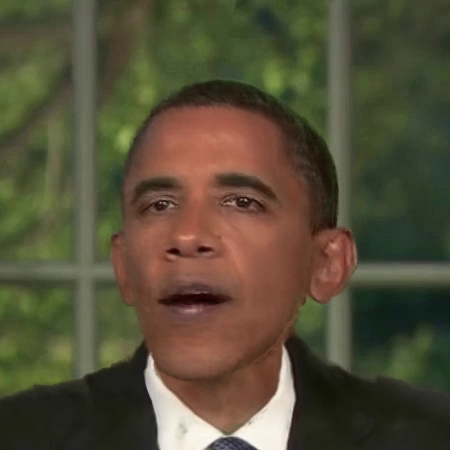}\hfill &
\includegraphics[width=\linewidth]{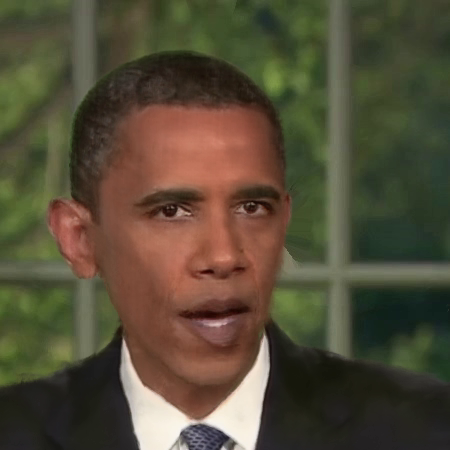}\hfill &
\includegraphics[width=\linewidth]{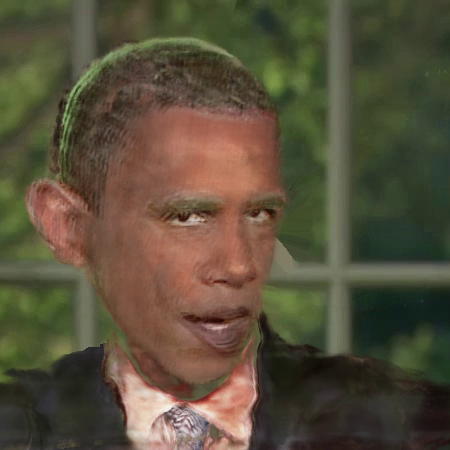}\hfill\\

\textbf{Talk3D (Ours)} &

\includegraphics[width=\linewidth]{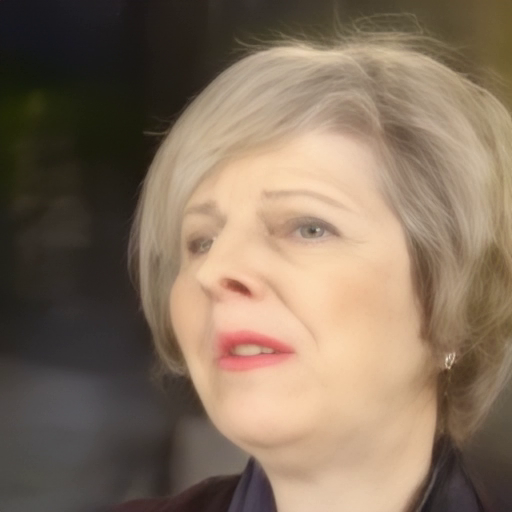}\hfill &
\includegraphics[width=\linewidth]{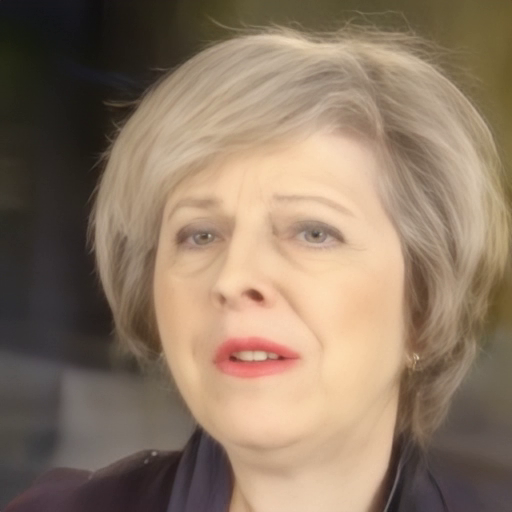}\hfill &
\includegraphics[width=\linewidth]{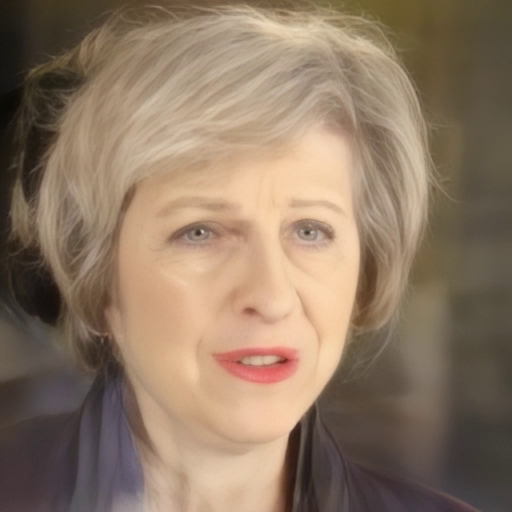}\hfill &
\includegraphics[width=\linewidth]{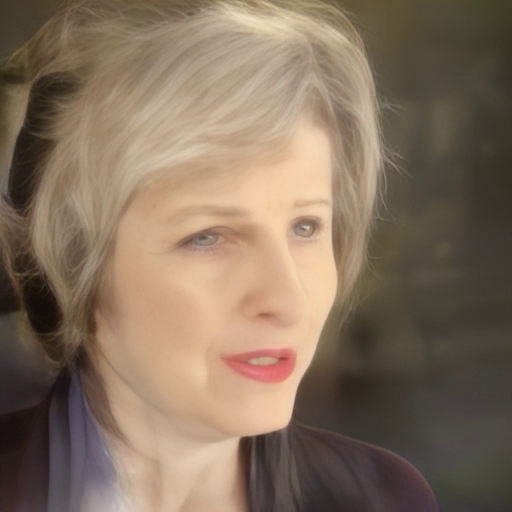}\hfill &
\includegraphics[width=\linewidth]{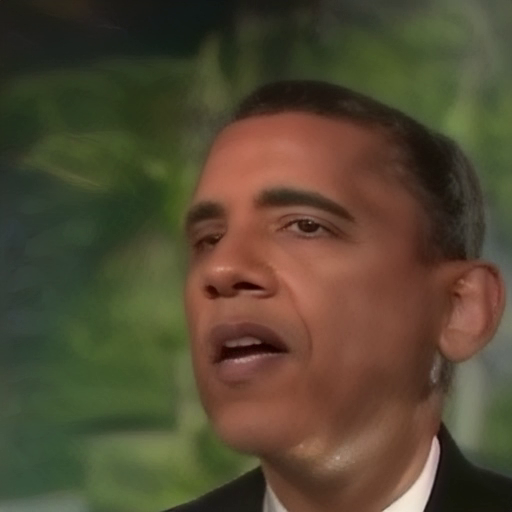}\hfill &
\includegraphics[width=\linewidth]{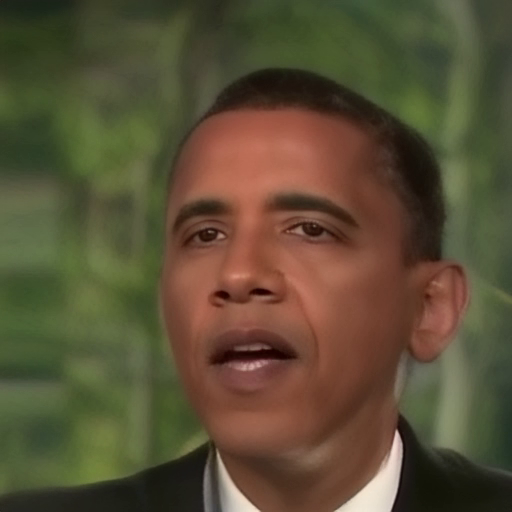}\hfill &
\includegraphics[width=\linewidth]{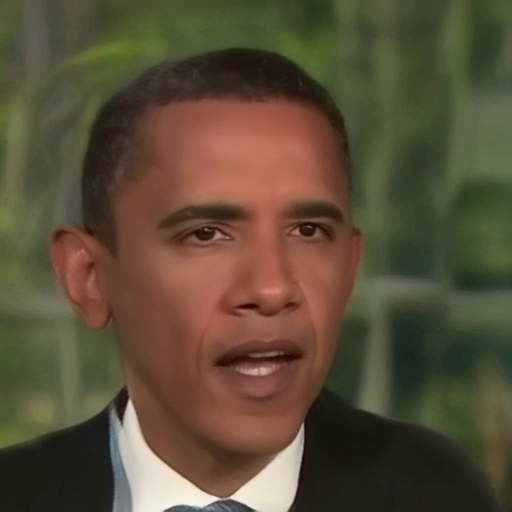}\hfill &
\includegraphics[width=\linewidth]{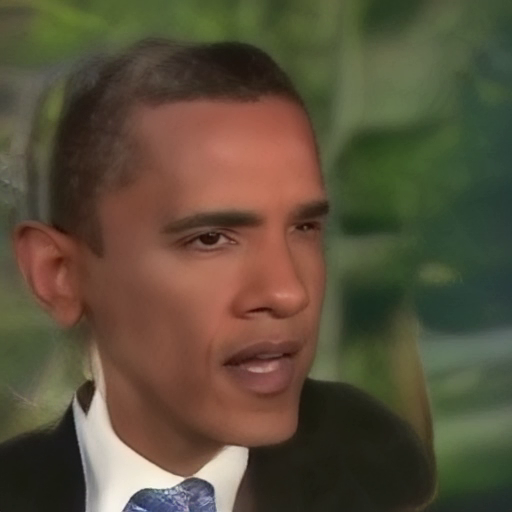}\hfill\\

\end{tabular}
\vspace{-5pt}
\caption{\textbf{Visualization of synthesized portraits from head poses unseen during training.} We show a randomly selected frame from synthesized talking portraits using different rendered at different yaw and pitch ($\mathbf{y}$, $\mathbf{p}$) angles. Our method demonstrates its robustness on rendering facial images at large head angles which are rarely shown in the training video.}
\label{fig:ood_pose}
\vspace{-10pt}
\end{figure}

\paragrapht{Novel pose synthesis evaluation. }
The result of the \emph{novel-view synthesis} setting is shown in \tabref{tab:ood_pose}. We compare our method against prior works capable of explicit control of the camera viewpoint. 
While the majority of methods exhibit comparable performance in frontal view rendering, a notable decline in scores is observed for other NeRF-based techniques when the viewpoint is rotated to different angles. 
In contrast, our method demonstrates consistently high scores across all metrics, highlighting its efficacy in maintaining performance across diverse viewing angles compared to its counterparts.

\paragrapht{Self-driven evaluation. } 
The \emph{self-driven} evaluation results are presented in \tabref{tab:selfdriven}. Our method achieves the best quality in most image quality metrics, while also showing the best lip synchronization among NeRF-based methods. Despite the one-shot 2D-based methods, Wav2Lip and PC-AVS present superior Sync scores, their subpar scores in image fidelity highlight their inadequacy in the accurate reconstruction of the specific portrait. Our method demonstrates superior performance in lip synchronization metrics, while also showing comparable image fidelity to other NeRF-based methods. 
\label{paragraph:selfdriven_quan}

\paragrapht{Cross-driven evaluation. }
The evaluation results for the \emph{cross-driven} setting, as depicted in \tabref{tab:crossdriven}, demonstrate the successful performance on general audio input to synthesize the corresponding lip movement. 
Our model consistently presents the highest scores on most comparisons among NeRF-based methods. The additional use of the sync loss function enforces the model to produce accurate lip shapes even from audio unseen during training. This distinguishes our approach from previous works, which cannot employ sync loss.

\begin{figure*}[!t]
\newcolumntype{M}[1]{>{\centering\arraybackslash}m{#1}}
\setlength{\tabcolsep}{0.2pt}
\renewcommand{\arraystretch}{0.1}
\centering
\scriptsize
\begin{tabular}{M{0.09\linewidth}@{\hskip 0.005\linewidth} M{0.11\linewidth}M{0.11\linewidth}M{0.11\linewidth}M{0.11\linewidth}  @{\hskip 0.01\linewidth} M{0.11\linewidth} M{0.11\linewidth} M{0.11\linewidth}M{0.11\linewidth}}

& \textcolor{red}{who}ever & \textcolor{red}{wo}rk & \vspace{1.7pt}\textcolor{red}{new} & \textcolor{red}{p}eacefully & \textcolor{red}{re}quire & with\textcolor{red}{ou}t & safe\textcolor{red}{ty} & $\langle$ mute $\rangle$ \\ \\

\makecell{\vspace{-6pt}\\ GT} &
\includegraphics[width=\linewidth]{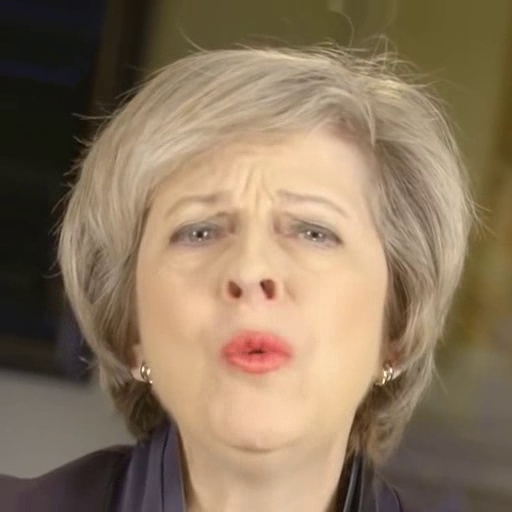}\hfill &
\includegraphics[width=\linewidth]{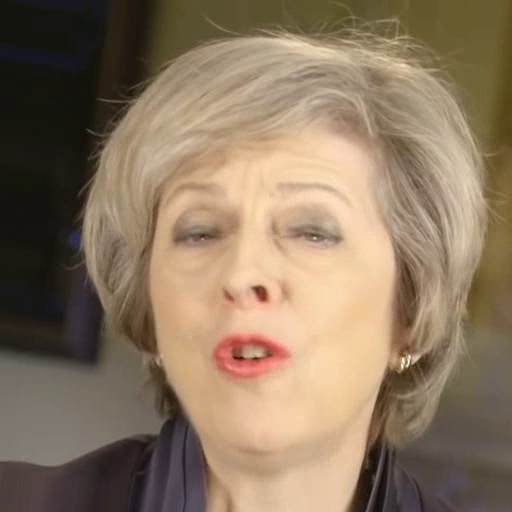}\hfill &
\includegraphics[width=\linewidth]{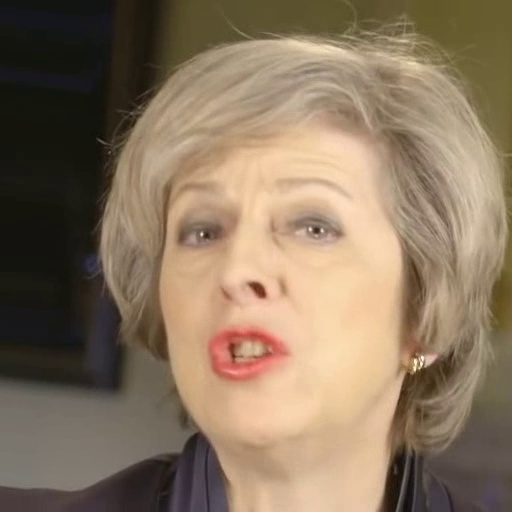}\hfill &
\includegraphics[width=\linewidth]{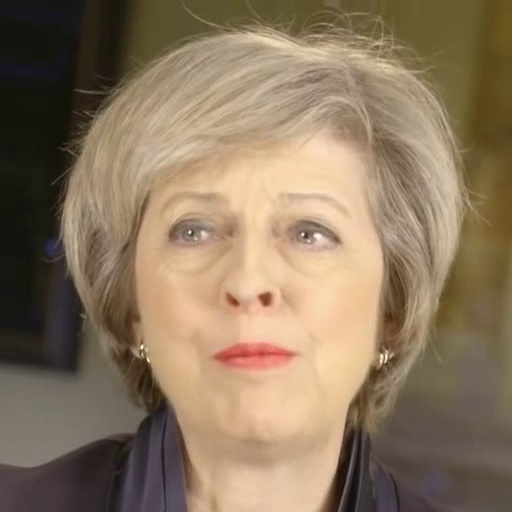}\hfill &
\includegraphics[width=\linewidth]{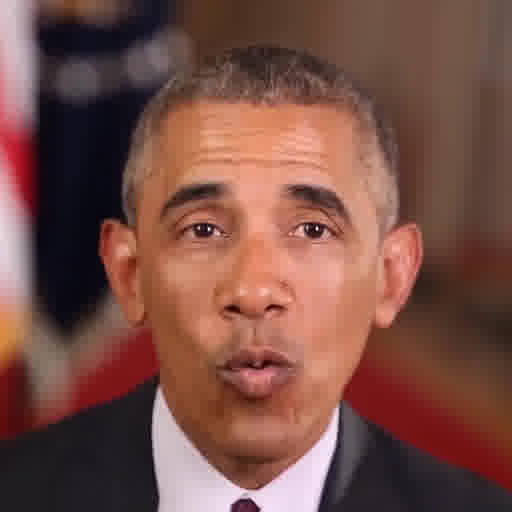}\hfill &
\includegraphics[width=\linewidth]{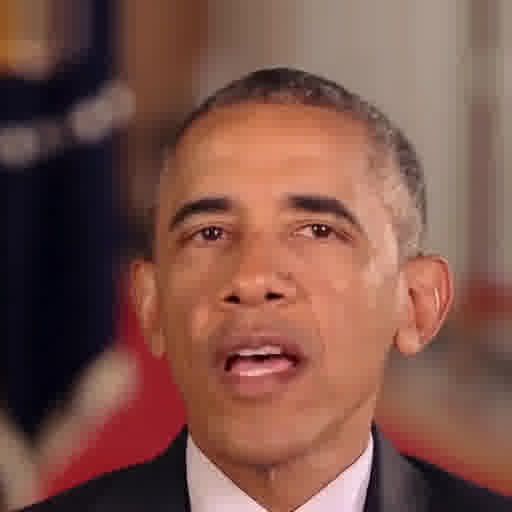}\hfill &
\includegraphics[width=\linewidth]{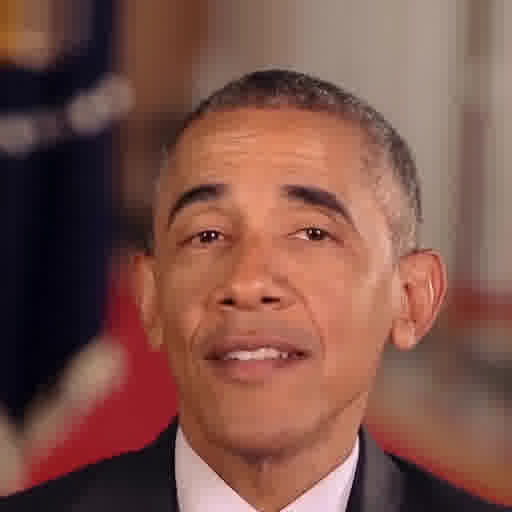}\hfill &
\includegraphics[width=\linewidth]{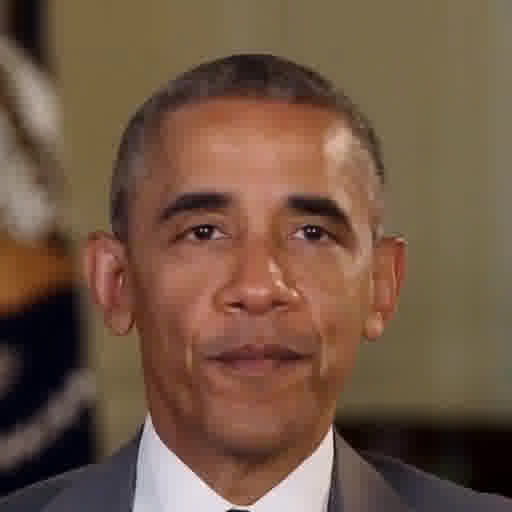}\hfill \\

Wav2Lip &
\includegraphics[width=\linewidth]{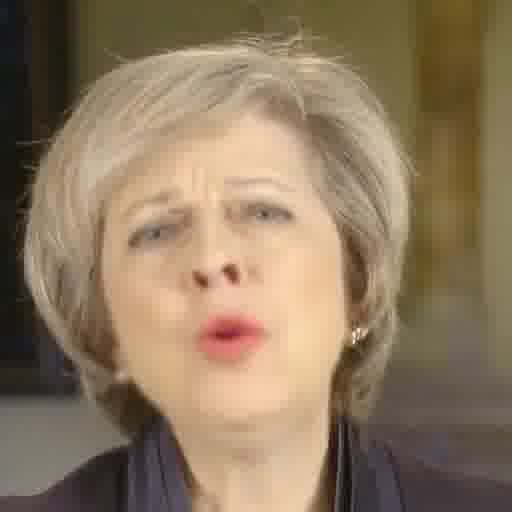}\hfill &
\includegraphics[width=\linewidth]{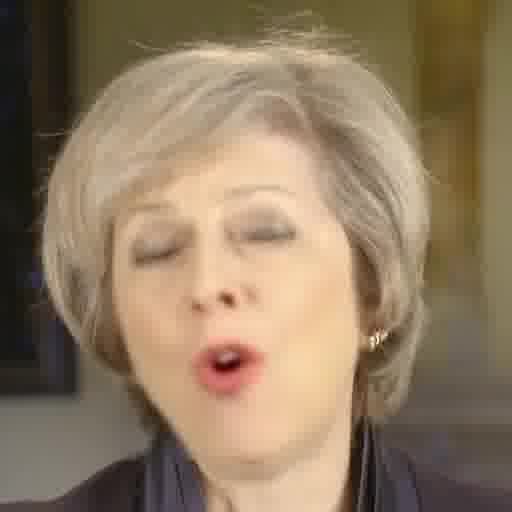}\hfill &
\includegraphics[width=\linewidth]{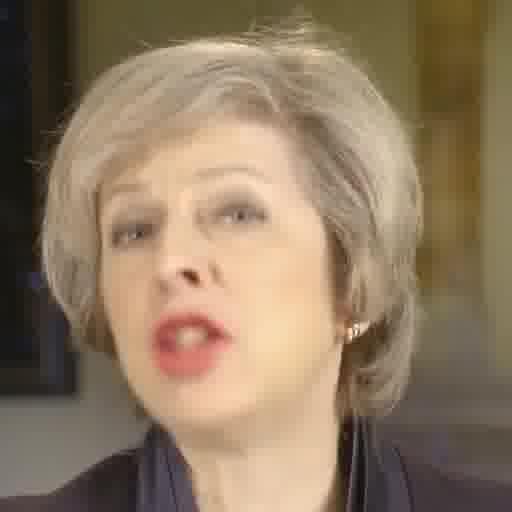}\hfill &
\includegraphics[width=\linewidth]{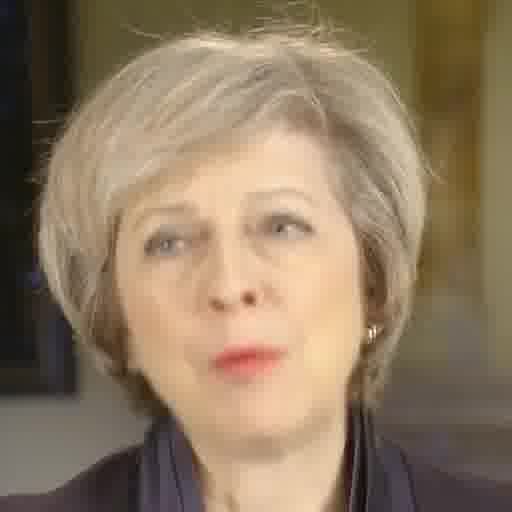}\hfill &
\includegraphics[width=\linewidth]{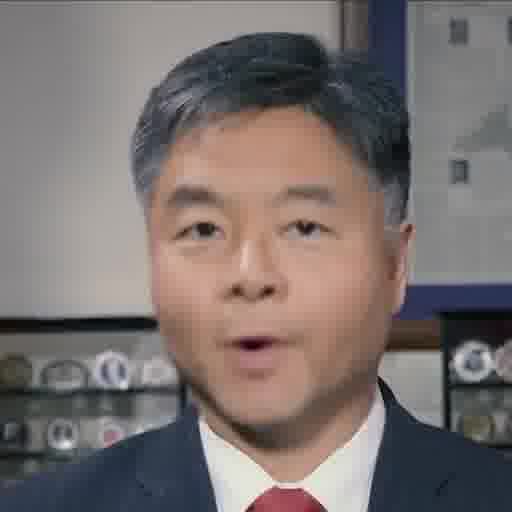}\hfill &
\includegraphics[width=\linewidth]{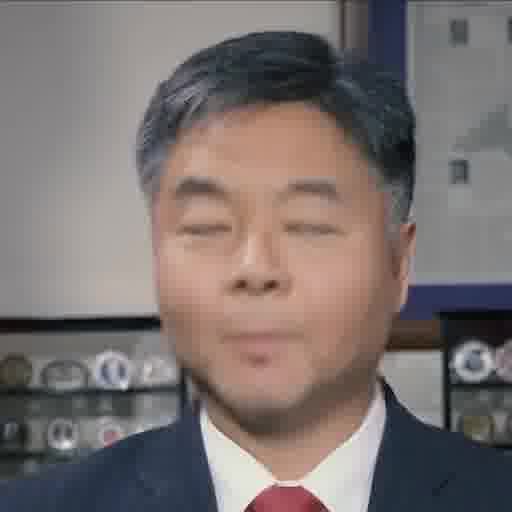}\hfill &
\includegraphics[width=\linewidth]{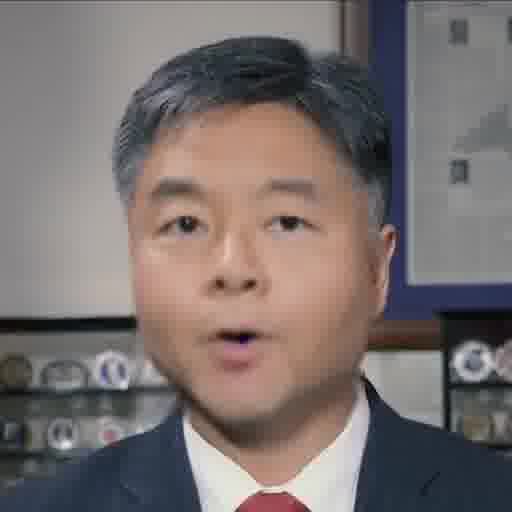} &
\includegraphics[width=\linewidth]{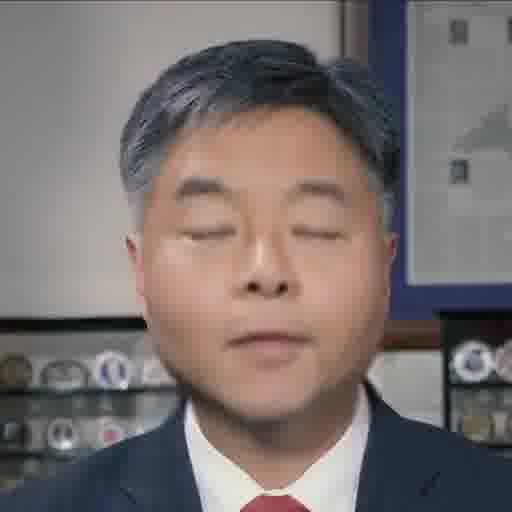}\hfill \\

PC-AVS &
\includegraphics[width=\linewidth]{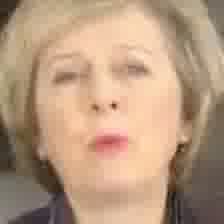}\hfill &
\includegraphics[width=\linewidth]{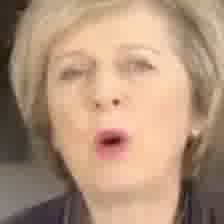}\hfill &
\includegraphics[width=\linewidth]{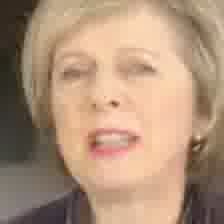}\hfill &
\includegraphics[width=\linewidth]{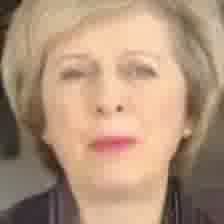}\hfill &
\includegraphics[width=\linewidth]{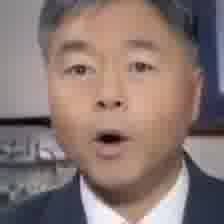}\hfill &
\includegraphics[width=\linewidth]{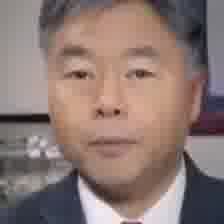}\hfill &
\includegraphics[width=\linewidth]{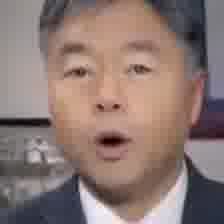}\hfill &
\includegraphics[width=\linewidth]{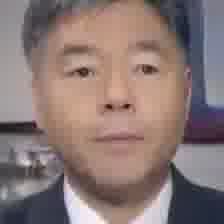}\hfill \\

AD-NeRF &
\includegraphics[width=\linewidth]{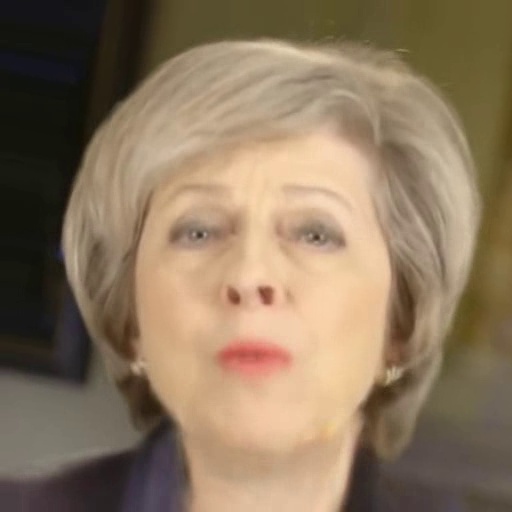}\hfill &
\includegraphics[width=\linewidth]{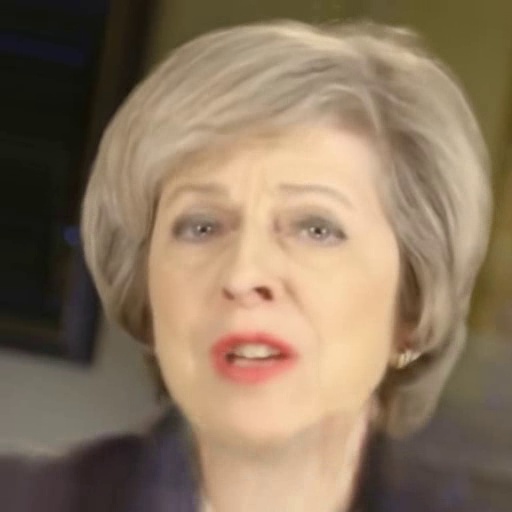}\hfill &
\includegraphics[width=\linewidth]{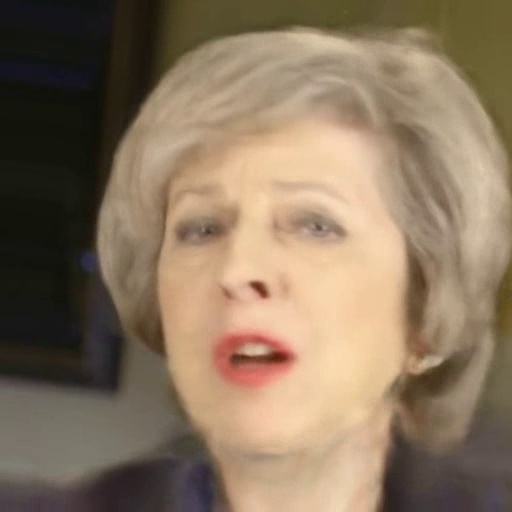}\hfill &
\includegraphics[width=\linewidth]{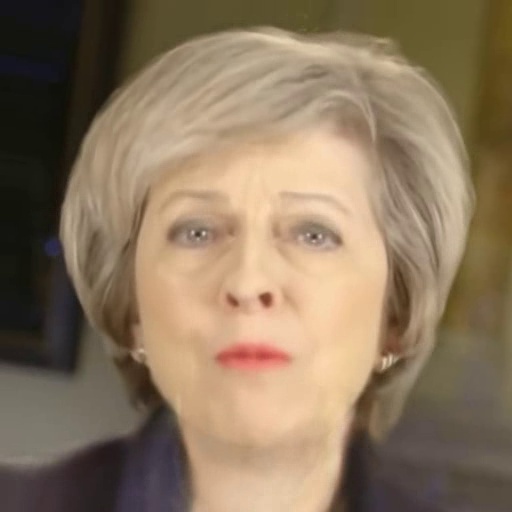}\hfill &
\includegraphics[width=\linewidth]{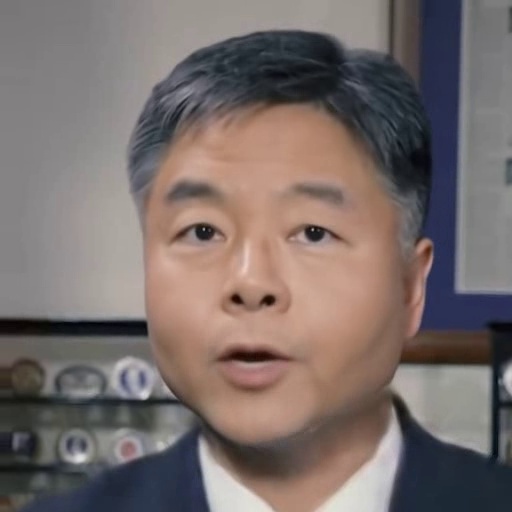}\hfill &
\includegraphics[width=\linewidth]{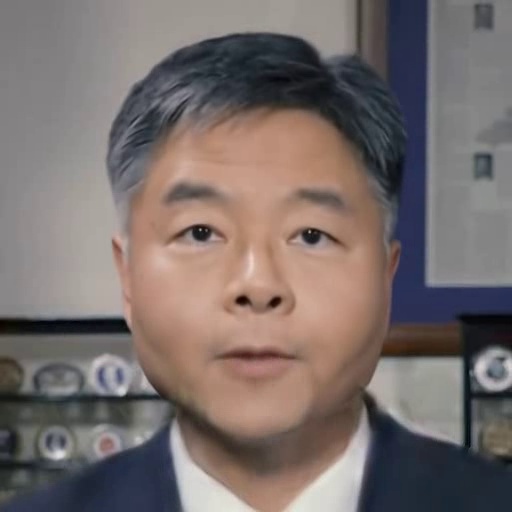}\hfill &
\includegraphics[width=\linewidth]{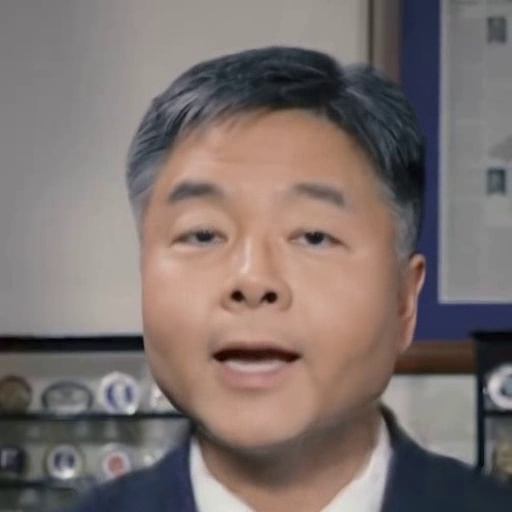}\hfill &
\includegraphics[width=\linewidth]{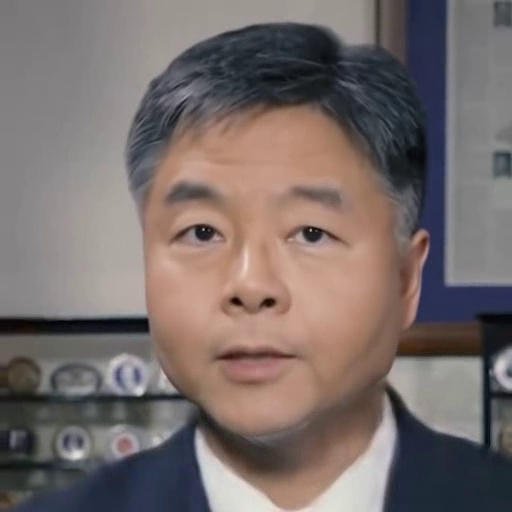}\hfill \\

RAD-NeRF &
\includegraphics[width=\linewidth]{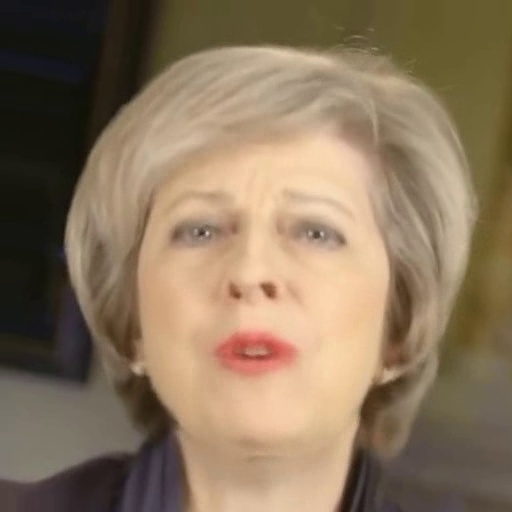}\hfill &
\includegraphics[width=\linewidth]{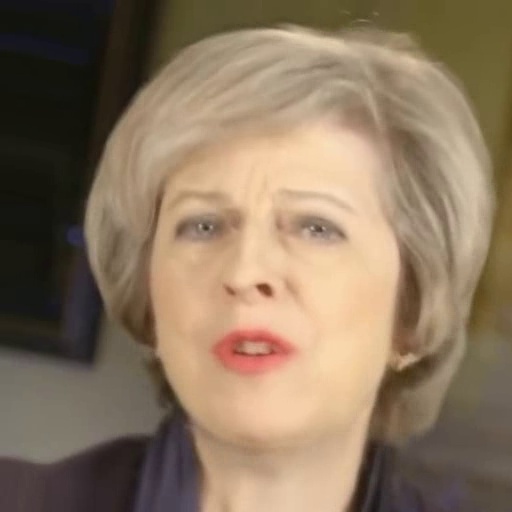}\hfill &
\includegraphics[width=\linewidth]{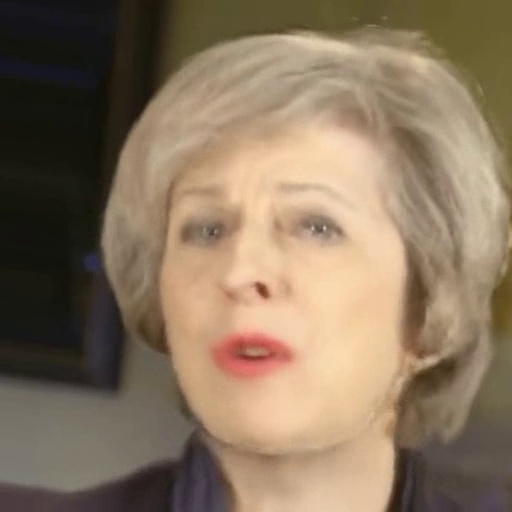}\hfill &
\includegraphics[width=\linewidth]{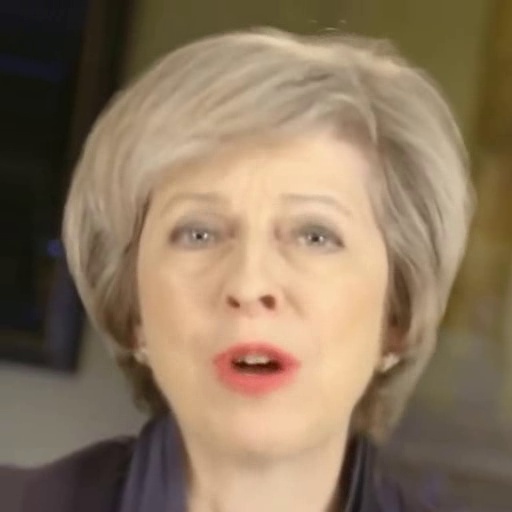}\hfill &
\includegraphics[width=\linewidth]{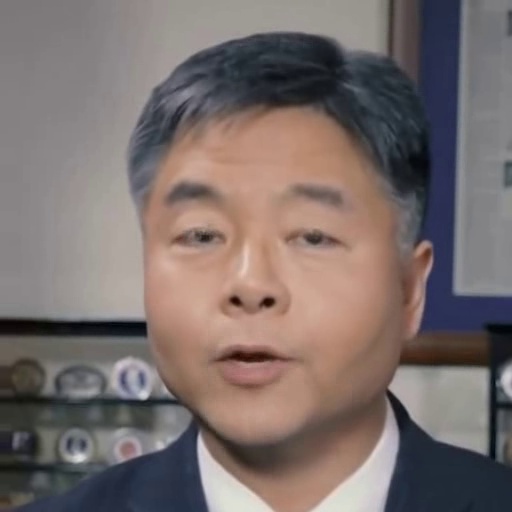}\hfill &
\includegraphics[width=\linewidth]{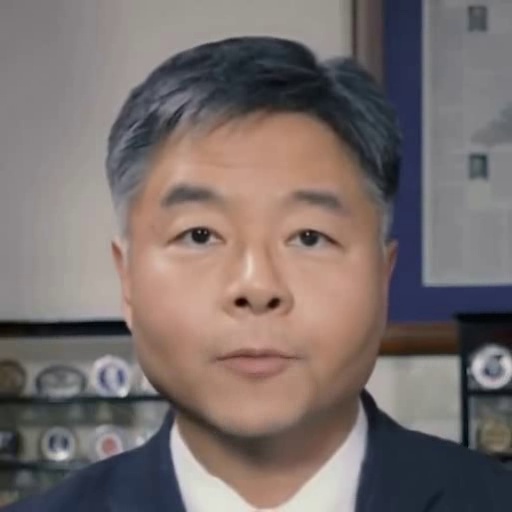}\hfill &
\includegraphics[width=\linewidth]{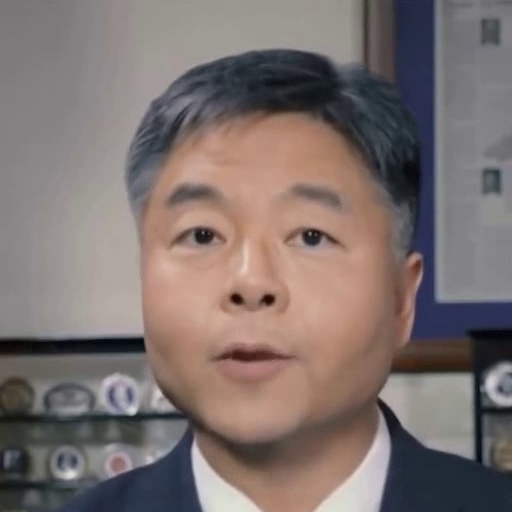}\hfill &
\includegraphics[width=\linewidth]{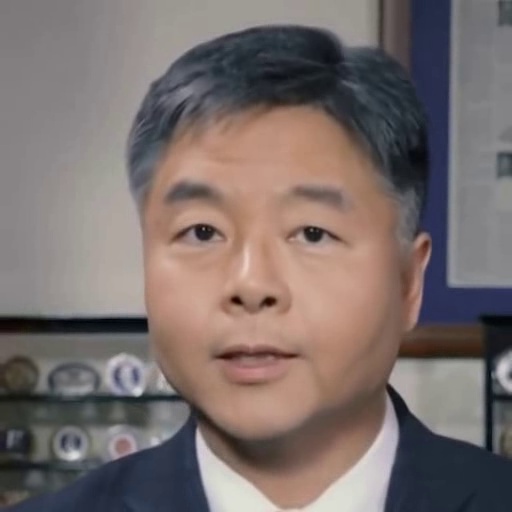}\hfill \\

ER-NeRF &
\includegraphics[width=\linewidth]{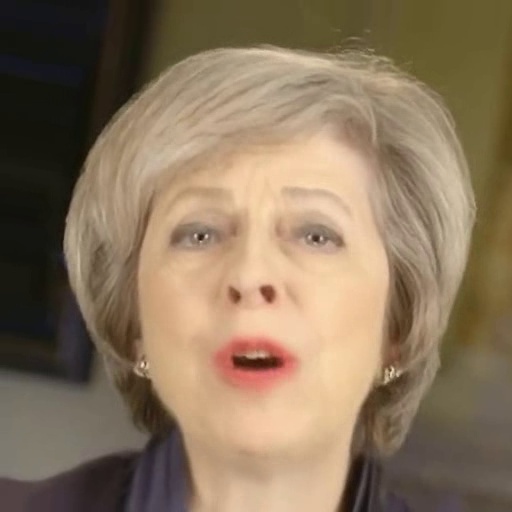}\hfill &
\includegraphics[width=\linewidth]{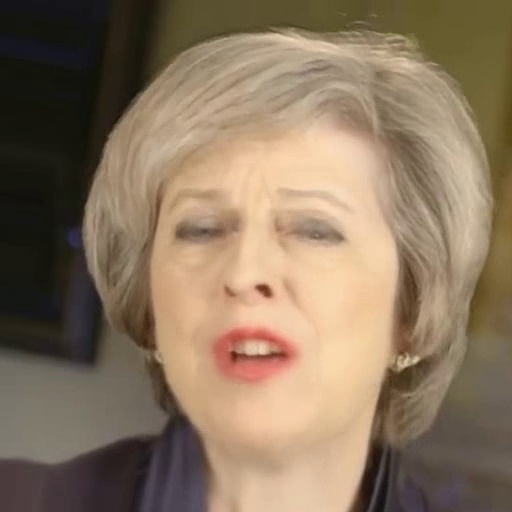}\hfill &
\includegraphics[width=\linewidth]{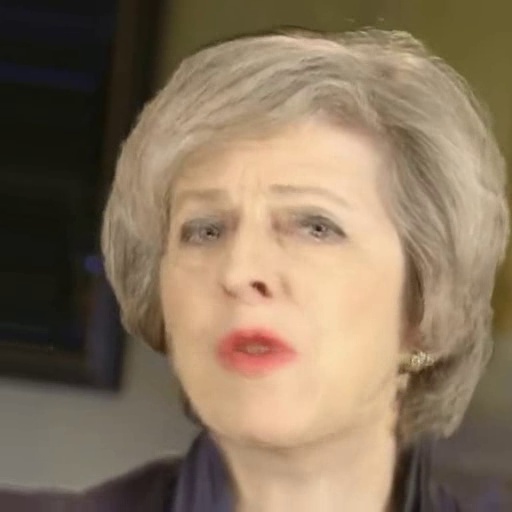}\hfill &
\includegraphics[width=\linewidth]{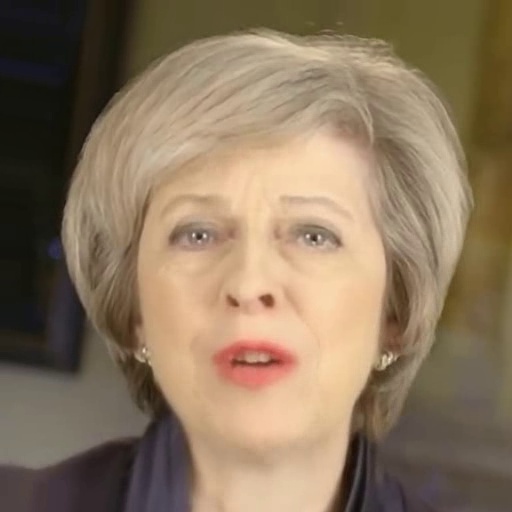}\hfill &
\includegraphics[width=\linewidth]{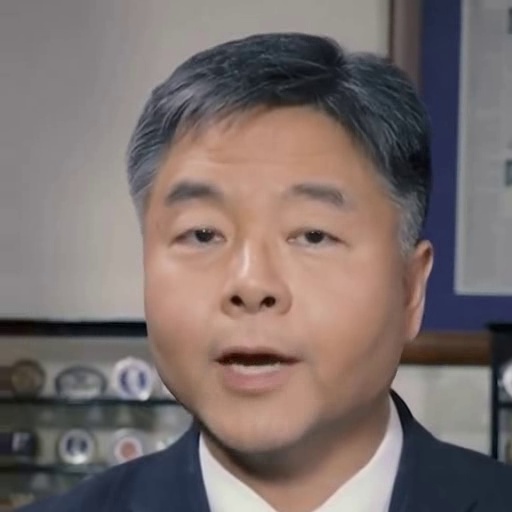}\hfill &
\includegraphics[width=\linewidth]{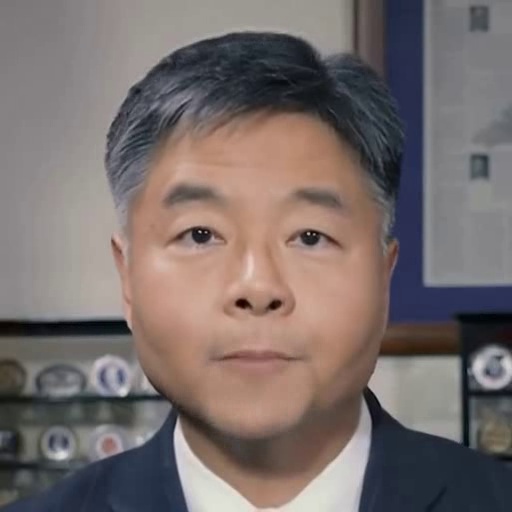}\hfill &
\includegraphics[width=\linewidth]{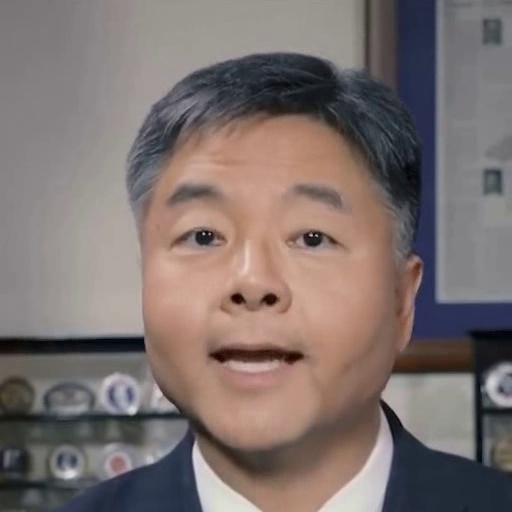}\hfill &
\includegraphics[width=\linewidth]{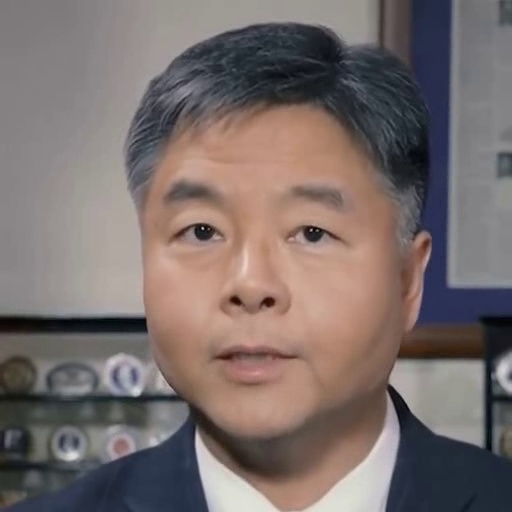}\hfill \\

\textbf{Talk3D (Ours)} &
\includegraphics[width=\linewidth]{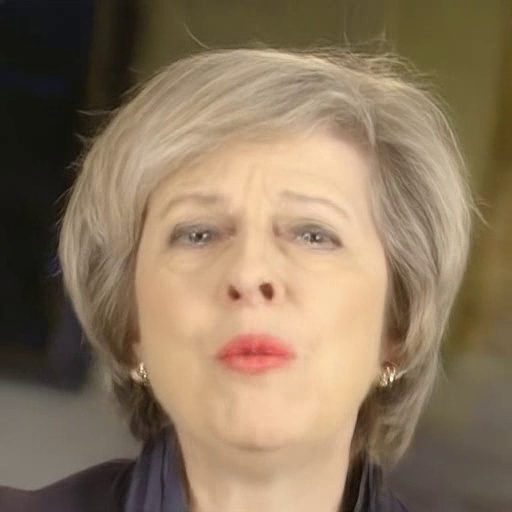}\hfill &
\includegraphics[width=\linewidth]{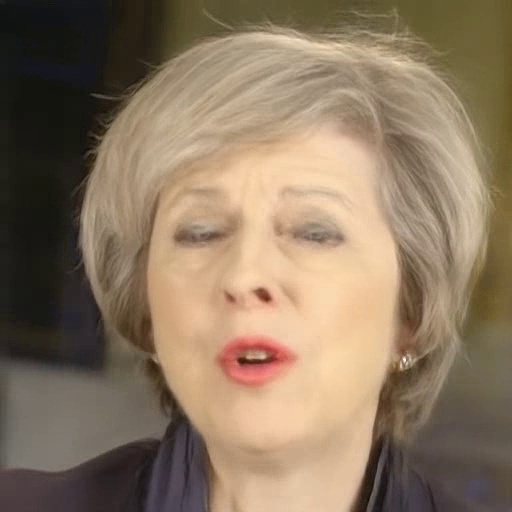}\hfill &
\includegraphics[width=\linewidth]{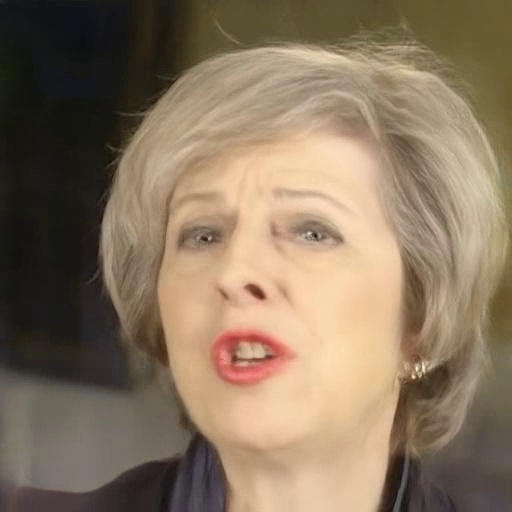}\hfill &
\includegraphics[width=\linewidth]{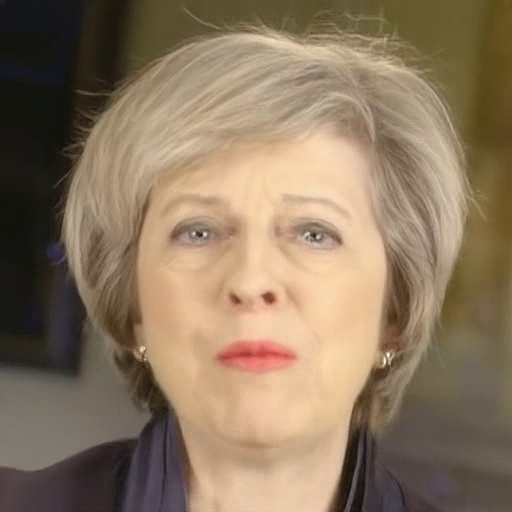}\hfill &
\includegraphics[width=\linewidth]{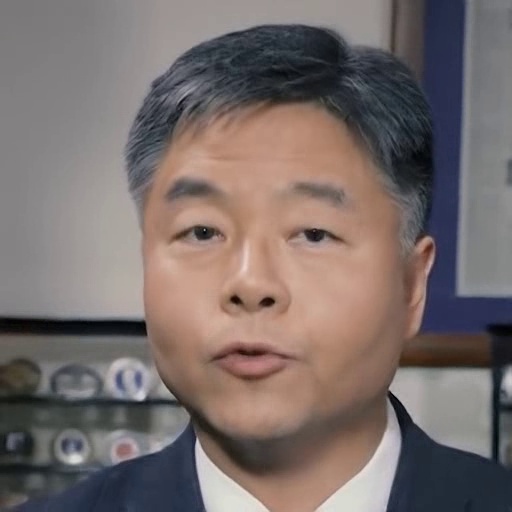}\hfill &
\includegraphics[width=\linewidth]{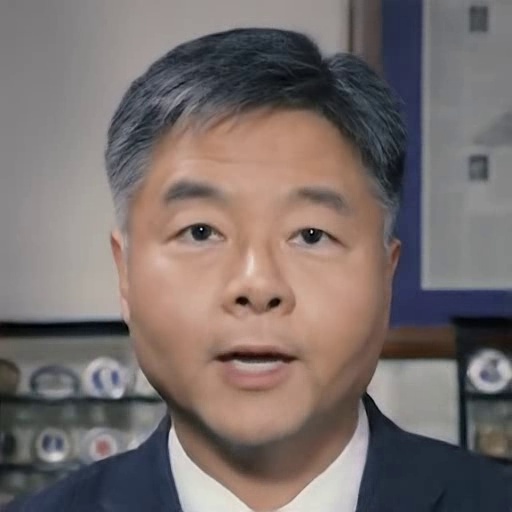}\hfill &
\includegraphics[width=\linewidth]{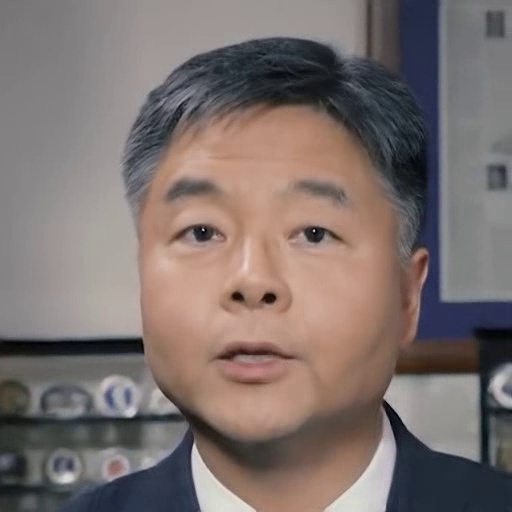}\hfill &
\includegraphics[width=\linewidth]{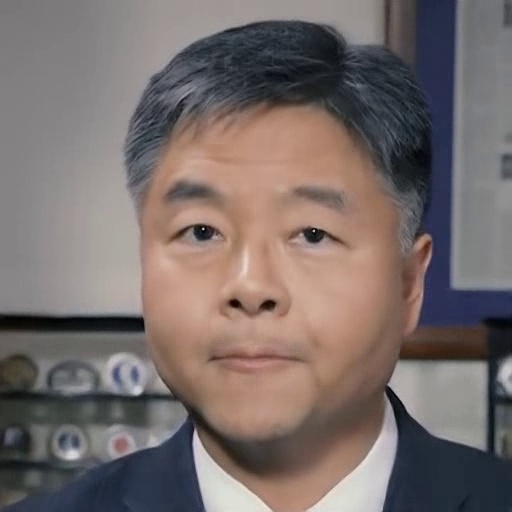}\hfill \\

\end{tabular}
\vspace{-5pt}
\caption{\textbf{The comparison of the keyframes and details of generated portraits.} We show visualizations of our method and previous methods under the self-driven setting (left side) and the cross-driven setting (right side). Best viewed in zoom.}
\label{fig:full_qual}
\vspace{-10pt}
\end{figure*}

\subsection{Qualitative Evaluation}
In this section, we present generated results from each of the evaluation settings. We first visualize the generated facial images rendered from various viewpoints. As can be seen in \figref{fig:ood_pose}, previous NeRF-based methods suffer from performance degradation when generated from camera angles far from the canonical viewpoint, with its results frequently showing inconsistent facial color and artifacts. 
Especially for the torso region, RAD-NeRF and ER-NeRF experience a substantial decline in quality caused by their torso modeling namely \emph{pseudo-3D deformable module}, showing irregular torso boundary and geometry deterioration.
For AD-NeRF, where the head and torso volumes are learned independently, the synthesized heads appear disembodied when viewed from side angles.
We also show the sampled results from \emph{self-driven} and \emph{cross-driven} experiments in \figref{fig:full_qual}. We choose four key frames from each of the two experiments to compare the lip-sync accuracy and reconstruction quality.
As mentioned in \ref{paragraph:selfdriven_quan}, although 2D-based methods such as Wav2lip and PC-AVS demonstrate high lip-sync accuracy, the generated results cannot fully reconstruct the given scene. 
Although the NeRF-based methods manage to create a full portrait, the separated rendering pipeline shows unnatural head movements at the neck part, and also shows blurry textures, especially at the hair part.
In contrast, our Talk3D demonstrates robust and accurate results, attributed to its unified generation process.
Notably, Talk3D avoids the issue of open lips in the absence of speech audio, demonstrating accurate results by appropriately closing the lips, an aspect where other NeRF-based methods falter.

\begin{table*}[t]
\vspace{-5pt}
\caption{\textbf{User study results.} The rating is of scale 1-5, the higher the better. The top, second-best, and third-best results are shown in \textcolor{red}{red}, \textcolor{orange}{orange}, and  \textcolor{yellow}{yellow}, respectively.}
\vspace{-5pt}
\definecolor{tabfirst}{rgb}{1, 0.7, 0.7} 
\definecolor{tabsecond}{rgb}{1, 0.85, 0.7} 
\definecolor{tabthird}{rgb}{1, 1, 0.7} 

\setlength{\tabcolsep}{4pt}
\resizebox{1\linewidth}{!}{
\centering

\begin{tabular}{llcccccc}
\toprule
Settings & Methods           & Wav2Lip \cite{prajwal2020wav2lip} & PC-AVS \cite{zhou2021pcavs}  & AD-NeRF \cite{guo2021adnerf}  & RAD-NeRF \cite{tang2022radnerf} & ER-NeRF~\cite{li2023ernerf} & \textbf{Talk3D(Ours)}  \\ \midrule
\multirow{3}{*}{\makecell{novel-view \\ synthesis}} &
Lip-sync Accuracy & $-$  & $-$  & $2.056$  & \cellcolor{tabthird}$2.411$  & \cellcolor{tabsecond} $2.983$  & \cellcolor{tabfirst}{$3.103$}  \\
&Image Quality     & $-$  & $-$  & $0.924$  & \cellcolor{tabthird}$1.417$  & \cellcolor{tabsecond}$2.532$  & \cellcolor{tabfirst}{$3.123$}  \\
&Video Realness    & $-$  & $-$  & \cellcolor{tabthird} $1.205$  & $0.834$  & \cellcolor{tabsecond}$2.163$  & \cellcolor{tabfirst}{$2.242$}  \\  \midrule
\multirow{3}{*}{self-driven} &
Lip-sync Accuracy & \cellcolor{tabfirst}{$3.455$}  & $2.511$  & $2.455$  & $2.636$  & \cellcolor{tabthird} $2.909$  & \cellcolor{tabsecond}{$3.394$}  \\
&Image Quality     & $2.623$  & $0.607$  & \cellcolor{tabthird} $3.723$  & $3.650$  & \cellcolor{tabsecond}$3.789$  & \cellcolor{tabfirst}{$3.970$}  \\
&Video Realness    & $2.868$  & $0.757$  & $2.936$  & \cellcolor{tabthird} $2.991$  & \cellcolor{tabsecond}$3.223$  & \cellcolor{tabfirst}{$3.467$}  \\  \midrule
\multirow{3}{*}{cross-driven} &
Lip-sync Accuracy & \cellcolor{tabsecond}$2.933$  & $1.767$  & \cellcolor{tabthird} $2.867$  & $2.467$  & $2.667$  & \cellcolor{tabfirst}{$3.301$}  \\
&Image Quality     & $2.967$  & $0.767$  & \cellcolor{tabsecond}$3.733$  & $3.441$  & \cellcolor{tabthird}$3.763$  & \cellcolor{tabfirst}{$3.798$}  \\
&Video Realness    & $2.801$  & $0.878$  & \cellcolor{tabsecond}$3.233$  & $2.731$  & \cellcolor{tabthird}$3.183$  & \cellcolor{tabfirst}{$3.267$}  \\  \bottomrule
\end{tabular}
}

\label{tab:user_study}
\end{table*}

\subsection{User Study}
We present a user study to assess the visual quality of the generated heads. We invited 31 participants to compare 9 randomly selected video clips from the quantitative evaluation of the main study. 
Utilizing the mean opinion scores (MOS) rating protocol, participants first provided ratings for the generated videos of the \emph{novel-view synthesis setting}, \emph{self-driven setting} and \emph{cross-driven setting}, each based on three criteria: (1) lip-sync accuracy; (2) image quality; and (3) video realness. 
The average scores for each method are presented in \tabref{tab:user_study}, revealing that our Talk3D outperforms most of the criteria. These results demonstrate the outstanding visual quality of our method, in light of both facial reconstruction and novel view synthesis. 

\begin{table}[!t]
\caption{\textbf{Ablation study} on the sync loss function.}
\vspace{-8pt}
\centering
\newcommand{\RomanNumeralCaps}[1]{\MakeUppercase{\romannumeral #1}}
\newcolumntype{M}[1]{>{\centering\arraybackslash}m{#1}}
\scalebox{0.8}{
\setlength{\tabcolsep}{4pt}
\begin{tabular}{@{}lccccc@{}}
\toprule
Method  & PSNR $\uparrow$ & LPIPS $\downarrow$ & LMD $\downarrow$ & AUE $\downarrow$ & Sync $\uparrow$ \\ \midrule
    Ground Truth  & -              & -            & 0           &  0       & $8.605$  \\\arrayrulecolor{black!50}\specialrule{0.1ex}{0.2ex}{0.2ex}
w/o sync &  $26.180$  & $0.068$  &   $3.149$ & $1.715$  & $6.137$ \\

All (Ours) & $26.799$        & $0.054$    &$3.227$  & $1.540$  & $6.529$  \\\arrayrulecolor{black!100} \bottomrule
\end{tabular}
}
\label{tab:loss_ablation}
\end{table}

\subsection{Ablation Study}
In this section, we present the ablation study to validate the efficacy of our primary contributions. All ablation studies are conducted under a slightly different setting than the \emph{self-driven} scenario, with the key distinction being the measurement of metrics on the entire image pixels.

\paragrapht{Use of the sync loss. }
Due to the computationally expensive nature of NeRF limits full-image rendering during training time,
prior NeRF-based works~\cite{guo2021adnerf, liu2022sspnerf, tang2022radnerf, li2023ernerf} solely employ pixel-based MSE loss and patch-wise LPIPS loss. On the other hand, leveraging the efficient representation of EG3D, our model is capable of utilizing full image-based loss functions such as the sync loss function. 
 In \tabref{tab:loss_ablation}, we assess the significance of the sync loss by comparing results without its utilization. While forgoing the sync loss function marginally enhances reconstruction accuracy, it is essential for generating well-synchronized lips.

\definecolor{tabfirst}{rgb}{1, 0.7, 0.7} 
\definecolor{tabsecond}{rgb}{1, 0.85, 0.7} 
\definecolor{tabthird}{rgb}{1, 1, 0.7} 

\begin{table}[!t]
\caption{\textbf{Ablation study} on use of each feature token.  }
\vspace{-8pt}
\centering
\newcolumntype{M}[1]{>{\centering\arraybackslash}m{#1}}
\setlength{\tabcolsep}{4pt}
\scalebox{0.85}{
\begin{tabular}{@{}lccccc@{}}
\toprule
Method  & PSNR $\uparrow$ & LPIPS $\downarrow$ & LMD $\downarrow$ & AUE $\downarrow$ & Sync $\uparrow$ \\ \midrule
    Ground Truth  & -              & -            & $3.322$        &  $1.815$      & $8.605$     \\
\arrayrulecolor{black!50}\specialrule{0.1ex}{0.2ex}{0.2ex}
w/o null-vec & $25.745$  & $0.064$              & \cellcolor{tabfirst} {$2.781$}            & \cellcolor{tabthird} {$1.650$}            & $6.267$         \\
w/o eye feature & $25.862$           & $0.062$    & $3.335$   & \cellcolor{tabsecond} {$1.598$}   & \cellcolor{tabthird} {$6.414$} \\ 
w/o landmark tokens & \cellcolor{tabsecond} {$26.195$}  & \cellcolor{tabsecond} {$0.059$}              & $3.392 $           & $1.719$         & $5.498$          \\
w/o angle tokens & \cellcolor{tabthird} {$26.152$}           & \cellcolor{tabthird} {$0.060$}     & \cellcolor{tabthird} {$3.313$}  & $1.920$   & \cellcolor{tabsecond} {$6.508$}  \\ 
\arrayrulecolor{black!50}\specialrule{0.1ex}{0.2ex}{0.2ex}
All (Ours) & \cellcolor{tabfirst} {$26.799$} \         & \cellcolor{tabfirst} {$0.054$}    & \cellcolor{tabsecond} {$3.227$}  & \cellcolor{tabfirst} {$1.540$}  & \cellcolor{tabfirst} {$6.529$}  \\ \arrayrulecolor{black!100}\bottomrule
\end{tabular}
}
\label{tab:ablation_tokens}

\vspace{15pt}
\caption{\textbf{Ablation study} on specific design selections for deltaplane prediction.}
\vspace{-8pt}
\setlength{\tabcolsep}{4pt}
\centering
\scalebox{0.85}{
    \begin{tabular}{@{}lccccc@{}}
    \toprule
    Method  & PSNR $\uparrow$ & LPIPS $\downarrow$ & LMD $\downarrow$ & AUE $\downarrow$ & Sync $\uparrow$ \\ \midrule
    Ground Truth  & -              & -            & $0$           &  $0$       & $8.605$  \\\arrayrulecolor{black!50}\specialrule{0.1ex}{0.2ex}{0.2ex}
    w/o deltaplane &$19.180$  &  $0.187$  & $4.675$  & $2.939$        & $1.192$        \\
    w/o attention & \cellcolor{tabthird} {$24.925$} &   \cellcolor{tabthird} {$0.071$}    & \cellcolor{tabthird} {$3.485$}          & \cellcolor{tabthird} {$2.115$}          & \cellcolor{tabthird} {$4.591$}         \\
    w/o split & $24.403$   &  $0.096$   & $3.793$  & $2.730$ & $1.024$ \\ 
    w/o rollout & \cellcolor{tabsecond}{$25.621$} &  \cellcolor{tabsecond} {$0.064$} &  \cellcolor{tabsecond}{$3.233$}   &   \cellcolor{tabsecond}{$1.962$}   & \cellcolor{tabsecond}{$6.438$}           \\ \arrayrulecolor{black!50}\specialrule{0.1ex}{0.2ex}{0.2ex}
All (Ours) & \cellcolor{tabfirst} {$26.799$} \         & \cellcolor{tabfirst}{$0.054$}    & \cellcolor{tabfirst} {$3.227$}  & \cellcolor{tabfirst} {$1.540$}  & \cellcolor{tabfirst} {$6.529$}  \\ \arrayrulecolor{black!100}\bottomrule
    \end{tabular}
}
\setlength{\abovecaptionskip}{4pt}

\label{tab:ablation_network_design}
\vspace{-10pt}
\end{table}

\paragrapht{Feature token selection.}
We also investigate the significance of using augmented conditions, such as eye blink, head rotation, and facial landmarks. In \tabref{tab:ablation_tokens}, we measure the impact of each feature on image fidelity by turning them on and off in turn. The lower PSNR scores are attributed to the low lip-sync accuracy due to the feature entanglement
or the absence of eye blink features preventing proper eye closure. We also show detailed visualizations in Sec. 4 of the supplementary material.

\paragrapht{Deltaplane predictor design. }
We further ablate on the network designs of deltaplane predictor.~\tabref{tab:ablation_network_design} shows four different design choices, which are predicting $\mathbf{w}$ latent vector instead of deltaplane (w/o deltaplane), replacing attention module to conditional affine layer (w/o attention), merging split-convolution layer into a single convolution (w/o split), and removing roll-out method utilized in \secref{paragraph:split_conv}. Our model exhibits superior image quality and lip-sync accuracy compared to other design choices, highlighting the effectiveness of our model architecture.

\section{Conclusion}

In this paper, we introduced \textbf{Talk3D}, a novel framework that incorporates 3D-aware GAN prior and a region-aware motion for high-fidelity 3D talking head synthesis. 
Our framework incorporates a personalized generator fine-tuned using the VIVE3D framework, allowing for the synthesis of 3D-aware talking head avatars with its realistic geometry and explicit rendering viewpoint control.
Furthermore, our proposed audio-guided attention U-Net architecture enhances the disentanglement of local variations within image frames, such as background, torso, and eye movements. 
Through extensive experiments, we demonstrate that our proposed model not only produces accurate lip movements corresponding to the input audio but also enables rendering from novel viewpoints, addressing limitations observed in previous state-of-the-art approaches.
We anticipate our work will significantly impact digital media experiences, and virtual interactions, and find applications in film-making, virtual avatars, and video conferencing.

\title{- Supplementary Material -} 

\appendix

\newpage
\noindent {\Large \textbf{Appendix}}
\vspace{10pt}

\noindent In the following, we provide implementation details and offer further analysis of our experiment along with extensive qualitative results. 
Specifically, in~\secref{supsec:implementation_details} we describe
the implementation details including data preprocessing pipeline, feature extraction, and network architecture. 
In~\secref{supsec:result_and_comparison}, we present additional experiments and comparisons that demonstrate the robustness and effectiveness of our method.
In~\secref{supsec:further_analysis}, we also provide further analyses to support the efficacy of our contributions, presenting qualitative results of ablation studies, and visualizations of attention map and triplane.
In~\secref{supsec:editing}, we also introduce semantic manipulation over the generated portraits.
Finally, in~\secref{supsec:video} and~\secref{supsec:broader_impact}, we briefly explain our supplementary video and further discuss the limitations and ethical considerations of our research. 

\section{Additional Implementation Details}
\label{supsec:implementation_details}
\subsection{Pre-processing}
We follow the same image cropping as VIVE3D~\cite{fruhstuck2023vive3d}. They detect the 6 facial landmarks from every video frame, using an off-the-shelf detector~\cite{bulat2017far} and perform Gaussian smoothing on the landmarks along the temporal axis to stabilize the transition of the cropping area. They additionally detect landmarks on a single reference image generated from the personalized generator. This reference image serves as an anchor for every frame to calculate the affine transformation matrices. Using these affine matrices, we calculate the cropping boundaries for each of the raw images. More specifically, they utilize a slightly wider cropping boundary compared to  EG3D~\cite{chan2022efficient} which employs Deep3DFace~\cite{deng2019accurate} for image cropping.

\subsection{Augmenting feature extraction} 
For the audio feature extraction, we follow~\cite{tang2022radnerf} which employ the pre-trained Wav2Vec~\cite{baevski2020wav2vec} model and further encode with several layers of 1D convolutions. On the other hand, the augmenting features such as the eye (scalar factor), head rotation angles (3-dimensional vector), and landmarks (6-dimensional vector) are comparatively low-dimensional feature vectors. Therefore, we upsample these augmenting features using the positional encodings and further encode with several layers of MLP. Each of the output features is a 64-dimensional feature token and is fed to our cross-attention network $\mathcal{F}_\mathrm{CA}$.

\subsection{Network architecture}
\subsubsection{Attention network.}
\begin{sloppypar}
Our deltaplane predictor $\mathcal{F}$ first encodes the 256-resolution triplane $\mathbf{P}$ into 32-resolution feature map $\mathbf{E}$, while its hidden dimension is upscaled from 32 to 256.
\end{sloppypar}
With given flattened
image feature vector $\mathbf{e}$ and conditioning tokens $\mathbf{t}_n$, our cross-attention layer predicts the low-resolution feature map $\mathbf{E}^\mathrm{out}_n$ as: 
\begin{equation}
    \mathbf{E}^\mathrm{out}_n = \mathcal{F}_{\mathrm{CA}}(\mathbf{e}, \mathbf{t}_n).
\end{equation}

Given learnable parameters of cross-attention layer $\mathbf{w}_\mathrm{q}$, $\mathbf{w}_\mathrm{k}$, $\mathbf{w}_\mathrm{v}$, the above process can be divided into the sub-processes as:
\begin{equation}
    Q = \mathbf{e} \mathbf{w}_\mathrm{q}, ~~K_{n} = \mathbf{t}_{n} \mathbf{w}_\mathrm{k}, ~~V_{n} = \mathbf{t}_{n} \mathbf{w}_\mathrm{v},
\end{equation}
\begin{equation}
    A_{n} = \mathrm{softmax}(Q K^{\intercal}_{n}), ~~ \mathbf{E}^\mathrm{out}_{n} = \mathrm{A}_{n} V_{n},
\end{equation}
where $Q$, $K_{n}$, $V_{n}$ denote query, key and value representation, and $A_{n}$ represents attention scores. Each of the parameters represents MLP with 1 layer and 64 hidden dimensions.

\subsubsection{Super-resolution module.}
We replaced the original super-resolution module in EG3D~\cite{chan2022efficient} with GFPGAN~\cite{wang2021towards}, which enhances rendering quality by reducing noise or artifacts in the background. Following the training strategy documented in the main paper, we fine-tune the pre-trained GFPGAN for a few epochs. 
For a fair comparison, all quantitative evaluations were measured on the results obtained by using the original EG3D's super-resolution module.

\section{Additional Results and Comparisons}
\label{supsec:result_and_comparison}
\subsection{Novel-view synthesis and depth information}
\begin{figure*}[!p]
\newcolumntype{M}[1]{>{\centering\arraybackslash}m{#1}}
\setlength{\tabcolsep}{0.5pt}
\renewcommand{\arraystretch}{0.25}
\centering
\scriptsize

\resizebox{.95\linewidth}{!}{
\begin{tabular}{ M{0.1\linewidth} M{0.18\linewidth} M{0.18\linewidth} M{0.18\linewidth} M{0.18\linewidth} M{0.18\linewidth} }

$\mathbf{y}$, $\mathbf{p}$
& $\texttt{+}30^{\circ},\texttt{+}20^{\circ}$ & $\texttt{+}15^{\circ},\texttt{+}10^{\circ}$ & $\texttt{+}0^{\circ},\texttt{+}0^{\circ}$ & $\texttt{-}15^{\circ},\texttt{-}10^{\circ}$ & $\texttt{-}30^{\circ},\texttt{-}20^{\circ}$ \\ \\

AD-NeRF &
 \includegraphics[width=\linewidth]{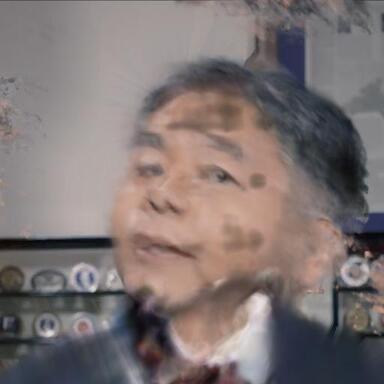}\hfill &
\includegraphics[width=\linewidth]{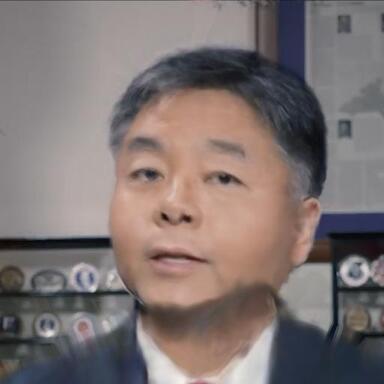}\hfill &
\includegraphics[width=\linewidth]{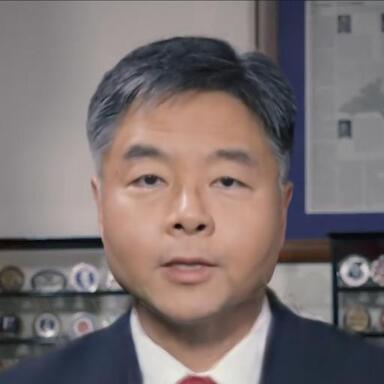}\hfill &
\includegraphics[width=\linewidth]{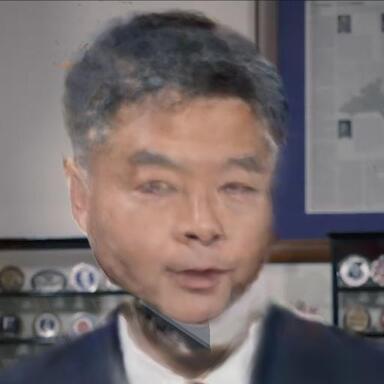}\hfill &
\includegraphics[width=\linewidth]{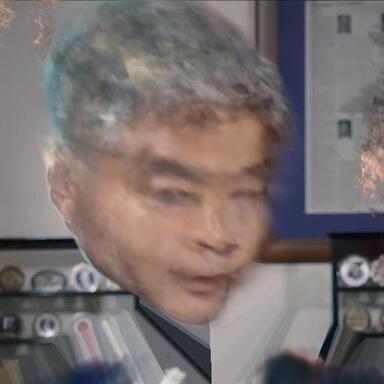}\hfill \\

&
 \includegraphics[width=\linewidth]{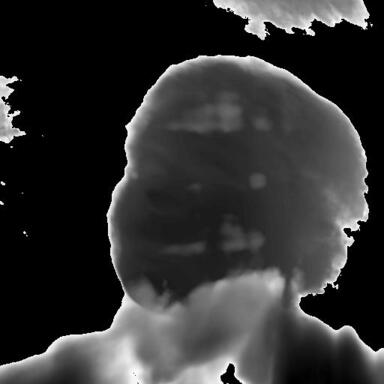}\hfill &
\includegraphics[width=\linewidth]{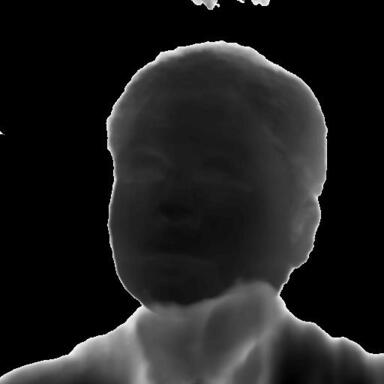}\hfill &
\includegraphics[width=\linewidth]{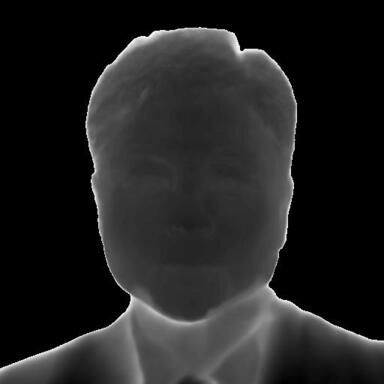}\hfill &
\includegraphics[width=\linewidth]{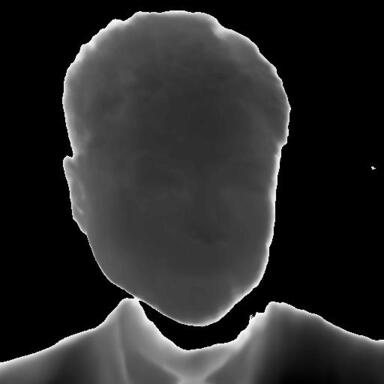}\hfill &
\includegraphics[width=\linewidth]{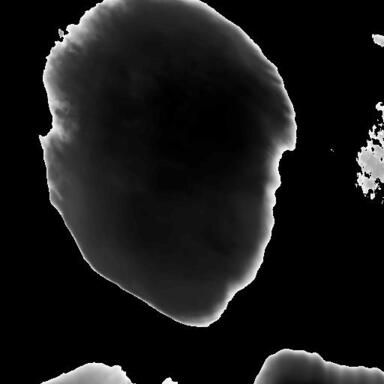}\hfill \\

RAD-NeRF &

 \includegraphics[width=\linewidth]{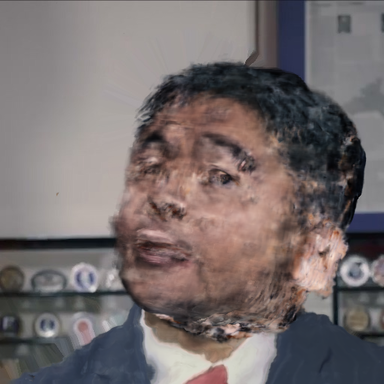}\hfill &
\includegraphics[width=\linewidth]{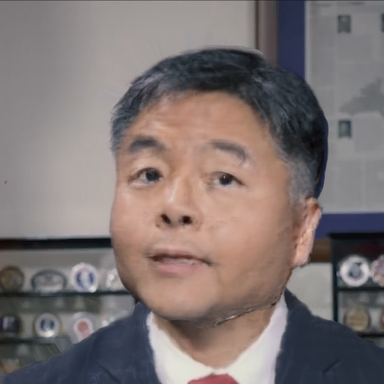}\hfill &
\includegraphics[width=\linewidth]{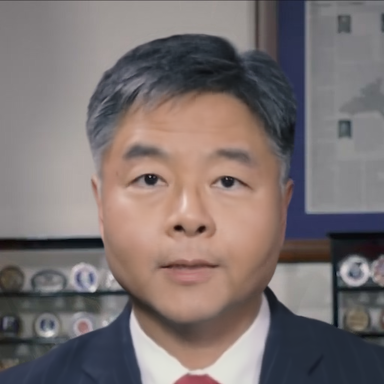}\hfill &
\includegraphics[width=\linewidth]{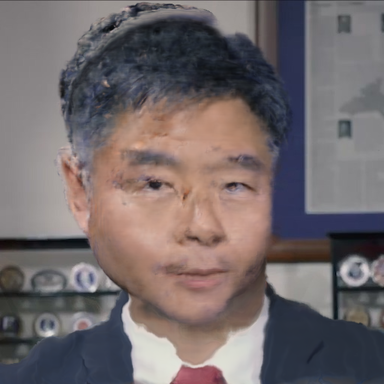}\hfill &
\includegraphics[width=\linewidth]{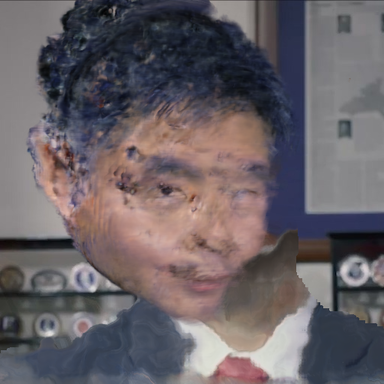}\hfill \\

&
 \includegraphics[width=\linewidth]{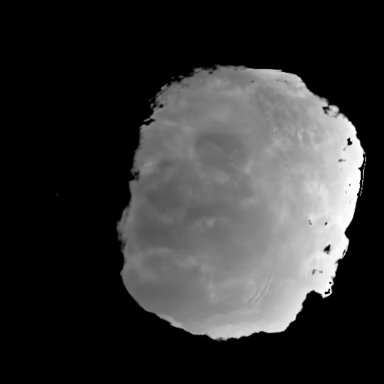}\hfill &
\includegraphics[width=\linewidth]{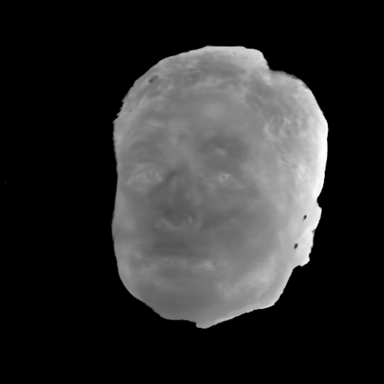}\hfill &
\includegraphics[width=\linewidth]{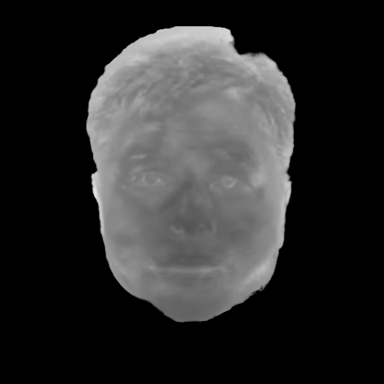}\hfill &
\includegraphics[width=\linewidth]{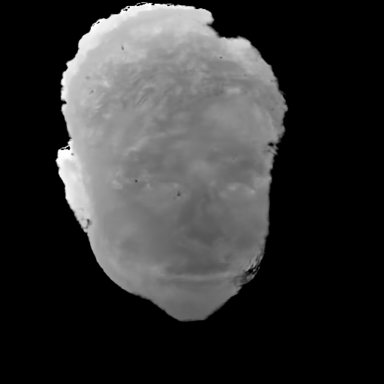}\hfill &
\includegraphics[width=\linewidth]{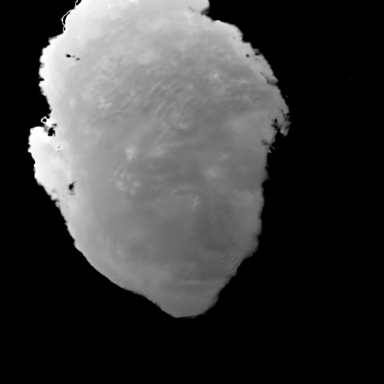}\hfill \\

ER- NeRF &
 \includegraphics[width=\linewidth]{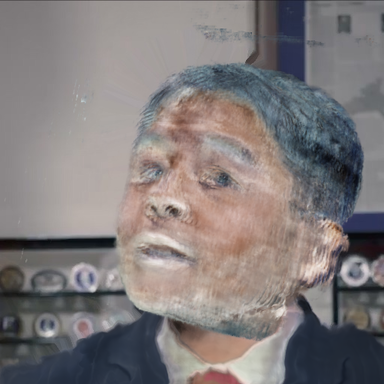}\hfill &
\includegraphics[width=\linewidth]{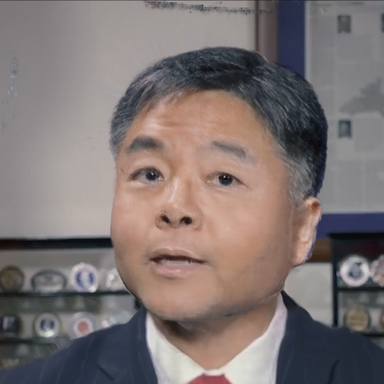}\hfill &
\includegraphics[width=\linewidth]{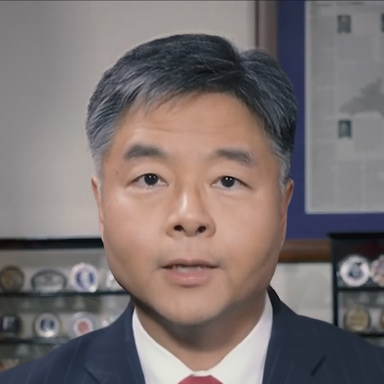}\hfill &
\includegraphics[width=\linewidth]{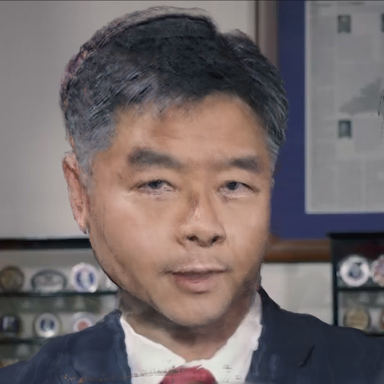}\hfill &
\includegraphics[width=\linewidth]{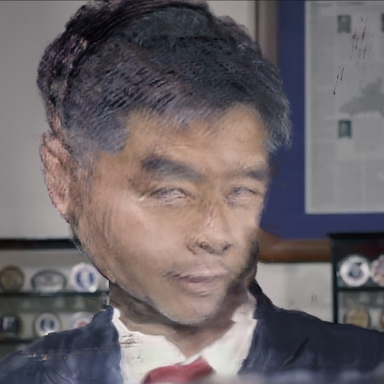}\hfill \\

&
 \includegraphics[width=\linewidth]{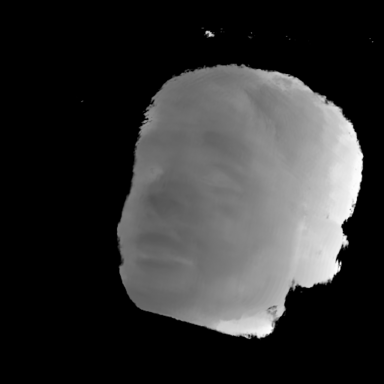}\hfill &
\includegraphics[width=\linewidth]{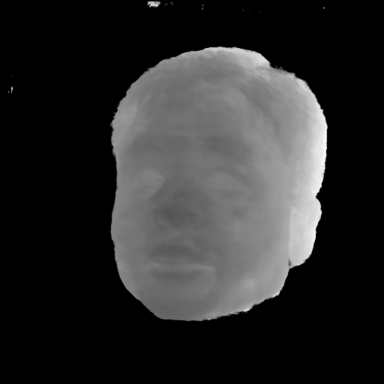}\hfill &
\includegraphics[width=\linewidth]{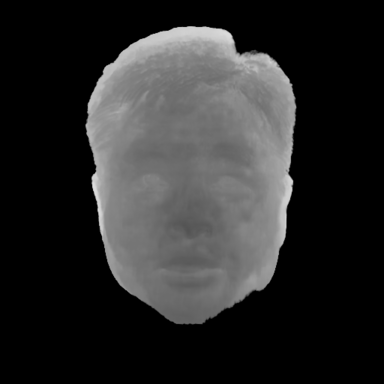}\hfill &
\includegraphics[width=\linewidth]{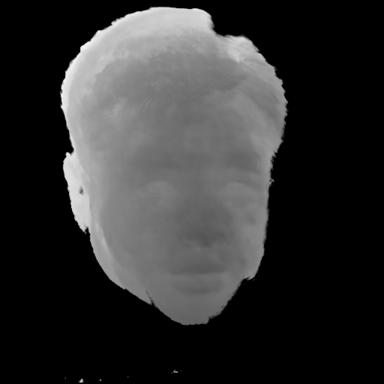}\hfill &
\includegraphics[width=\linewidth]{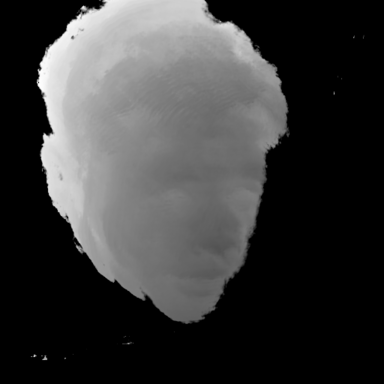}\hfill \\

\textbf{Talk3D (Ours)} &
 \includegraphics[width=\linewidth]{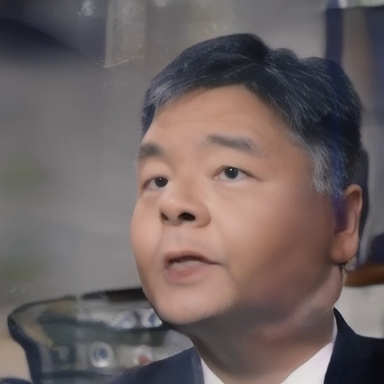}\hfill &
\includegraphics[width=\linewidth]{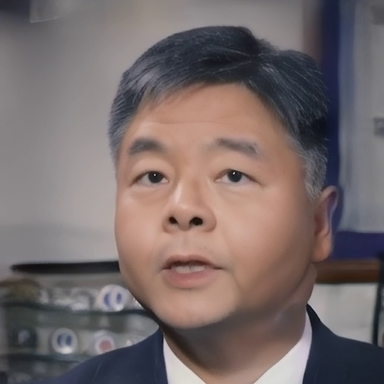}\hfill &
\includegraphics[width=\linewidth]{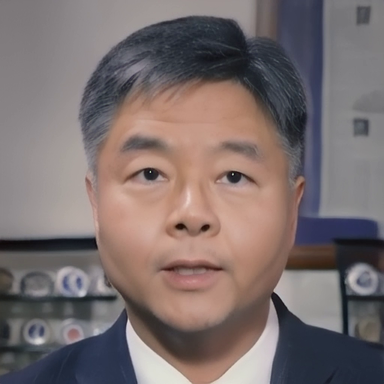}\hfill &
\includegraphics[width=\linewidth]{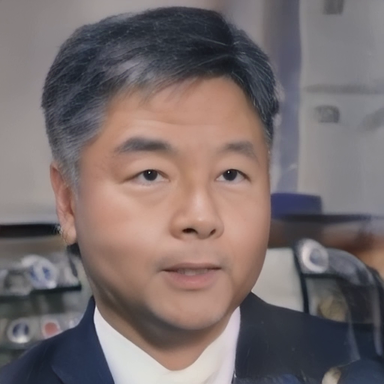}\hfill &
\includegraphics[width=\linewidth]{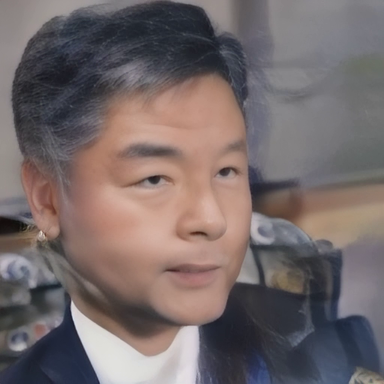}\hfill \\

&
 \includegraphics[width=\linewidth]{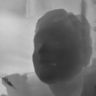}\hfill &
\includegraphics[width=\linewidth]{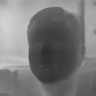}\hfill &
\includegraphics[width=\linewidth]{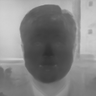}\hfill &
\includegraphics[width=\linewidth]{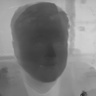}\hfill &
\includegraphics[width=\linewidth]{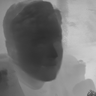}\hfill \\

\end{tabular}}
\vspace{-5pt}
\caption{\textbf{Visualization of synthesized portraits and depth map rendered from novel viewpoints.} We show a randomly selected frame from synthesized talking portraits (odd rows) and corresponding depth information (even rows) using different rendering viewpoints of yaw and pitch angles ($\mathbf{y}$, $\mathbf{p}$) with $15^{\circ}$, $10^{\circ}$ intervals. }
\label{supfig:ood_pose_depth}
\vspace{-13pt}
\end{figure*}

We demonstrate the robustness of Talk3D by generating images from the extreme viewpoints, shown in \figref{supfig:ood_pose_depth}. We compare our method with the previous NeRF-based methods~\cite{guo2021adnerf, tang2022radnerf, li2023ernerf} visualizing both the generated images and their corresponding depth maps.  
Note that, the other NeRF-based methods do not synthesize the background and therefore lack depth information in that particular area. Furthermore, they frequently show the head and torso separation due to their separative volumetric representation for torso rendering. Especially, RAD-NeRF~\cite{tang2022radnerf} and ER-NeRF~\cite{li2023ernerf} employ a 2D deformation neural field for torso rendering, thus they are not capable of generating a realistic torso geometry. In contrast, our model successfully constructs the entire image as a single NeRF representation, providing depth information for all parts of the synthesized portrait.

\subsection{Additional dataset and qualitative results}
\begin{figure*}[!p]
\newcolumntype{M}[1]{>{\centering\arraybackslash}m{#1}}
\setlength{\tabcolsep}{0.2pt}
\renewcommand{\arraystretch}{0.12}
\centering
\scriptsize

\resizebox{\linewidth}{!}{
\begin{tabular}{M{0.09\linewidth}@{\hskip 0.005\linewidth}  M{0.12\linewidth}M{0.12\linewidth}M{0.12\linewidth}  M{0.12\linewidth} @{\hskip 0.01\linewidth}  M{0.12\linewidth} M{0.12\linewidth}M{0.12\linewidth}M{0.12\linewidth}}

& \textcolor{red}{I} & \textcolor{red}{do} & \textcolor{red}{con}tinue & str\textcolor{red}{ong}er & \textcolor{red}{wo}rld & f\textcolor{red}{or}ms & wi\textcolor{red}{ll n}ot & \textcolor{red}{ho}wever \\ \\

\makecell{\vspace{-12pt}\\ Ground \\ Truth} &
\includegraphics[width=\linewidth]{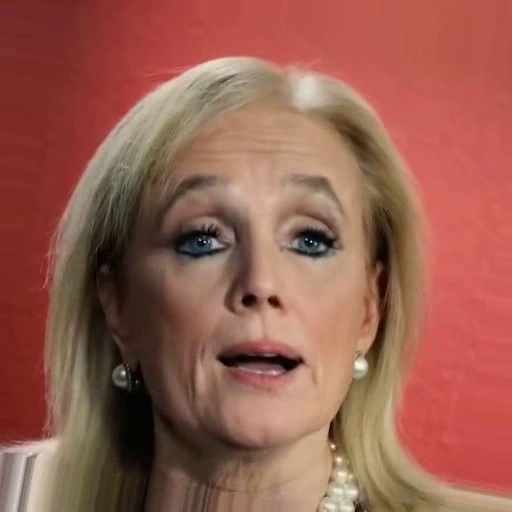}\hfill &
\includegraphics[width=\linewidth]{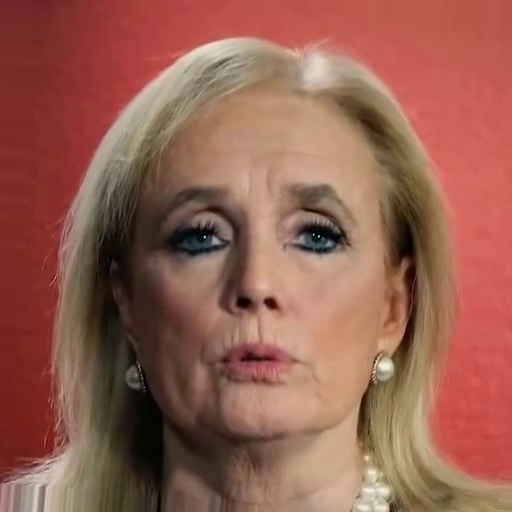}\hfill &
\includegraphics[width=\linewidth]{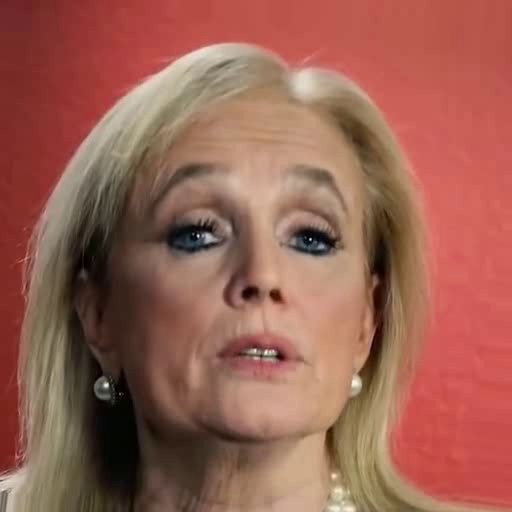}\hfill &
\includegraphics[width=\linewidth]{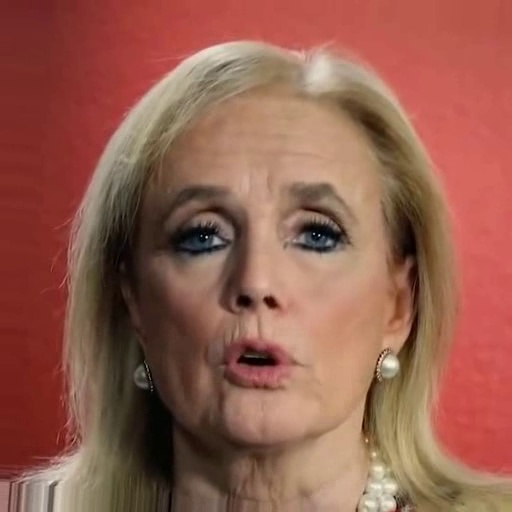}\hfill &
\includegraphics[width=\linewidth]{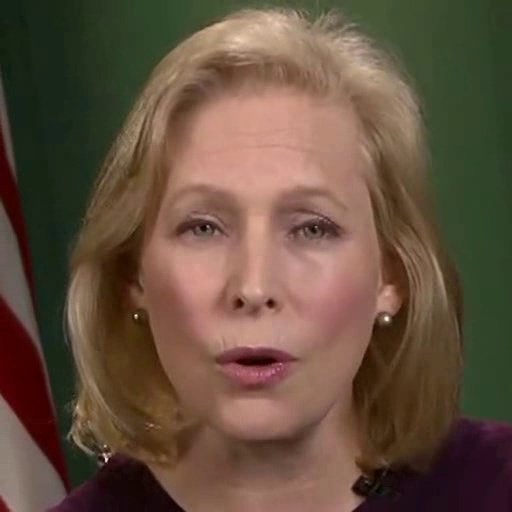}\hfill &
\includegraphics[width=\linewidth]{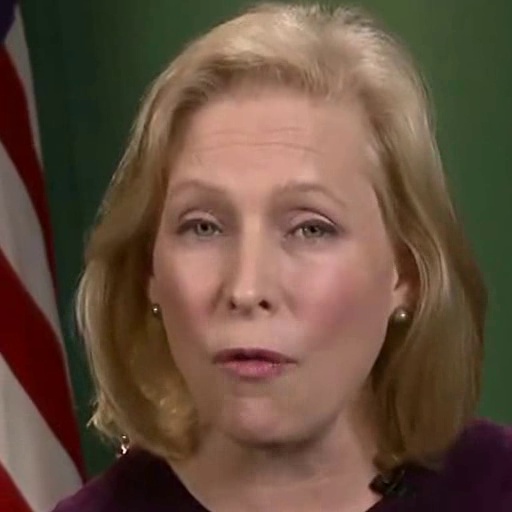}\hfill &
\includegraphics[width=\linewidth]{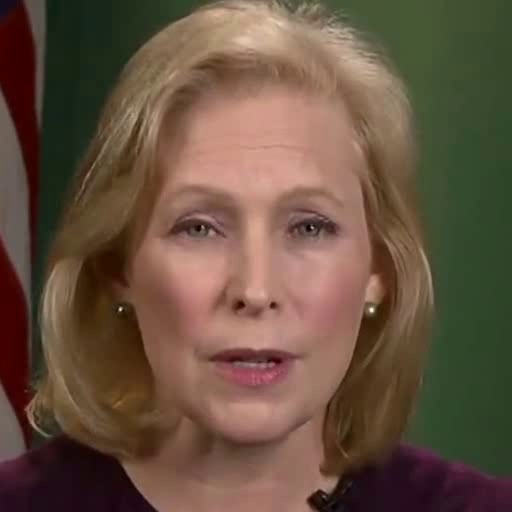}\hfill &
\includegraphics[width=\linewidth]{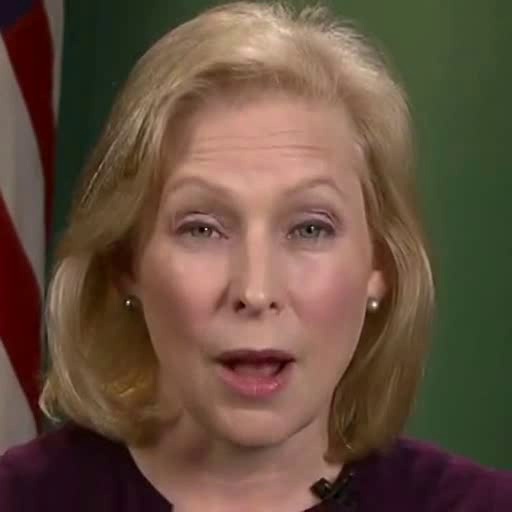}\hfill \\

Wav2Lip &
\includegraphics[width=\linewidth]{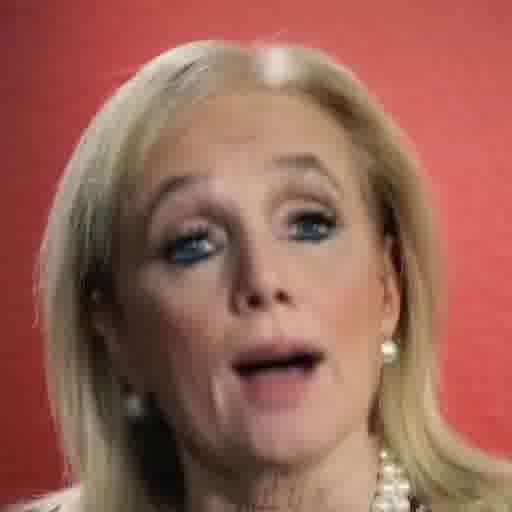}\hfill &
\includegraphics[width=\linewidth]{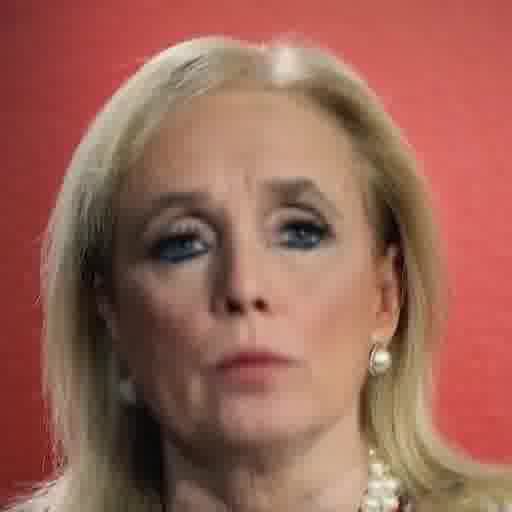}\hfill &
\includegraphics[width=\linewidth]{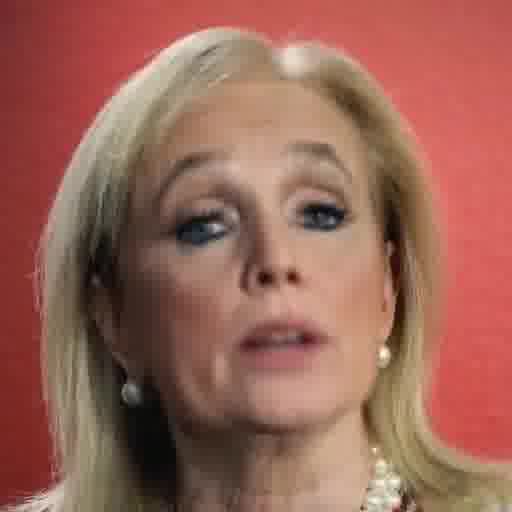}\hfill &
\includegraphics[width=\linewidth]{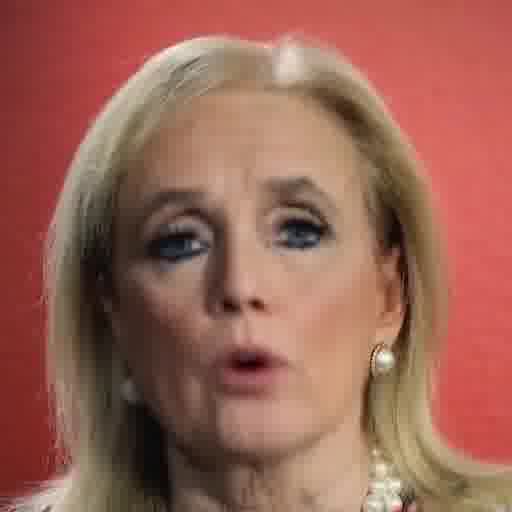}\hfill &
\includegraphics[width=\linewidth]{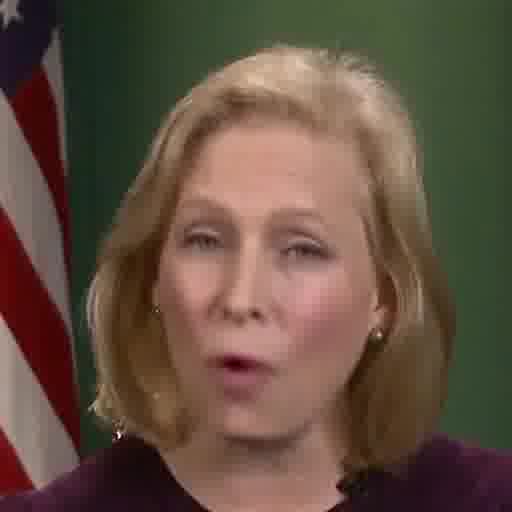}\hfill &
\includegraphics[width=\linewidth]{supplementary/fig2/w2l_kir_186.jpg} &
\includegraphics[width=\linewidth]{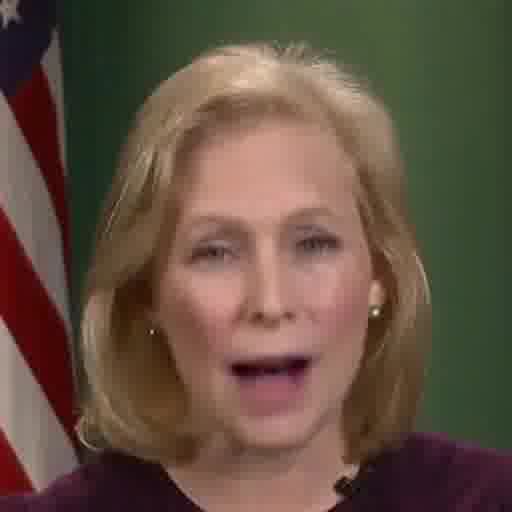}\hfill &
\includegraphics[width=\linewidth]{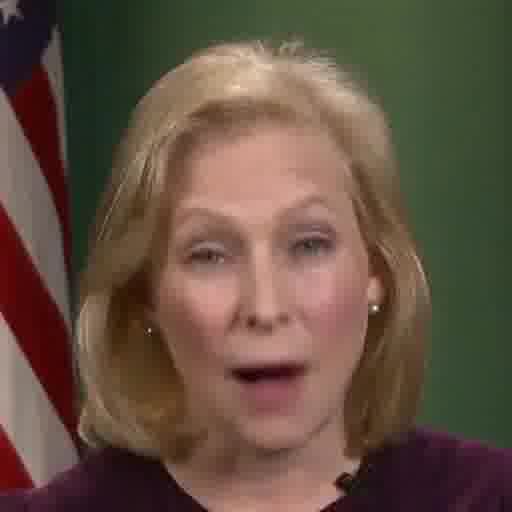}\hfill \\

PC-AVS &
\includegraphics[width=\linewidth]{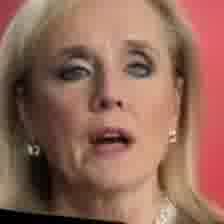}\hfill &
\includegraphics[width=\linewidth]{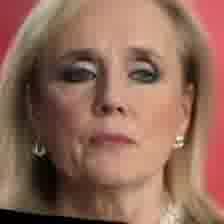}\hfill &
\includegraphics[width=\linewidth]{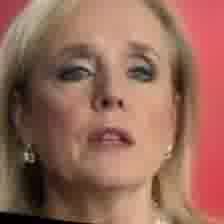}\hfill &
\includegraphics[width=\linewidth]{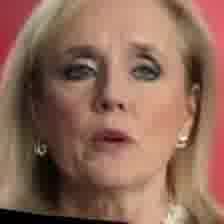}\hfill &
\includegraphics[width=\linewidth]{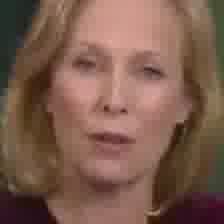}\hfill &
\includegraphics[width=\linewidth]{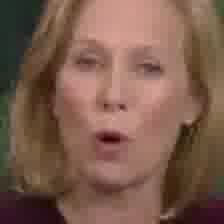}\hfill &
\includegraphics[width=\linewidth]{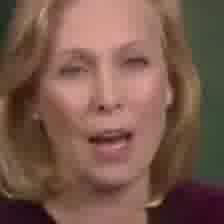}\hfill &
\includegraphics[width=\linewidth]{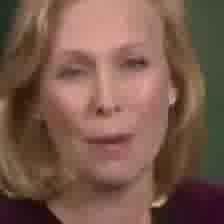}\hfill \\

AD-NeRF &
\includegraphics[width=\linewidth]{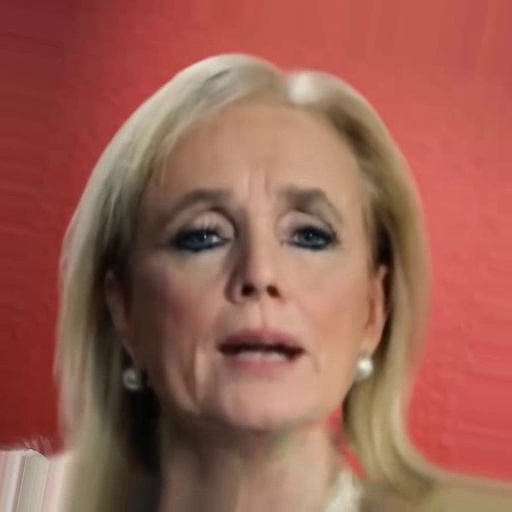}\hfill &
\includegraphics[width=\linewidth]{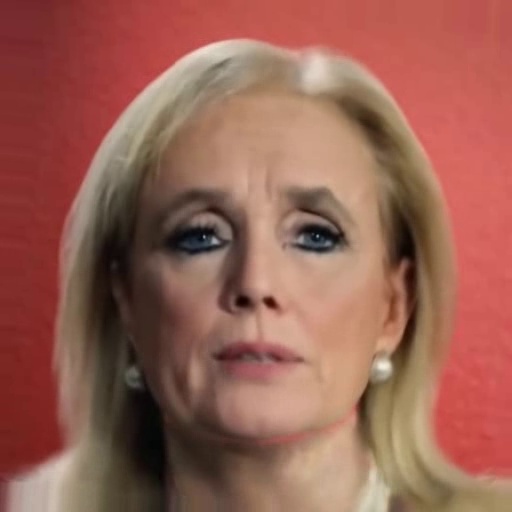}\hfill &
\includegraphics[width=\linewidth]{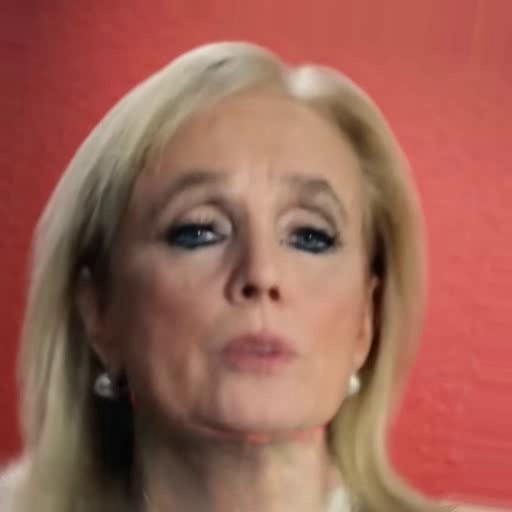}\hfill &
\includegraphics[width=\linewidth]{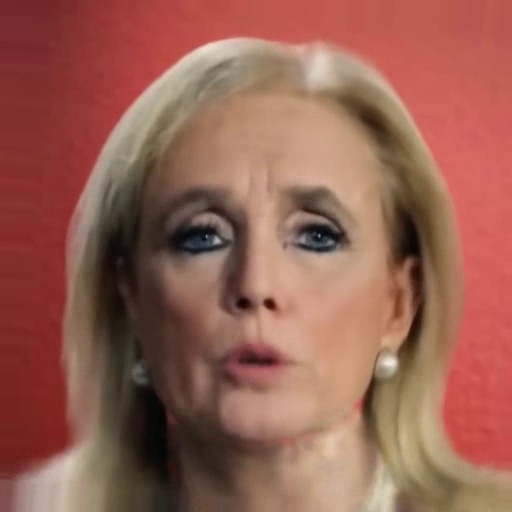}\hfill &
\includegraphics[width=\linewidth]{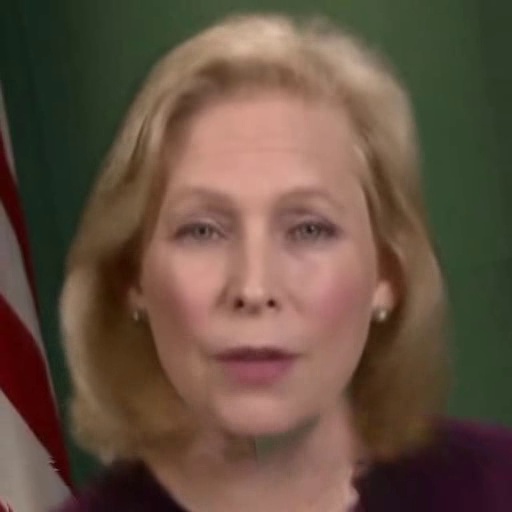}\hfill &
\includegraphics[width=\linewidth]{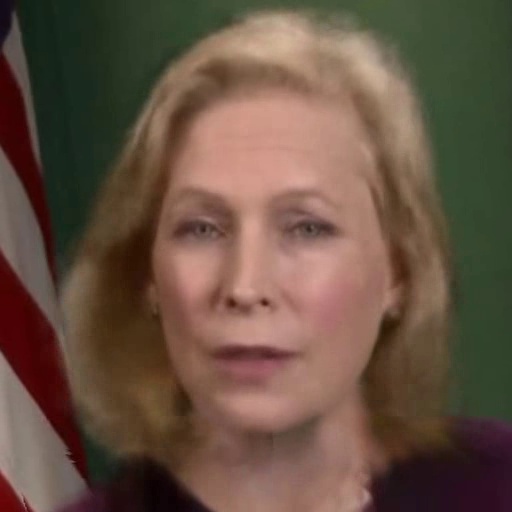}\hfill &
\includegraphics[width=\linewidth]{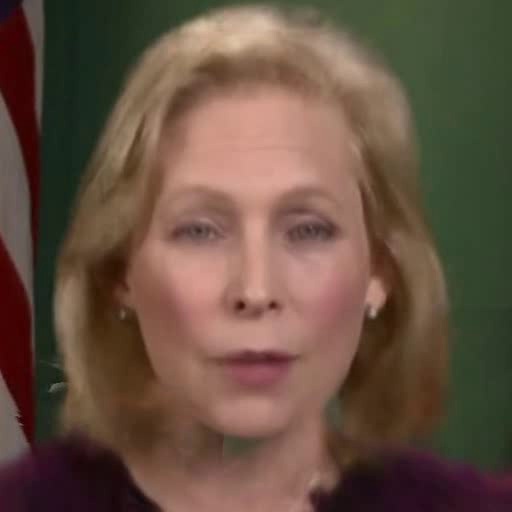}\hfill &
\includegraphics[width=\linewidth]{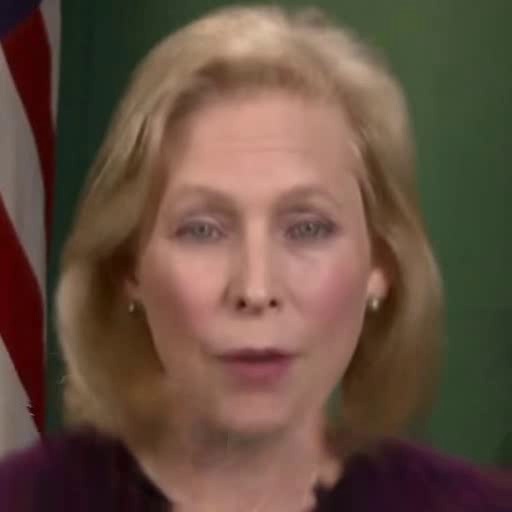}\hfill \\

RAD-NeRF &
\includegraphics[width=\linewidth]{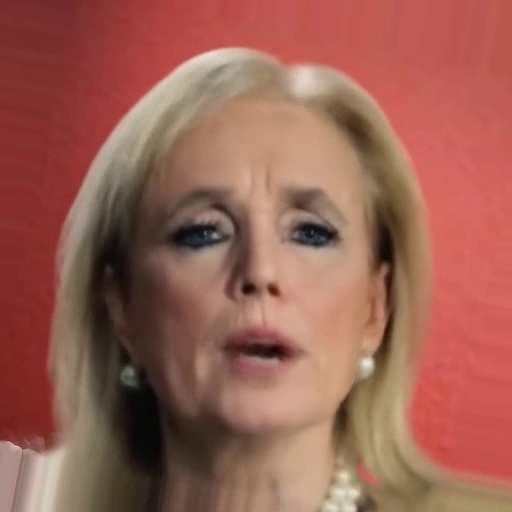}\hfill &
\includegraphics[width=\linewidth]{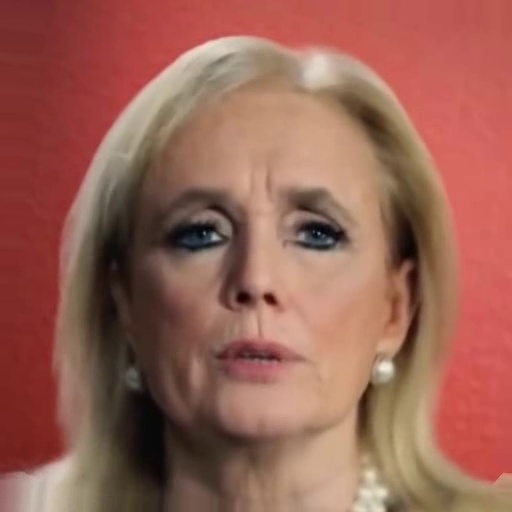}\hfill &
\includegraphics[width=\linewidth]{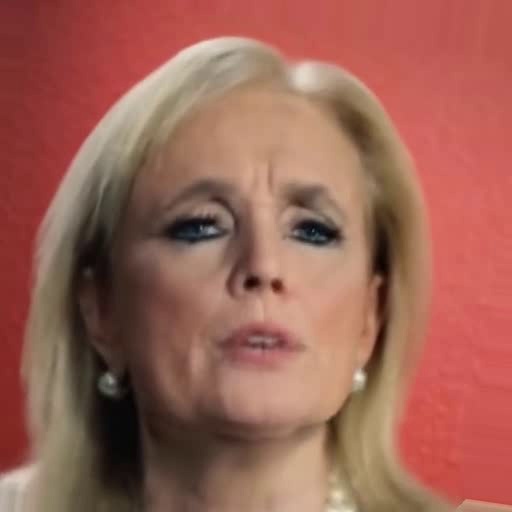}\hfill &
\includegraphics[width=\linewidth]{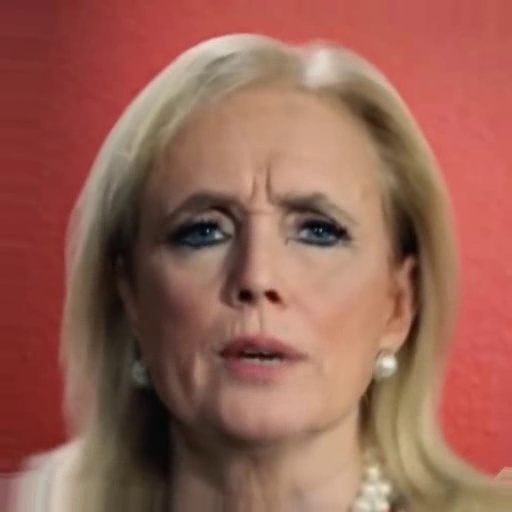}\hfill &
\includegraphics[width=\linewidth]{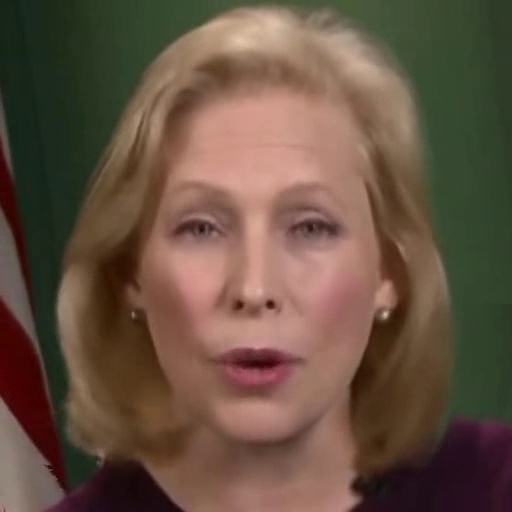}\hfill &
\includegraphics[width=\linewidth]{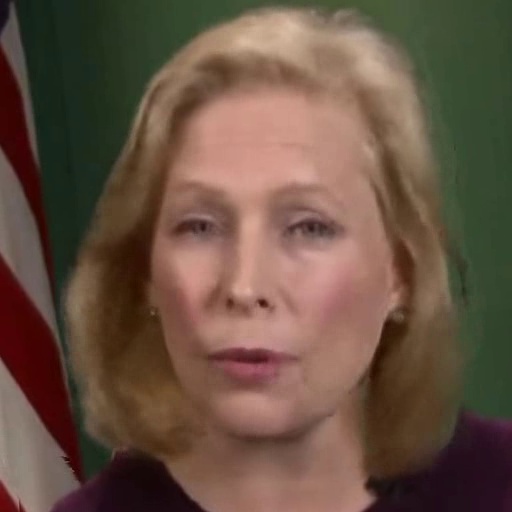}\hfill &
\includegraphics[width=\linewidth]{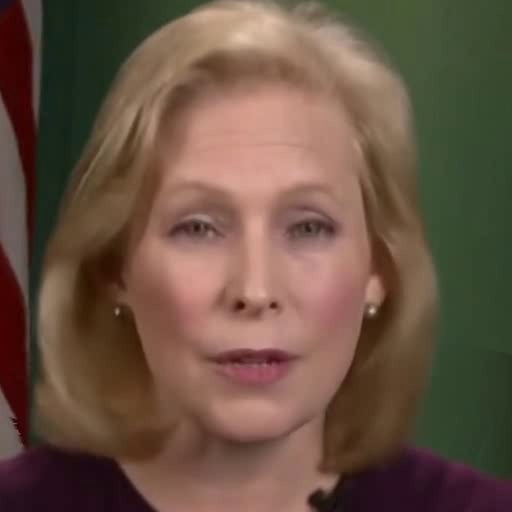}\hfill &
\includegraphics[width=\linewidth]{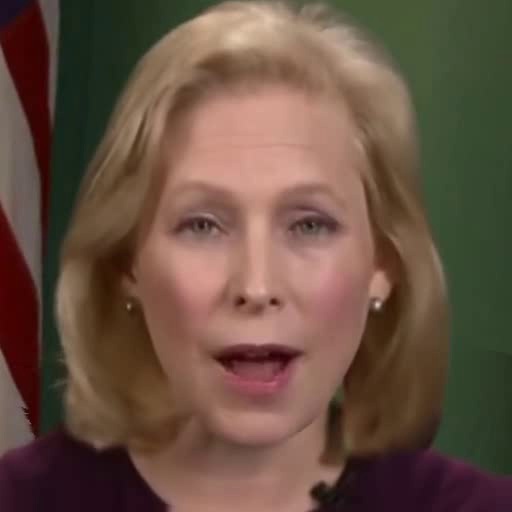}\hfill \\

ER-NeRF &
\includegraphics[width=\linewidth]{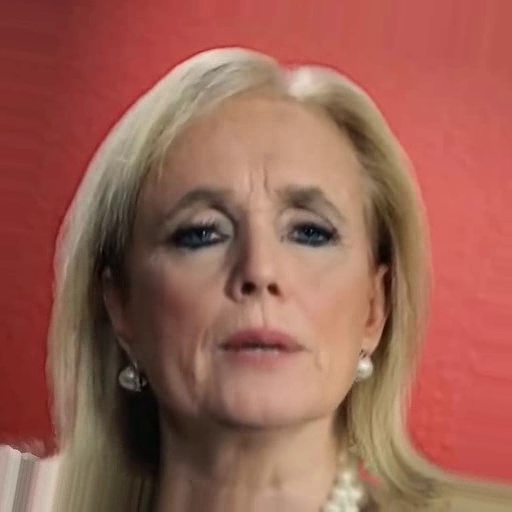}\hfill &
\includegraphics[width=\linewidth]{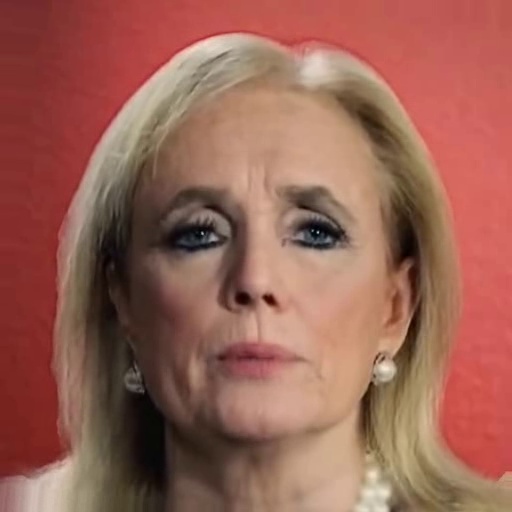}\hfill &
\includegraphics[width=\linewidth]{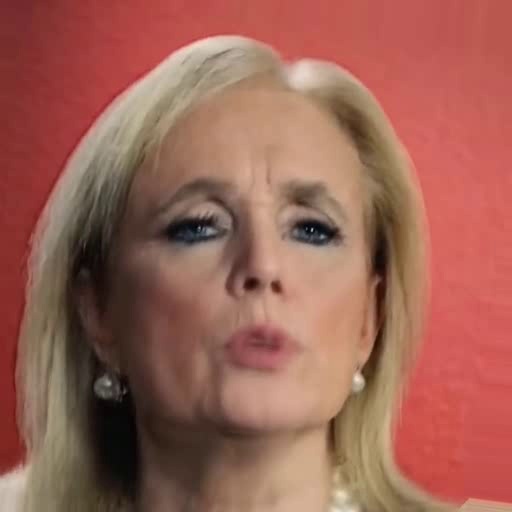}\hfill &
\includegraphics[width=\linewidth]{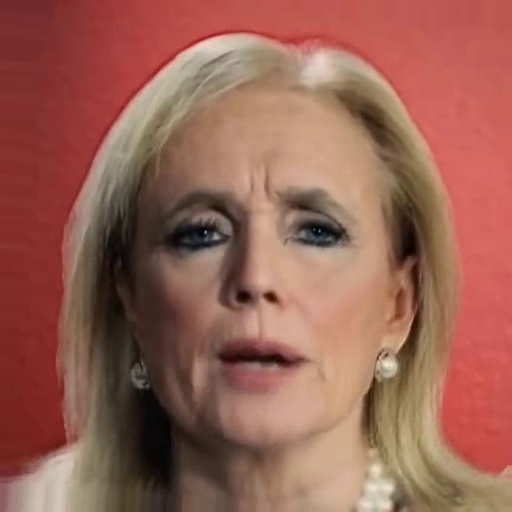}\hfill &
\includegraphics[width=\linewidth]{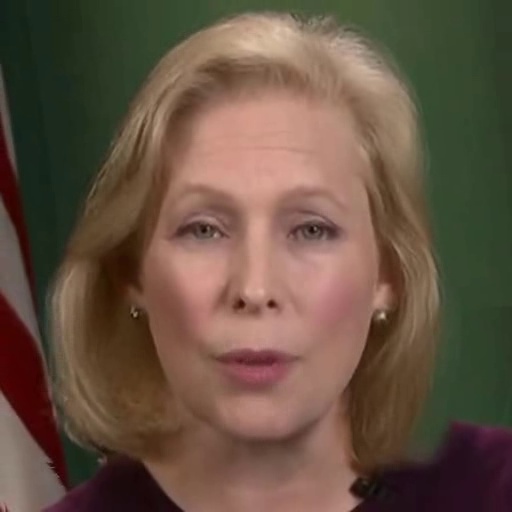}\hfill &
\includegraphics[width=\linewidth]{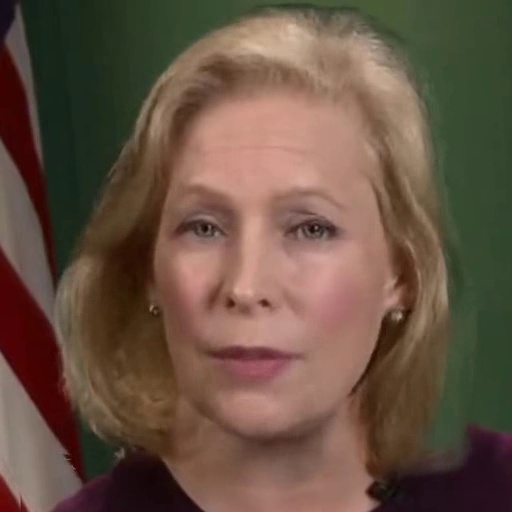}\hfill &
\includegraphics[width=\linewidth]{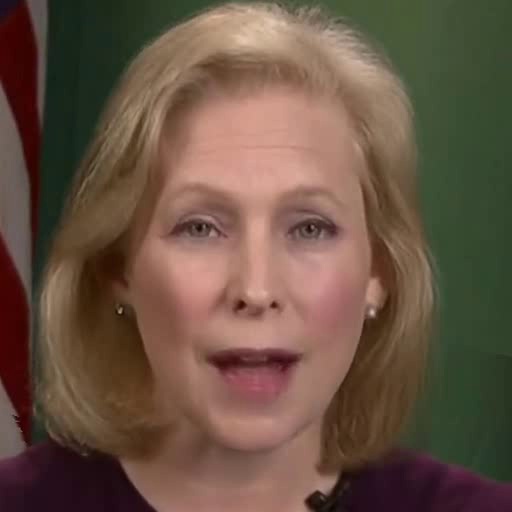}\hfill &
\includegraphics[width=\linewidth]{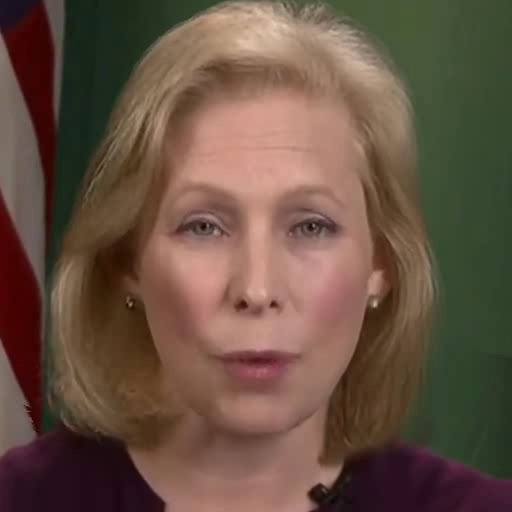}\hfill \\

\textbf{Talk3D (Ours)} &
\includegraphics[width=\linewidth]{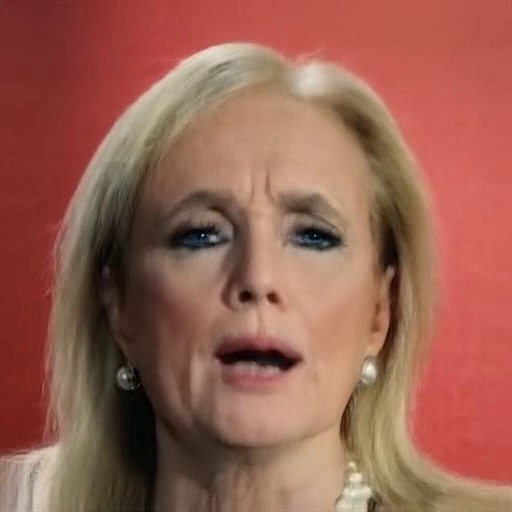}\hfill &
\includegraphics[width=\linewidth]{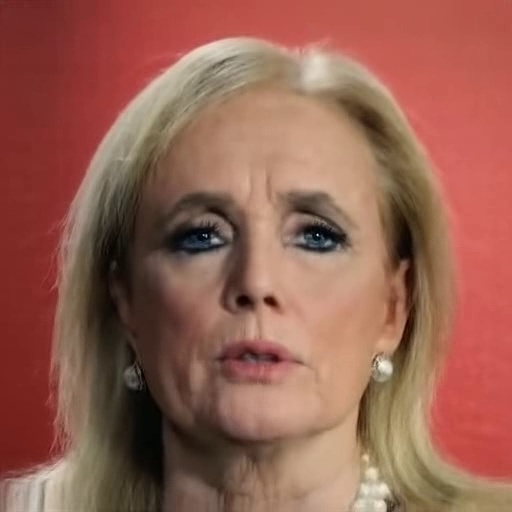}\hfill &
\includegraphics[width=\linewidth]{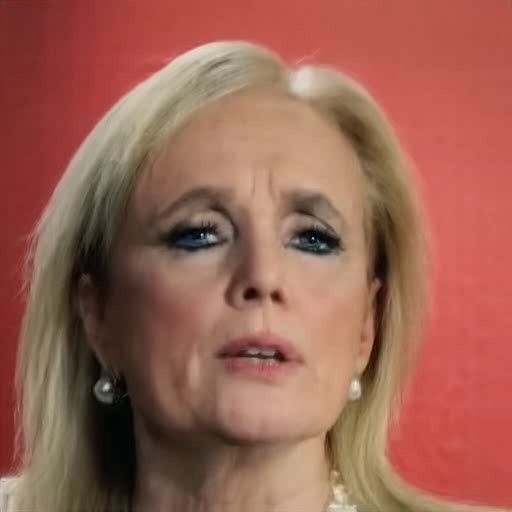}\hfill &
\includegraphics[width=\linewidth]{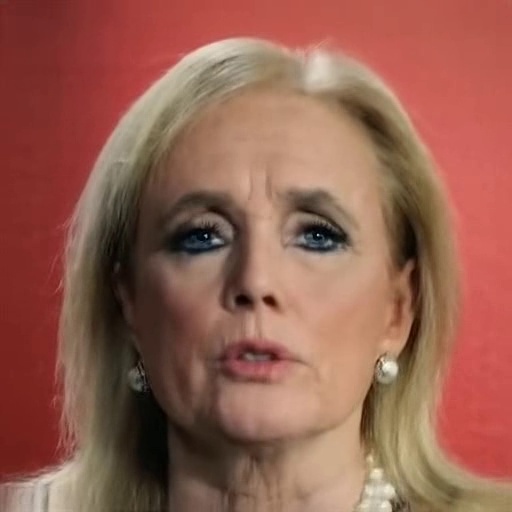}\hfill &
\includegraphics[width=\linewidth]{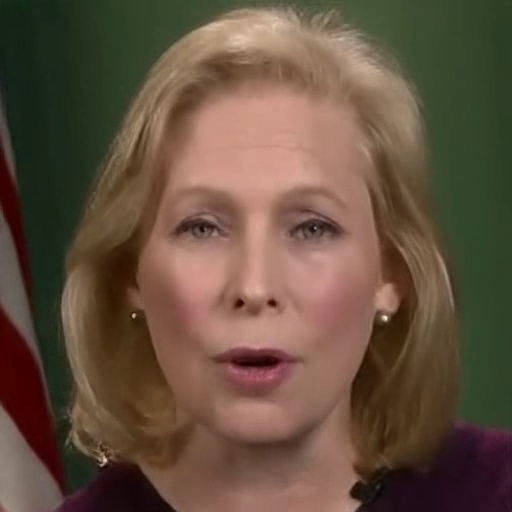}\hfill &
\includegraphics[width=\linewidth]{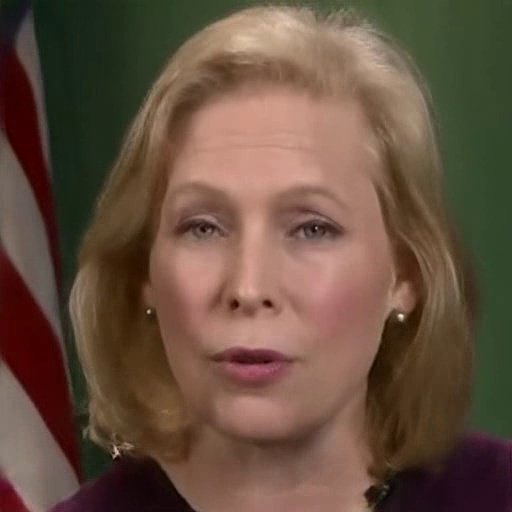}\hfill &
\includegraphics[width=\linewidth]{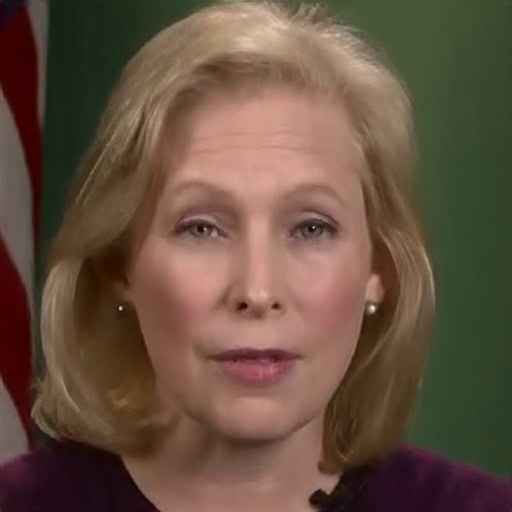}\hfill &
\includegraphics[width=\linewidth]{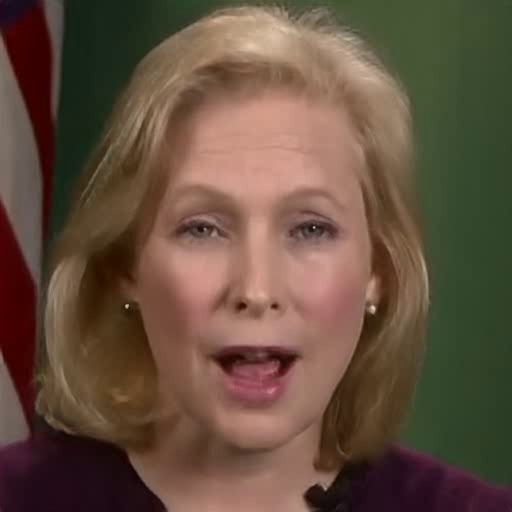}\hfill \\

\end{tabular}}
\vspace{-5pt}
\caption{\textbf{The \emph{self-driven} comparison of the key frames and details of generated portraits.} We show visualizations of our method and related methods under the self-driven setting. Best viewed in zoom.}
\label{supfig:self_driven}
\vspace{-10pt}
\end{figure*}

\begin{figure*}[!p]
\newcolumntype{M}[1]{>{\centering\arraybackslash}m{#1}}
\setlength{\tabcolsep}{0.2pt}
\renewcommand{\arraystretch}{0.1}
\centering
\scriptsize

\resizebox{\linewidth}{!}{
\begin{tabular}{M{0.09\linewidth}@{\hskip 0.005\linewidth}  M{0.12\linewidth}M{0.12\linewidth}M{0.12\linewidth}  M{0.12\linewidth} @{\hskip 0.01\linewidth}  M{0.12\linewidth} M{0.12\linewidth}M{0.12\linewidth}M{0.12\linewidth}}

& \textcolor{red}{he}lp & refi\textcolor{red}{na}nce & -fac\textcolor{red}{ture} & \vspace{2pt}\textcolor{red}{ea}rn & for\textcolor{red}{get} & m\textcolor{red}{i}ne & \textcolor{red}{shi}ne & $\langle$ mute $\rangle$ \\ \\

\makecell{\vspace{-12pt}\\ Ground \\ Truth} &
\includegraphics[width=\linewidth]{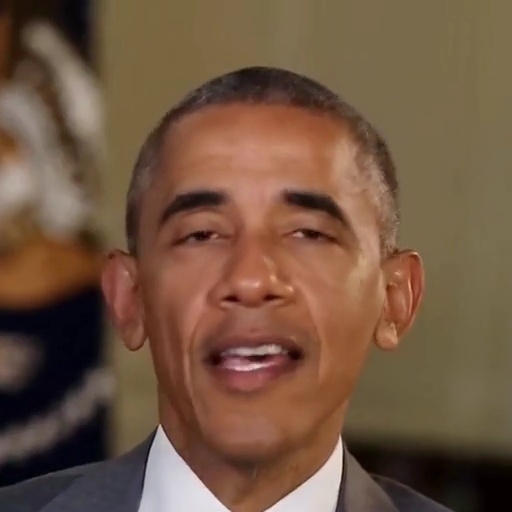}\hfill &
\includegraphics[width=\linewidth]{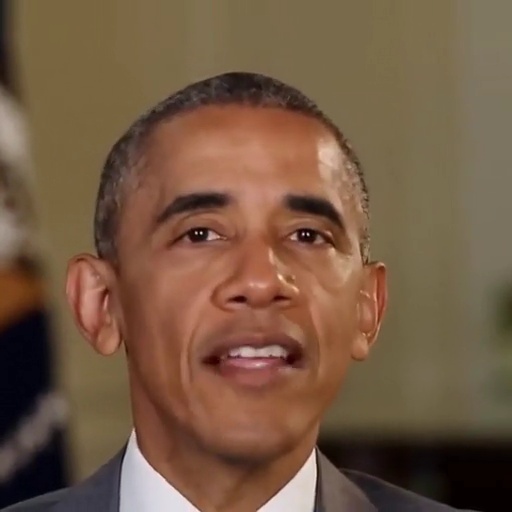}\hfill &
\includegraphics[width=\linewidth]{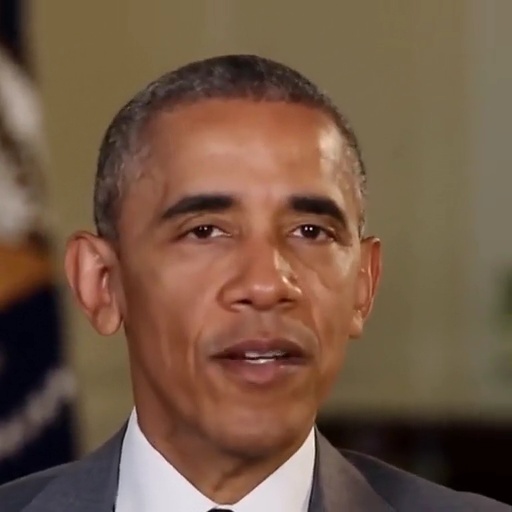}\hfill &
\includegraphics[width=\linewidth]{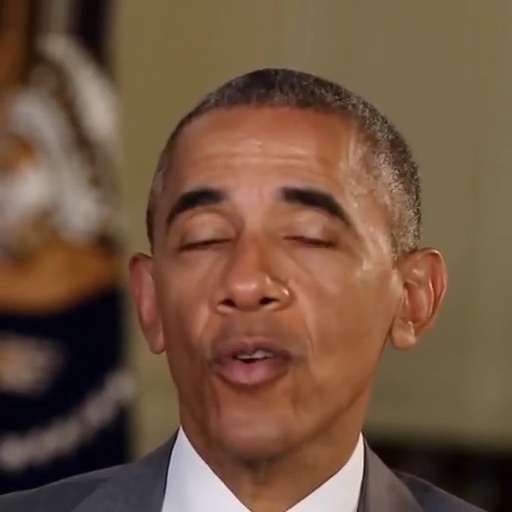}\hfill &
\includegraphics[width=\linewidth]{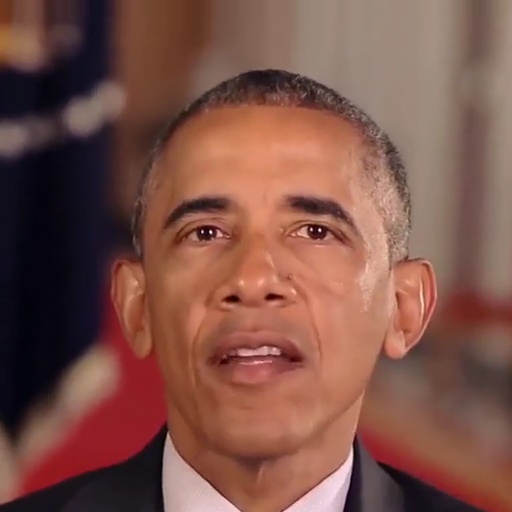}\hfill &
\includegraphics[width=\linewidth]{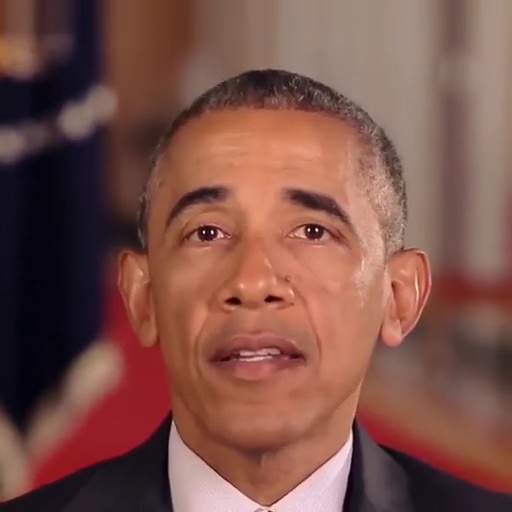}\hfill &
\includegraphics[width=\linewidth]{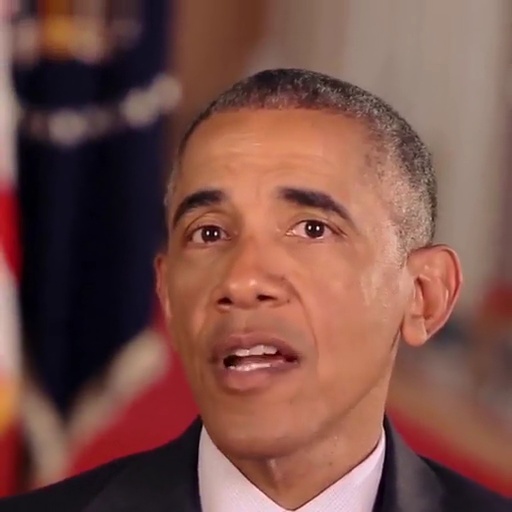}\hfill &
\includegraphics[width=\linewidth]{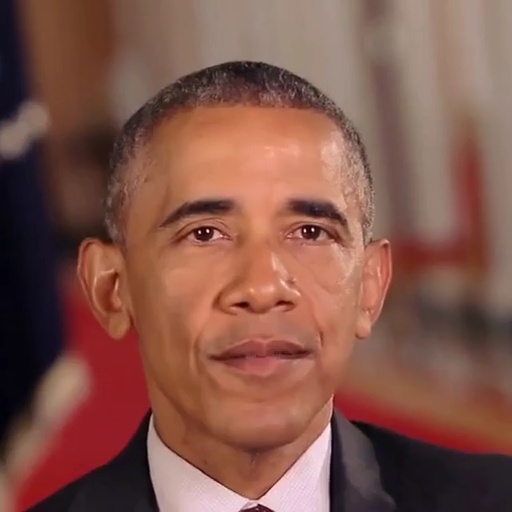}\hfill \\

Wav2Lip &
\includegraphics[width=\linewidth]{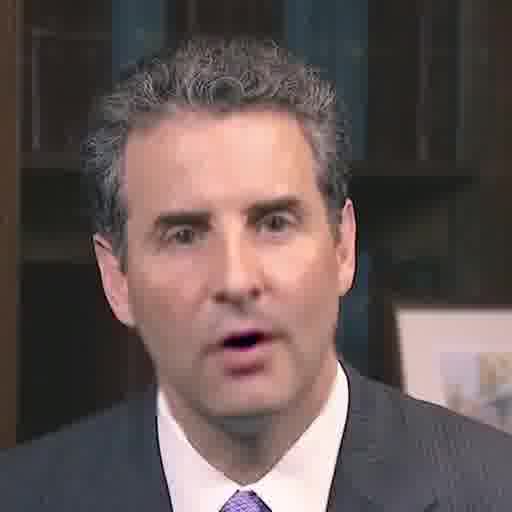}\hfill &
\includegraphics[width=\linewidth]{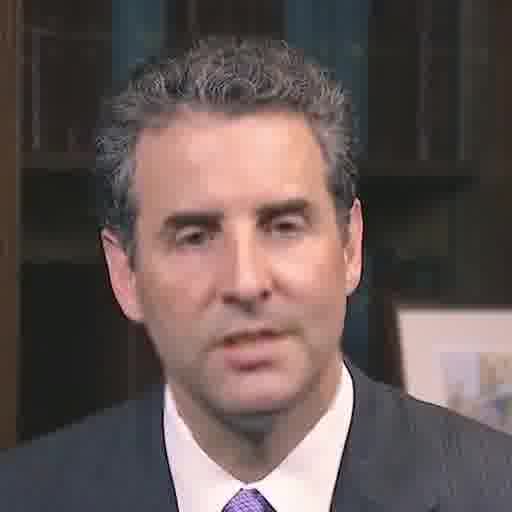}\hfill &
\includegraphics[width=\linewidth]{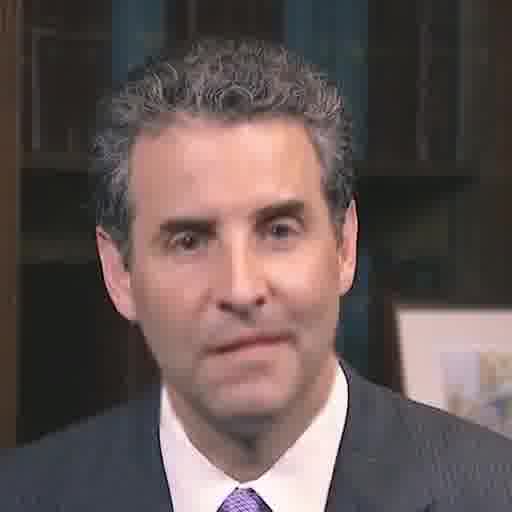}\hfill &
\includegraphics[width=\linewidth]{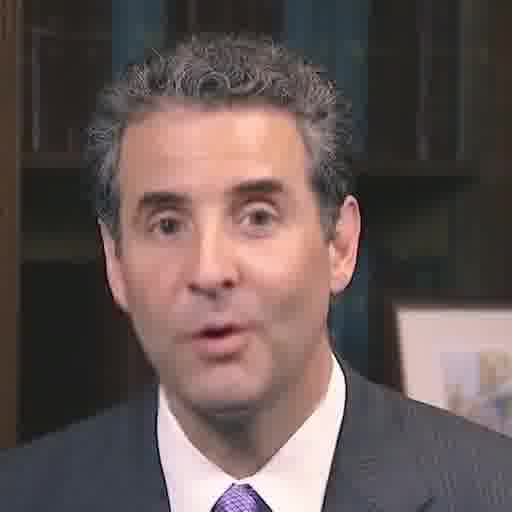}\hfill &
\includegraphics[width=\linewidth]{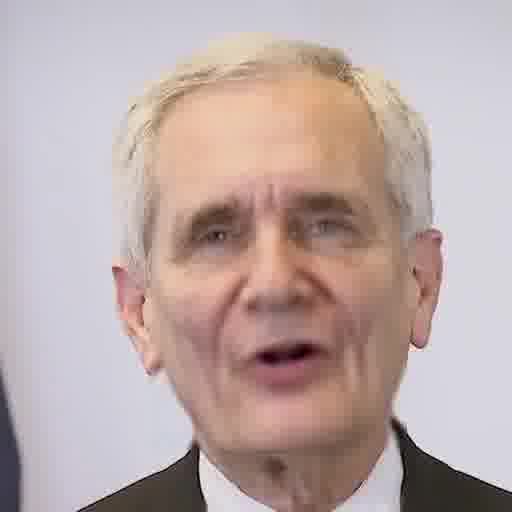}\hfill &
\includegraphics[width=\linewidth]{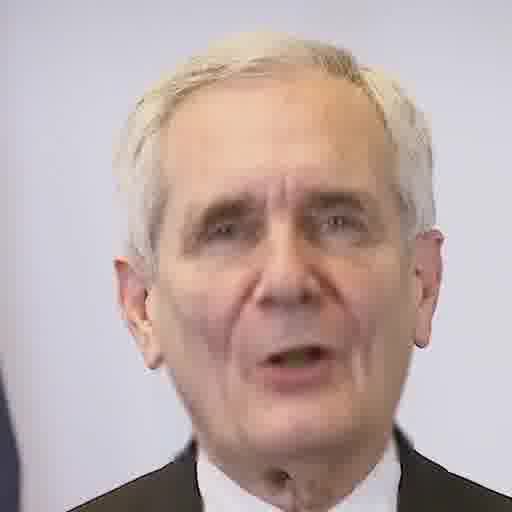} &
\includegraphics[width=\linewidth]{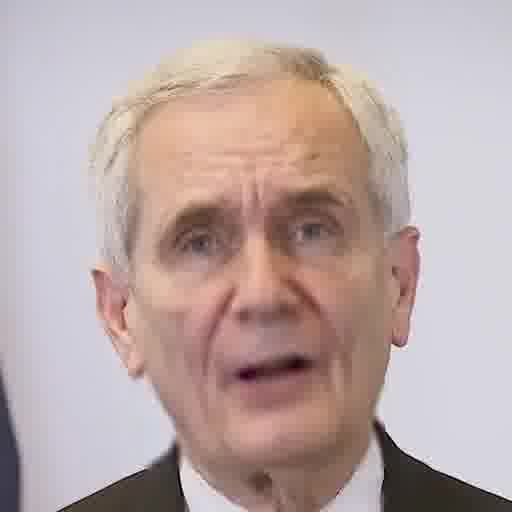}\hfill &
\includegraphics[width=\linewidth]{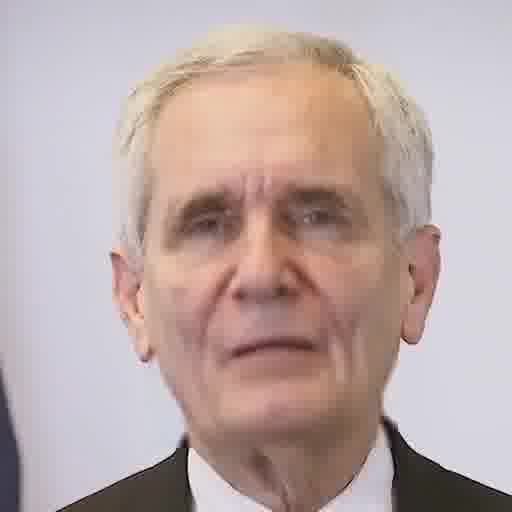}\hfill \\

PC-AVS &
\includegraphics[width=\linewidth]{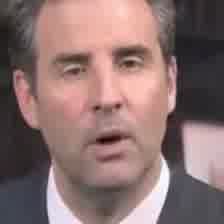}\hfill &
\includegraphics[width=\linewidth]{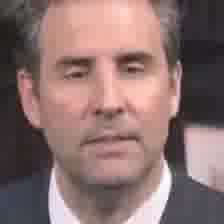}\hfill &
\includegraphics[width=\linewidth]{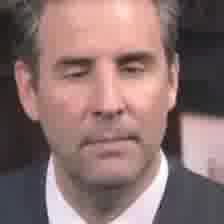}\hfill &
\includegraphics[width=\linewidth]{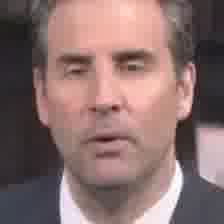}\hfill &
\includegraphics[width=\linewidth]{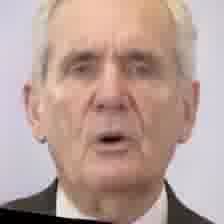}\hfill &
\includegraphics[width=\linewidth]{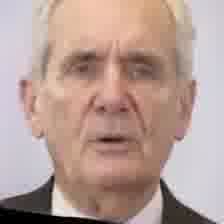}\hfill &
\includegraphics[width=\linewidth]{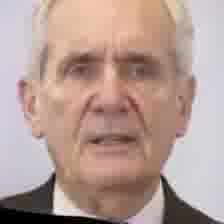}\hfill &
\includegraphics[width=\linewidth]{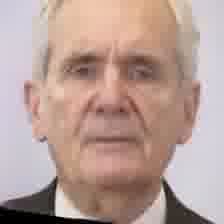}\hfill \\

AD-NeRF &
\includegraphics[width=\linewidth]{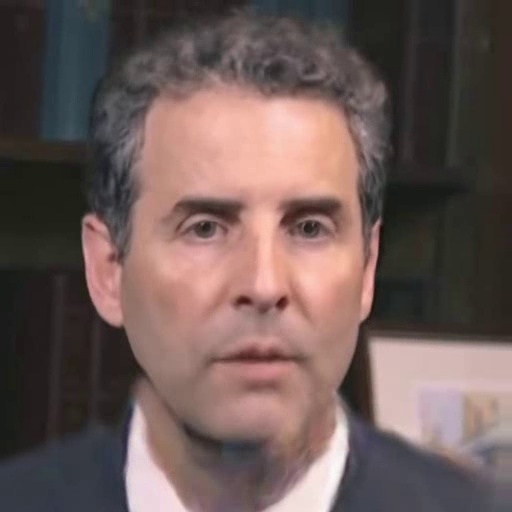}\hfill &
\includegraphics[width=\linewidth]{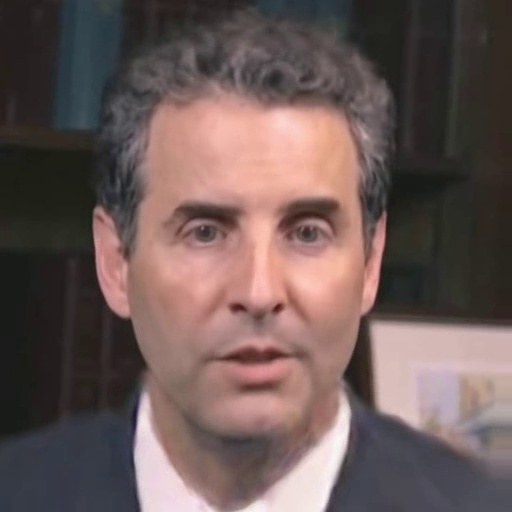}\hfill &
\includegraphics[width=\linewidth]{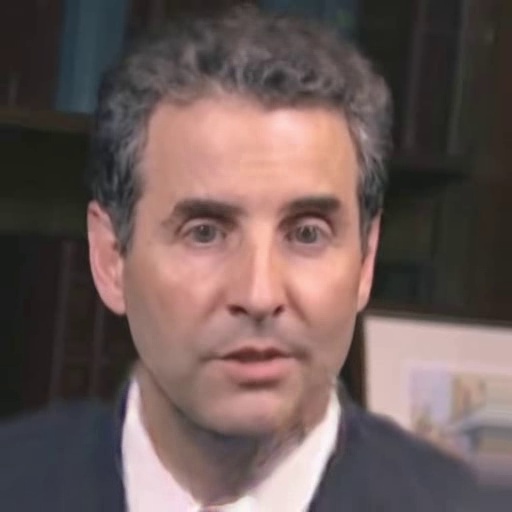}\hfill &
\includegraphics[width=\linewidth]{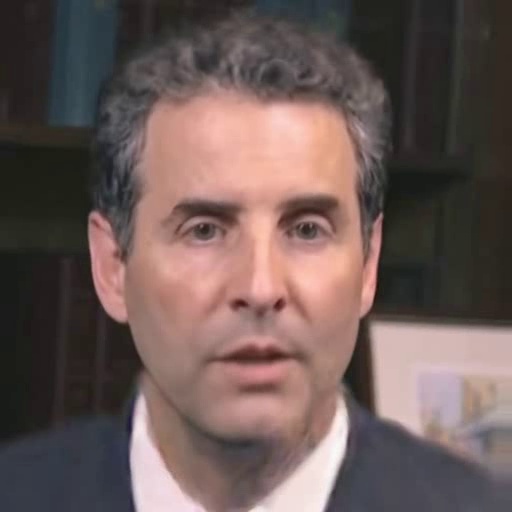}\hfill &
\includegraphics[width=\linewidth]{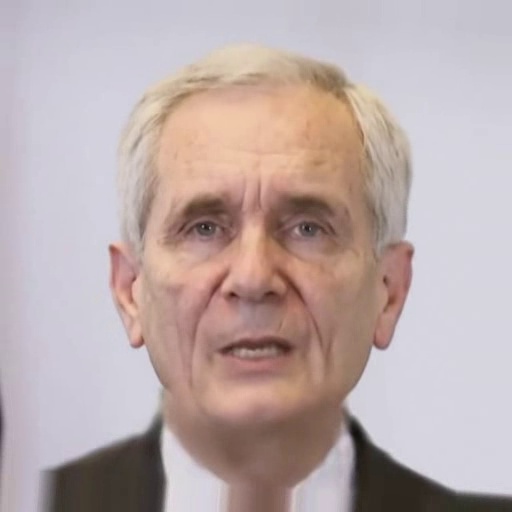}\hfill &
\includegraphics[width=\linewidth]{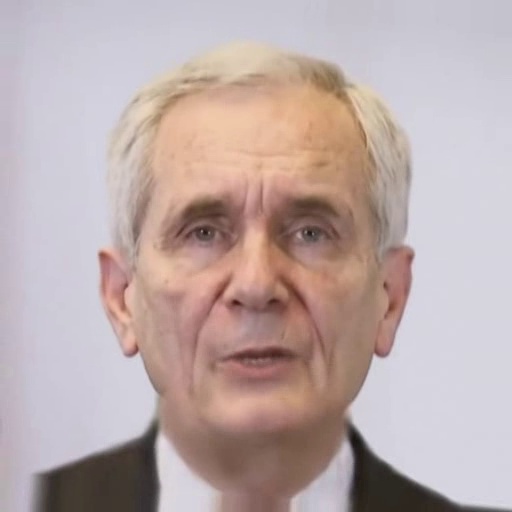}\hfill &
\includegraphics[width=\linewidth]{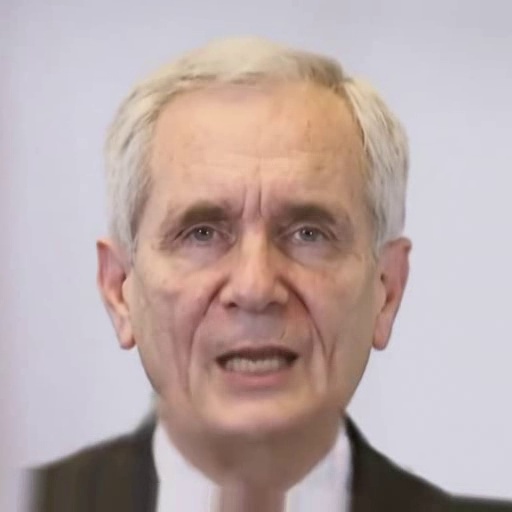}\hfill &
\includegraphics[width=\linewidth]{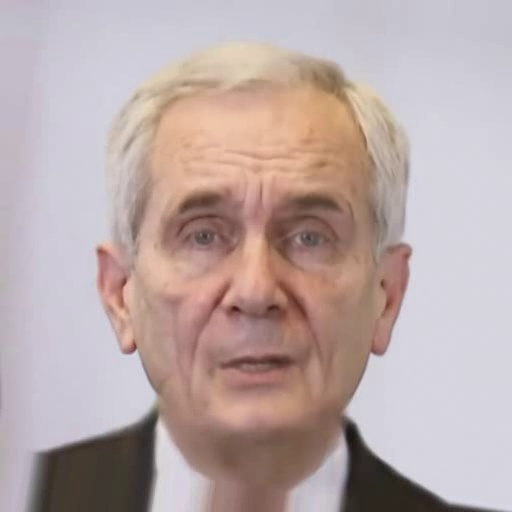}\hfill \\

RAD-NeRF &
\includegraphics[width=\linewidth]{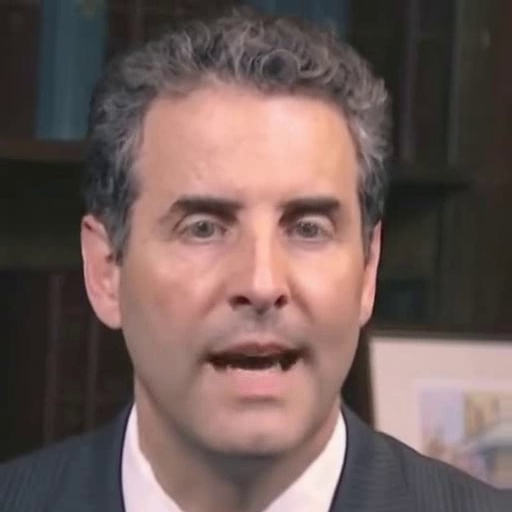}\hfill &
\includegraphics[width=\linewidth]{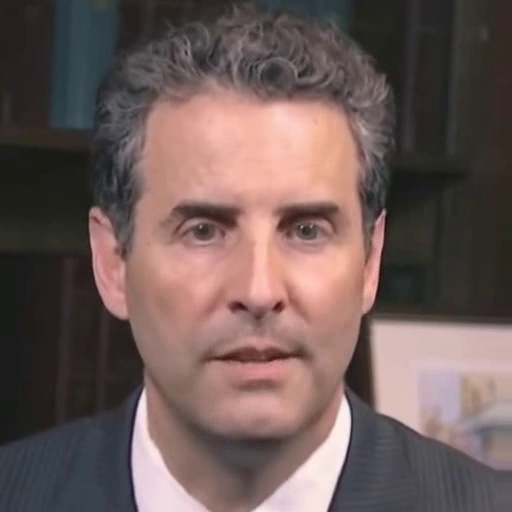}\hfill &
\includegraphics[width=\linewidth]{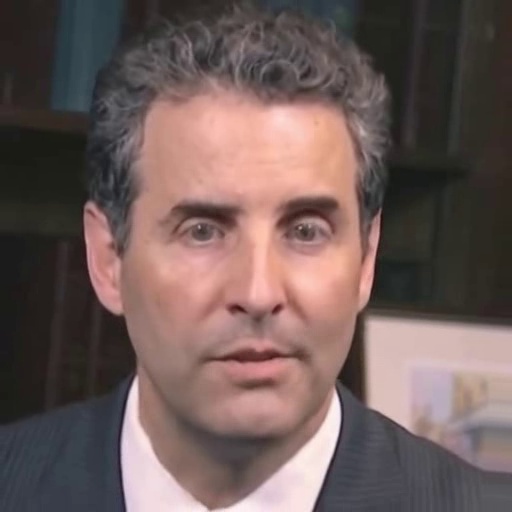}\hfill &
\includegraphics[width=\linewidth]{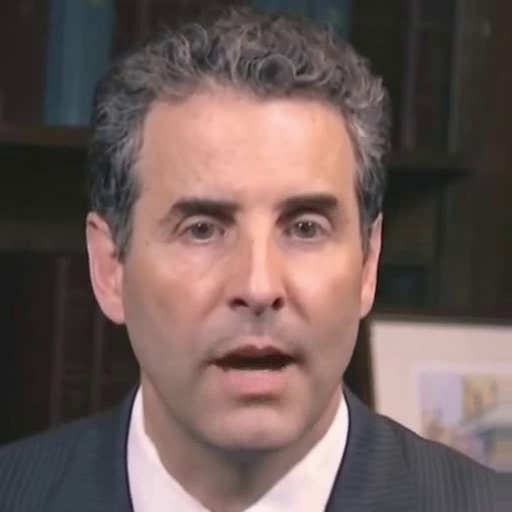}\hfill &
\includegraphics[width=\linewidth]{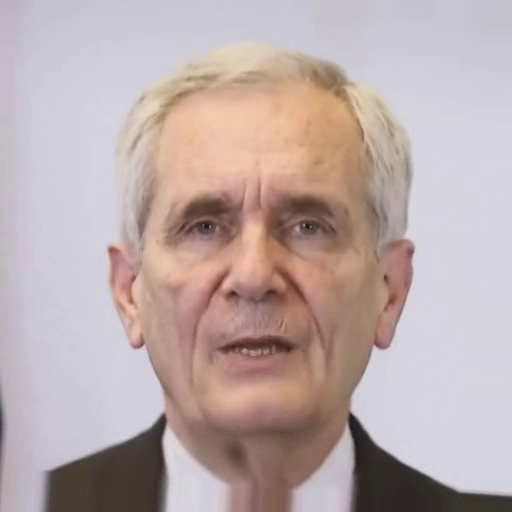}\hfill &
\includegraphics[width=\linewidth]{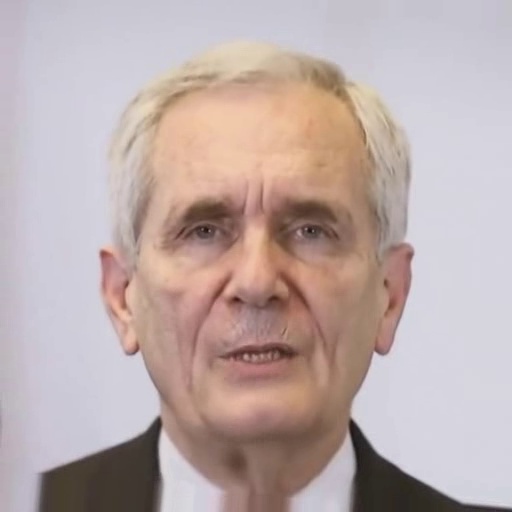}\hfill &
\includegraphics[width=\linewidth]{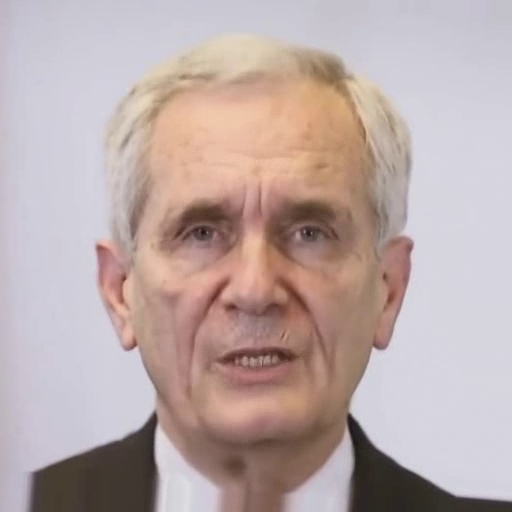}\hfill &
\includegraphics[width=\linewidth]{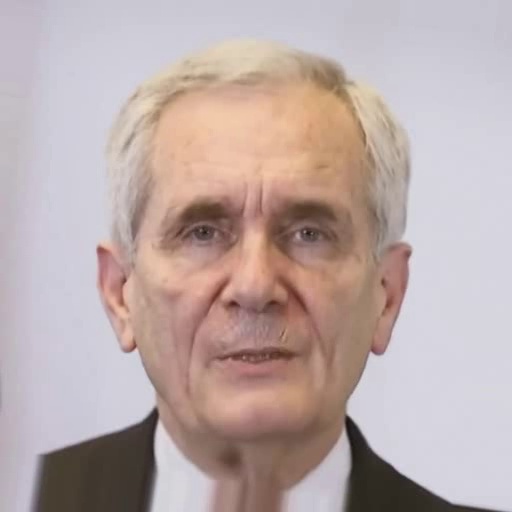}\hfill \\

ER-NeRF &
\includegraphics[width=\linewidth]{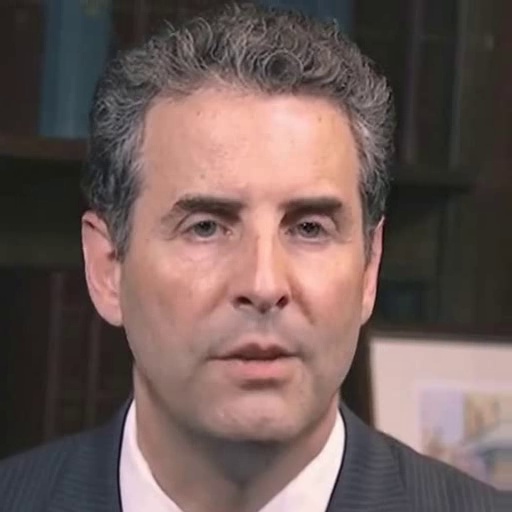}\hfill &
\includegraphics[width=\linewidth]{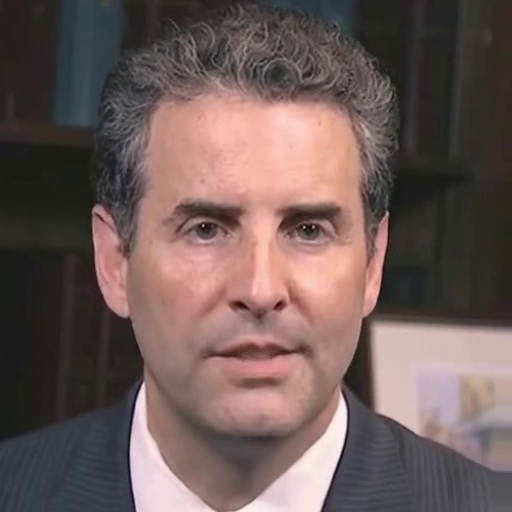}\hfill &
\includegraphics[width=\linewidth]{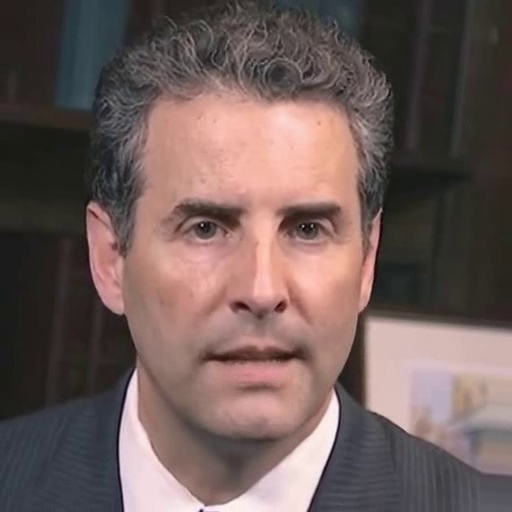}\hfill &
\includegraphics[width=\linewidth]{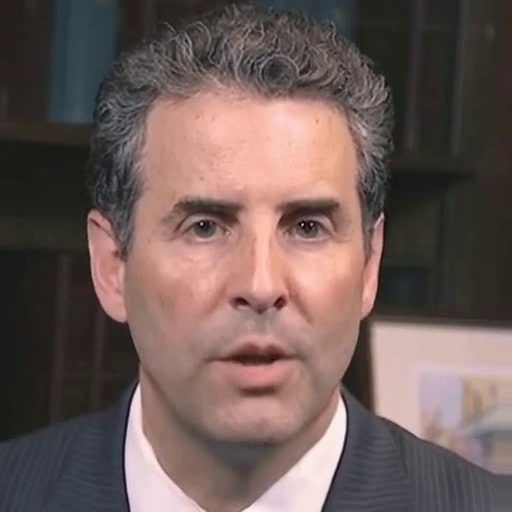}\hfill &
\includegraphics[width=\linewidth]{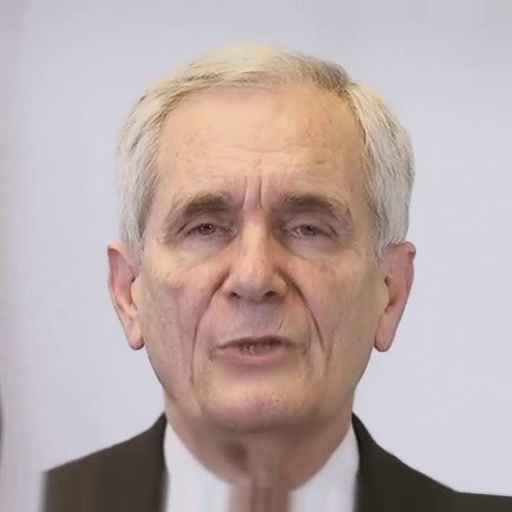}\hfill &
\includegraphics[width=\linewidth]{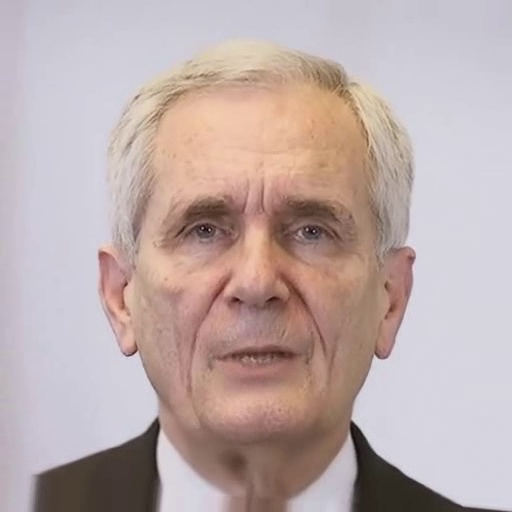}\hfill &
\includegraphics[width=\linewidth]{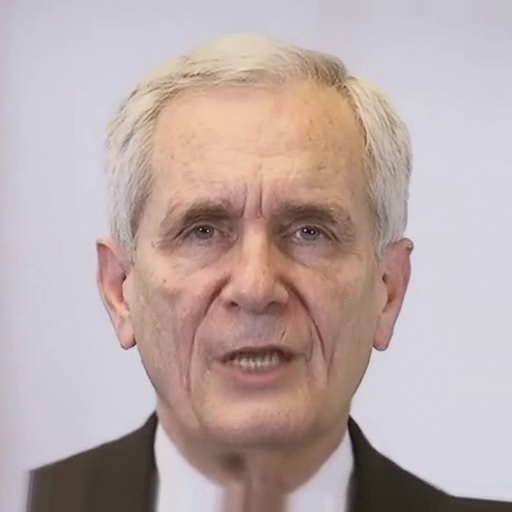}\hfill &
\includegraphics[width=\linewidth]{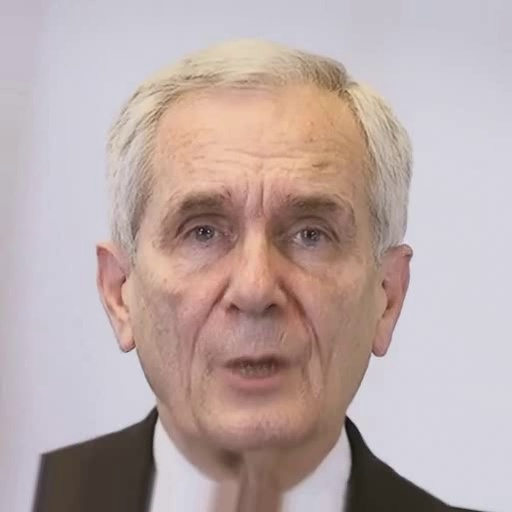}\hfill \\

\textbf{Talk3D (Ours)} &
\includegraphics[width=\linewidth]{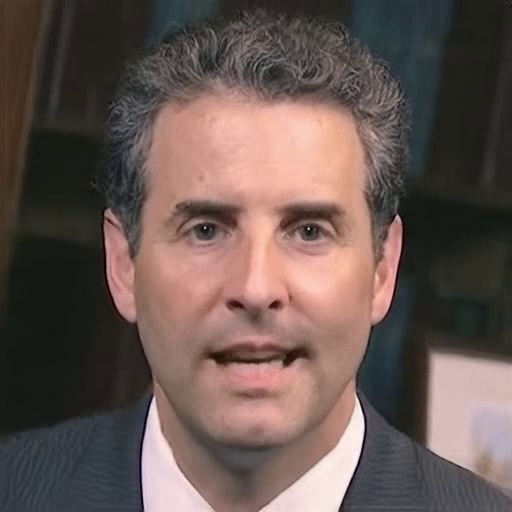}\hfill &
\includegraphics[width=\linewidth]{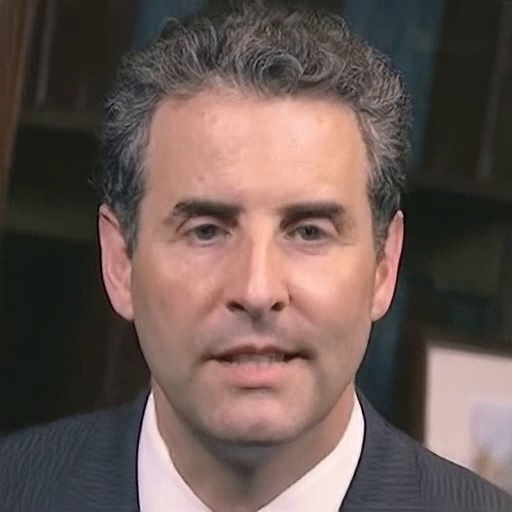}\hfill &
\includegraphics[width=\linewidth]{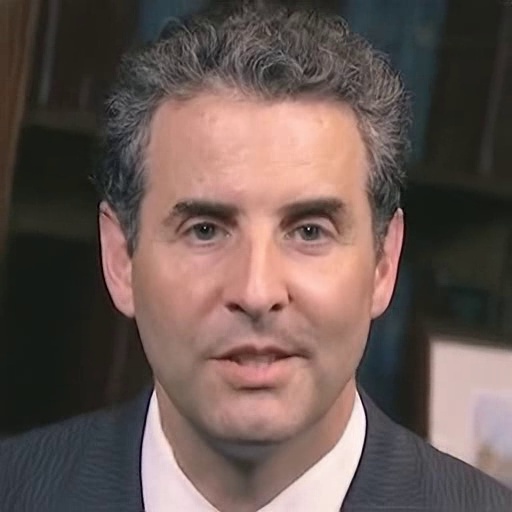}\hfill &
\includegraphics[width=\linewidth]{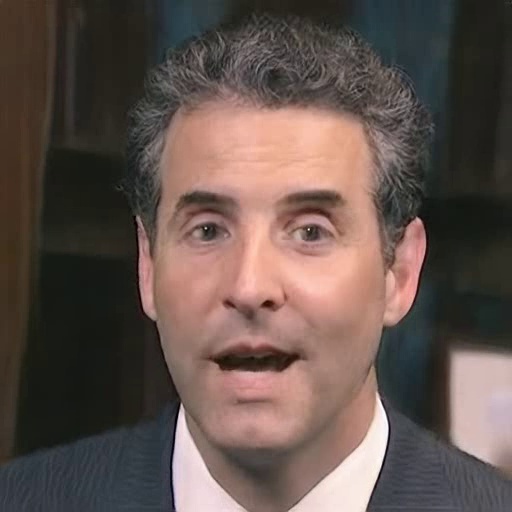}\hfill &
\includegraphics[width=\linewidth]{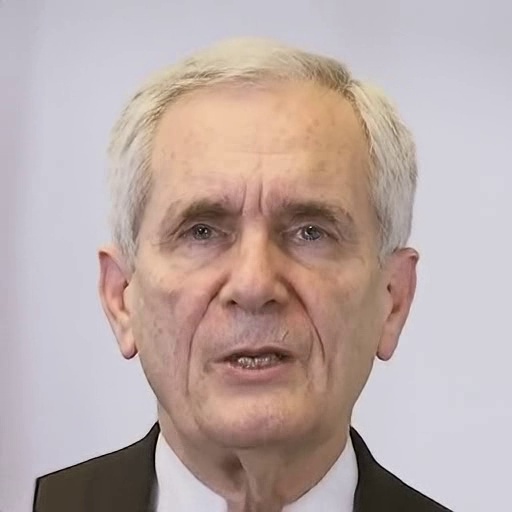}\hfill &
\includegraphics[width=\linewidth]{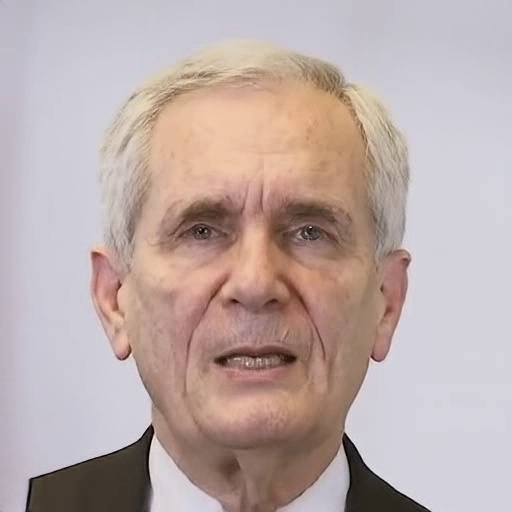}\hfill &
\includegraphics[width=\linewidth]{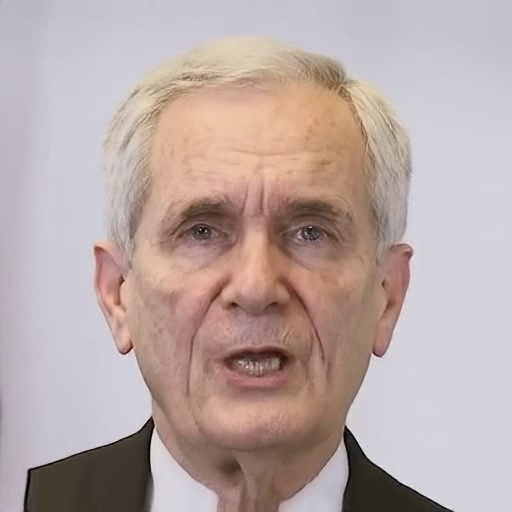}\hfill &
\includegraphics[width=\linewidth]{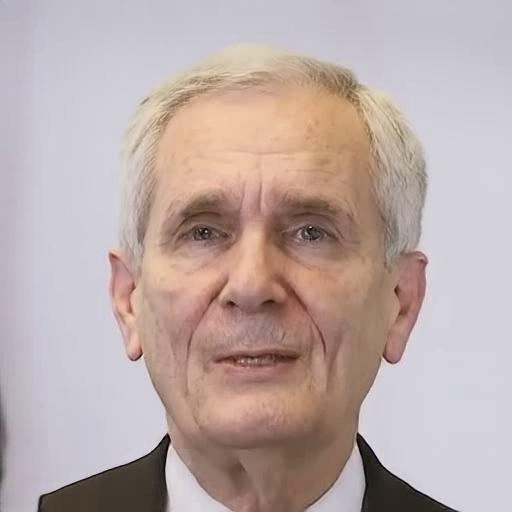}\hfill \\

\end{tabular}}
\vspace{-5pt}
\caption{\textbf{The \emph{cross-driven} comparison of the key frames and details of generated portraits.} We show visualizations of our method and related methods under the cross-driven setting. Best viewed in zoom.}
\label{supfig:cross_driven}
\vspace{-10pt}
\end{figure*}

\begin{sloppypar}
To further demonstrate the generalizability of our method, besides the dataset~\cite{guo2021adnerf} in the main paper, we also conduct experiments on datasets from HDTF~\cite{zhang2021hdtf}, which are the YouTube video clips containing in-the-wild talking portraits of 720p or 1080p resolution.
We utilize the same facial cropping method of VIVE3D~\cite{fruhstuck2023vive3d} and split each video for both training and validation.
We show additional generated results on HDTF dataset under the \emph{self-driven} setting and the \emph{cross-driven} setting in \figref{supfig:self_driven} and \figref{supfig:cross_driven}, respectively. 
\end{sloppypar}

\subsection{Comparison with HFA-GP }
HFA-GP~\cite{bai2023hfagp} is a similar work that utilizes a personalized 3D-aware generative prior for talking head synthesis. Considering the high relevance of HFA-GP, it is crucial to compare HFA-GP and our model thoroughly.
However, the implementation code of HFA-GP was incomplete, lacking essential components crucial for reproducing the experimental setup outlined in the original paper. 
We attempted to fix errors in the code, yet the outcomes differed significantly from those reported in the original paper. These discrepancies are not entirely fair to them, thus the results in \tabref{suptab:hfagp_quan} and \figref{supfig:hfagp_qual} are limited to provide an accurate comparison for the model performance.

\subsection{Comparison with GeneFace }
GeneFace~\cite{ye2022geneface} is another NeRF-based talking portrait synthesis method, but instead of directly conditioning the NeRF model on audio features, it employs audio-to-motion mapping trained on a corpus of diverse talking heads. We compare our method with GeneFace on \tabref{suptab:geneface} and \figref{supfig:geneface}. Note that the GeneFace is designed for different settings, the comparisons are not entirely fair and are just for reference.


\begin{table*}[t]
\centering

\caption{\textbf{Quantitative comparison against HFA-GP~\cite{bai2023hfagp} at the \emph{self-driven} setting.} The HFA-GP results are not an accurate representation of their work and are just for reference. }
\vspace{-5pt}

\setlength{\tabcolsep}{3.7mm}
\newcolumntype{M}[1]{>{\centering\arraybackslash}m{#1}}
\resizebox{1.\linewidth}{!}{
\begin{tabular}{ m{0.20\linewidth}  M{0.10\linewidth} M{0.10\linewidth} M{0.10\linewidth} M{0.10\linewidth} M{0.10\linewidth} M{0.10\linewidth} M{0.10\linewidth} M{0.10\linewidth} } 
\toprule
Methods & PSNR $\uparrow$ & SSIM $\uparrow$  & LPIPS $\downarrow$ & FID $\downarrow$ & LMD $\downarrow$ & AUE $\downarrow$ & Sync $\uparrow$ & IDSIM $\uparrow$ \\
\midrule
Ground Truth  & N/A            &  N/A               & $0$              & $0$              & $0$              & $0$         & $9.077$   & $1$            \\ \midrule
  
HFA-GP~\cite{bai2023hfagp} & $19.612$ & $0.712$ & $0.164$ & $25.397$ & $5.161$ & $3.146$ & $0.474$ & $0.913$ \\
\textbf{Talk3D (Ours)} & {$30.185$}       & {$0.895$} & {$0.027$} & {$8.626$} & {$2.932$}    & {$1.920$}          & {$7.383$}  & {$0.944$}         \\ \arrayrulecolor{black!100} \bottomrule 
\end{tabular}
}

\vspace{-10pt}
\label{suptab:hfagp_quan}
\end{table*}

\begin{figure*}[!htp]
\newcolumntype{M}[1]{>{\centering\arraybackslash}m{#1}}
\setlength{\tabcolsep}{0.2pt}
\renewcommand{\arraystretch}{0.1}
\centering
\scriptsize

\resizebox{\linewidth}{!}{
\begin{tabular}{M{0.09\linewidth}@{\hskip 0.005\linewidth}  M{0.12\linewidth}M{0.12\linewidth}M{0.12\linewidth}  M{0.12\linewidth} @{\hskip 0.01\linewidth}  M{0.12\linewidth} M{0.12\linewidth}M{0.12\linewidth}M{0.12\linewidth}}

& o\textcolor{red}{f} & sh\textcolor{red}{o}p & \textcolor{red}{s}tart & b\textcolor{red}{ack} & g\textcolor{red}{re}w & with\textcolor{red}{ou}t & \textcolor{red}{ar}ound & presen\textcolor{red}{ts} \\ \\

\makecell{\vspace{-12pt}\\ Ground \\ Truth} &
\includegraphics[width=\linewidth]{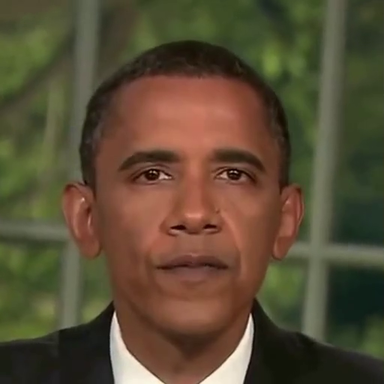}\hfill &
\includegraphics[width=\linewidth]{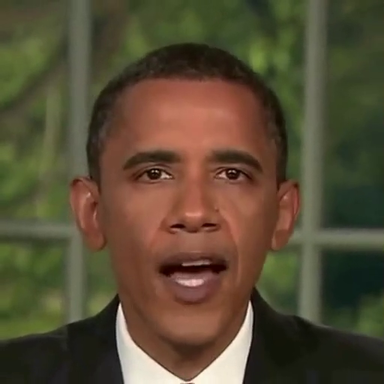}\hfill &
\includegraphics[width=\linewidth]{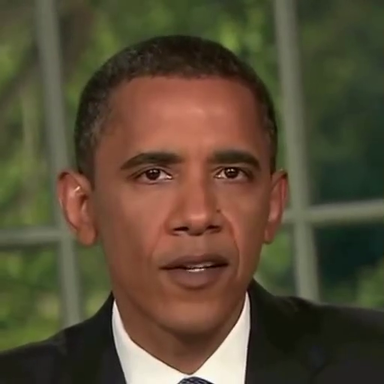}\hfill &
\includegraphics[width=\linewidth]{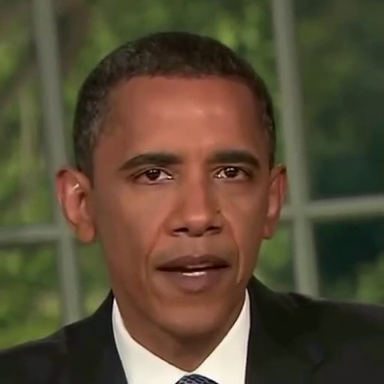}\hfill &
\includegraphics[width=\linewidth]{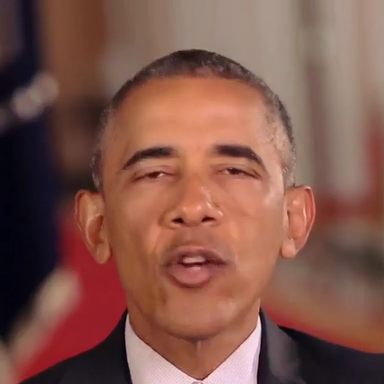}\hfill &
\includegraphics[width=\linewidth]{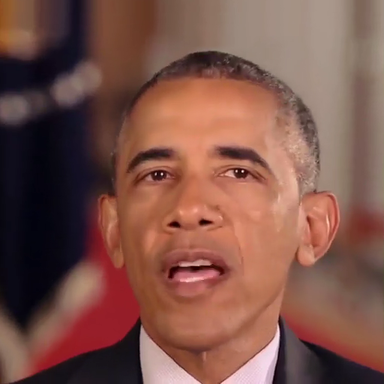}\hfill &
\includegraphics[width=\linewidth]{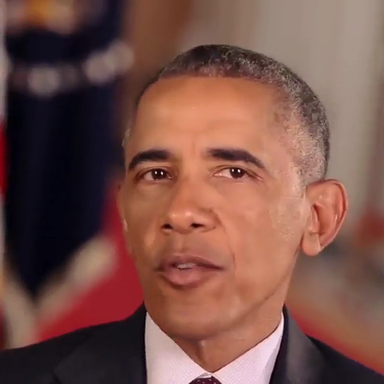}\hfill &
\includegraphics[width=\linewidth]{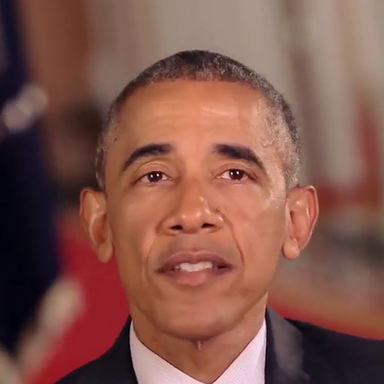}\hfill \\

HFA-GP &
\includegraphics[width=\linewidth]{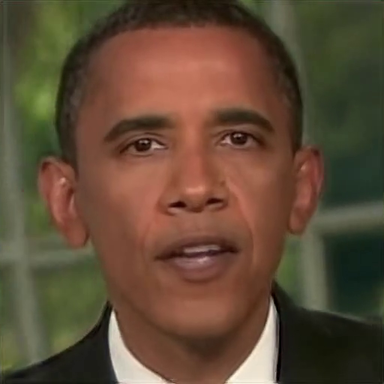}\hfill &
\includegraphics[width=\linewidth]{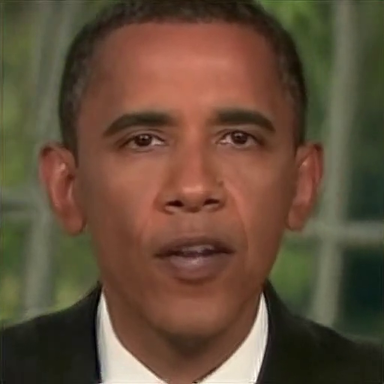}\hfill &
\includegraphics[width=\linewidth]{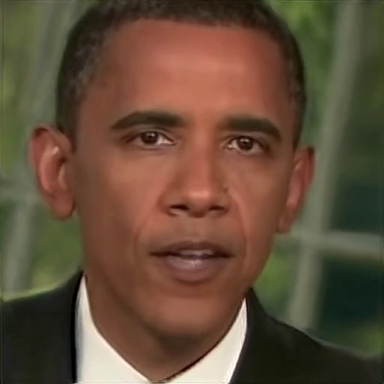}\hfill &
\includegraphics[width=\linewidth]{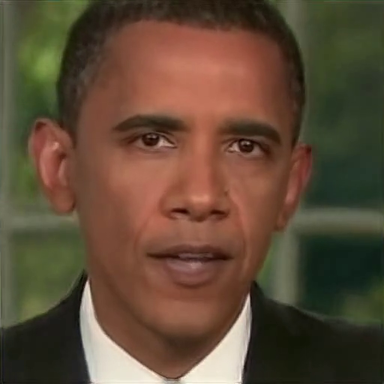}\hfill &
\includegraphics[width=\linewidth]{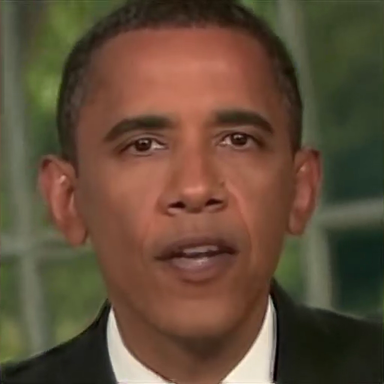}\hfill &
\includegraphics[width=\linewidth]{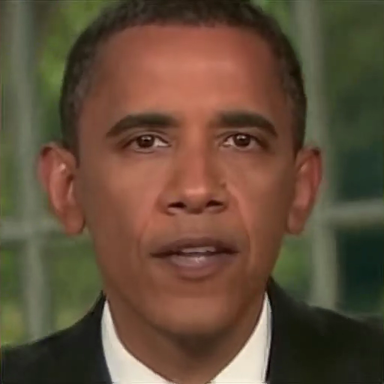}\hfill &
\includegraphics[width=\linewidth]{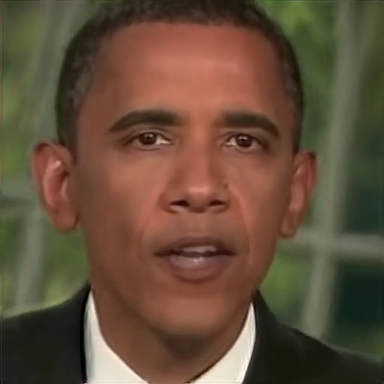}\hfill &
\includegraphics[width=\linewidth]{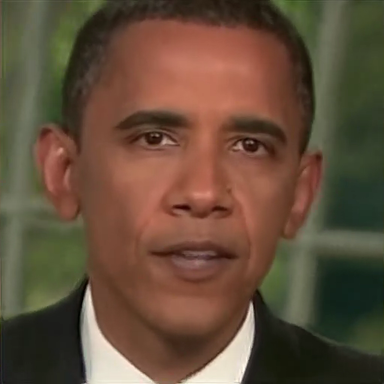}\hfill \\

\textbf{Talk3D (Ours)} &
\includegraphics[width=\linewidth]{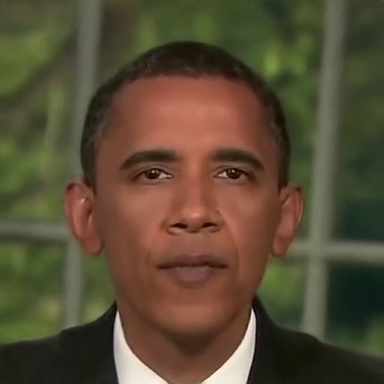}\hfill &
\includegraphics[width=\linewidth]{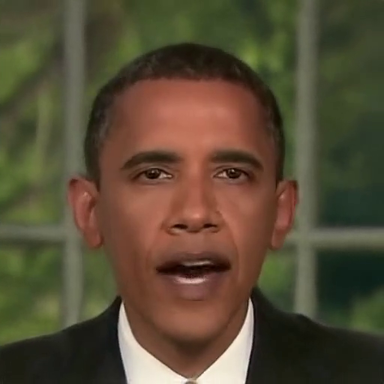}\hfill &
\includegraphics[width=\linewidth]{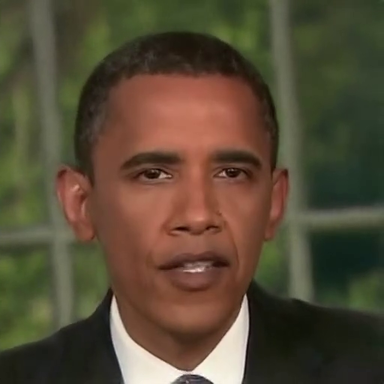}\hfill &
\includegraphics[width=\linewidth]{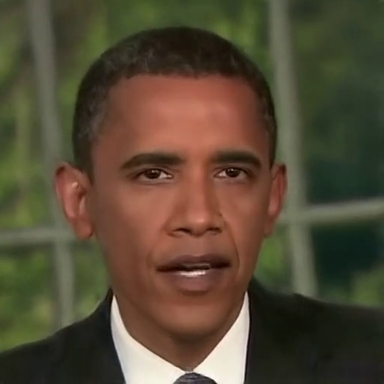}\hfill &
\includegraphics[width=\linewidth]{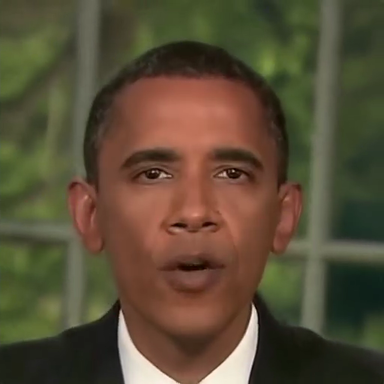}\hfill &
\includegraphics[width=\linewidth]{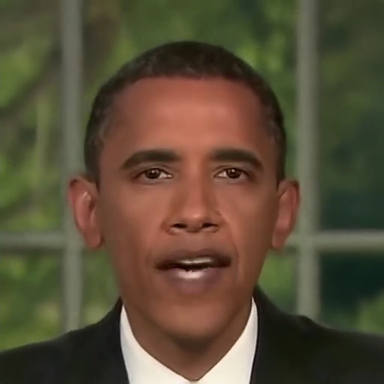}\hfill &
\includegraphics[width=\linewidth]{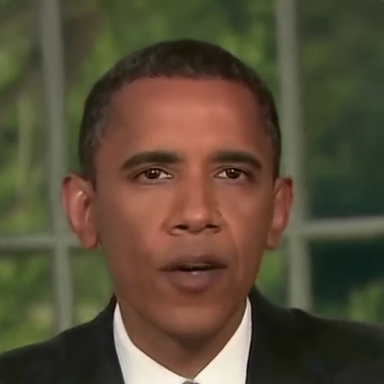}\hfill &
\includegraphics[width=\linewidth]{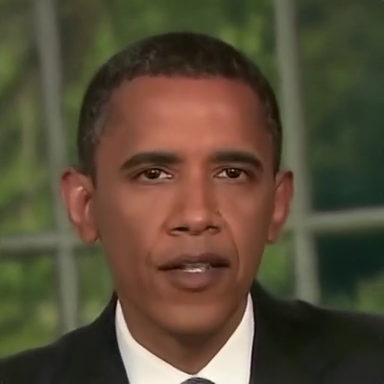}\hfill \\

\end{tabular}}
\vspace{-5pt}
\caption{\textbf{The comparison of the key frames and details of generated portraits.} We show visualizations of our method and HFA-GP under the self-driven setting (left side) and the cross-driven setting (right side).}
\label{supfig:hfagp_qual}
\vspace{-10pt}
\end{figure*}

\begin{table*}[t]
\centering

\caption{\textbf{Quantitative comparison against GeneFace~\cite{ye2022geneface} at the \emph{self-driven} setting.} Across all evaluation metrics, Talk3D exhibits consistently higher performance than GeneFace.}
\vspace{-5pt}

\setlength{\tabcolsep}{3.7mm}
\newcolumntype{M}[1]{>{\centering\arraybackslash}m{#1}}
\resizebox{1.\linewidth}{!}{
\begin{tabular}{ m{0.20\linewidth}  M{0.10\linewidth} M{0.10\linewidth} M{0.10\linewidth} M{0.10\linewidth} M{0.10\linewidth} M{0.10\linewidth} M{0.10\linewidth} M{0.10\linewidth} } 
\toprule
Methods & PSNR $\uparrow$ & SSIM $\uparrow$  & LPIPS $\downarrow$ & FID $\downarrow$ & LMD $\downarrow$ & AUE $\downarrow$ & Sync $\uparrow$ & IDSIM $\uparrow$ \\
\midrule
Ground Truth  & N/A            &  N/A               & $0$              & $0$              & $0$              & $0$         & $9.077$   & $1$            \\ \midrule
  
GeneFace~\cite{ye2022geneface} & $26.305$ & $0.832$ & $0.069$ & $15.30$ & $4.948$ & $2.758$ & $6.200$ & $0.909$ \\
\textbf{Talk3D (Ours)} & {$30.185$}       & {$0.895$} & {$0.027$} & {$8.626$} & {$2.932$}    & {$1.920$}          & {$7.383$}  & {$0.944$}         \\ \arrayrulecolor{black!100} \bottomrule 
\end{tabular}
}
\vspace{-10pt}
\label{suptab:geneface}
\end{table*}

 \begin{figure*}[!htp]
\newcolumntype{M}[1]{>{\centering\arraybackslash}m{#1}}
\setlength{\tabcolsep}{0.2pt}
\renewcommand{\arraystretch}{0.1}
\centering
\scriptsize

\resizebox{1.\linewidth}{!}{
\begin{tabular}{M{0.1\linewidth}@{\hskip 0.005\linewidth}  M{0.15\linewidth}  M{0.15\linewidth}  M{0.15\linewidth}   @{\hskip 0.01\linewidth}  M{0.15\linewidth}   M{0.15\linewidth}  M{0.15\linewidth}   }

& \textcolor{red}{peo}ple & $\langle$ mute $\rangle$  & majo\textcolor{red}{ri}ty & fa\textcolor{red}{th}er & li\textcolor{red}{ke} & every\textcolor{red}{da}y \\ \\

\makecell{\vspace{-12pt}\\ Ground \\ Truth} &
\includegraphics[width=\linewidth]{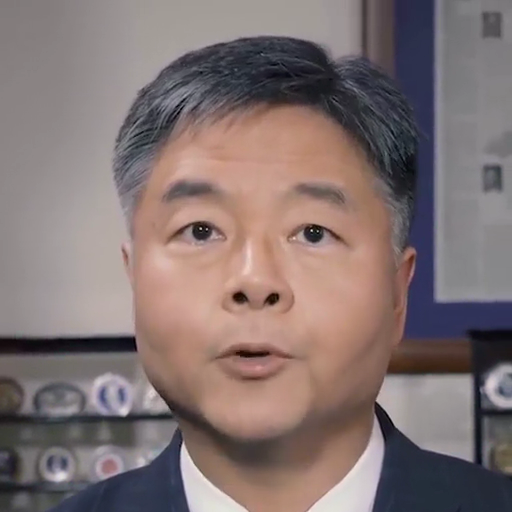}\hfill &
\includegraphics[width=\linewidth]{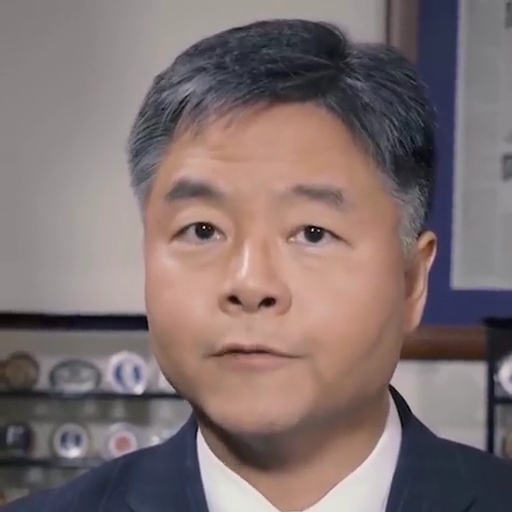}\hfill &
\includegraphics[width=\linewidth]{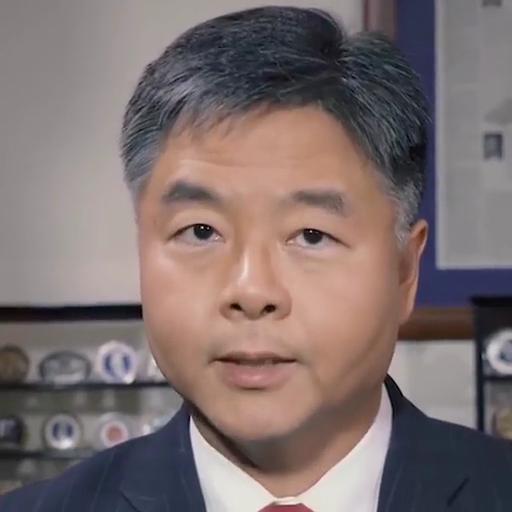}\hfill &\includegraphics[width=\linewidth]{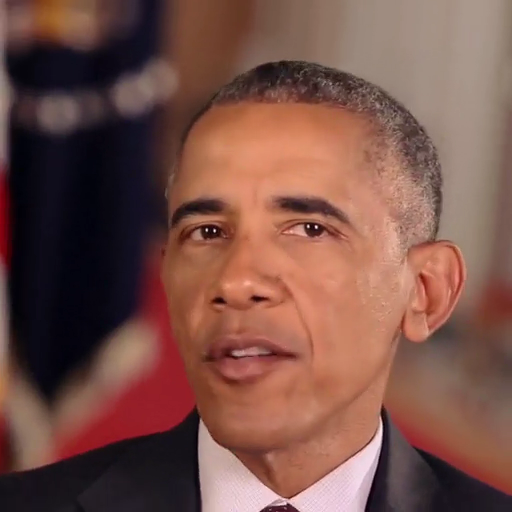}\hfill &
\includegraphics[width=\linewidth]{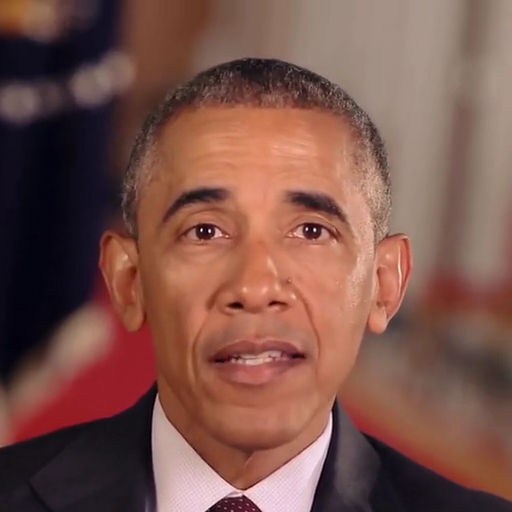}\hfill &
\includegraphics[width=\linewidth]{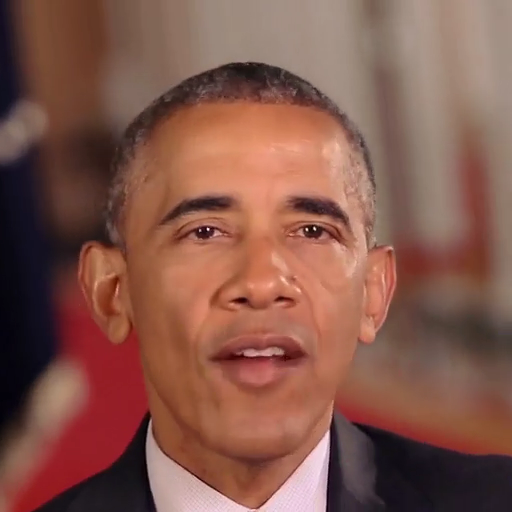}\hfill \\

Gene- Face &
\includegraphics[width=\linewidth]{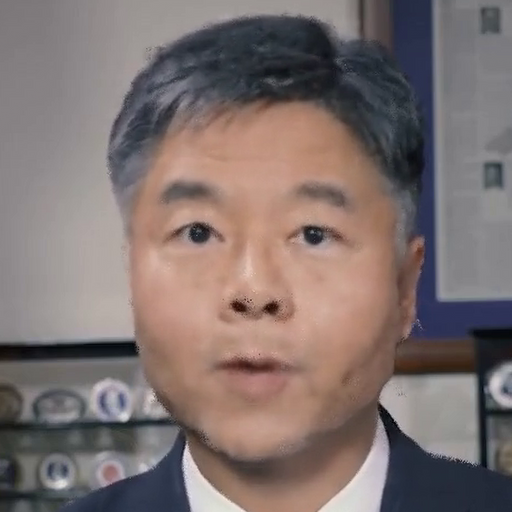}\hfill &
\includegraphics[width=\linewidth]{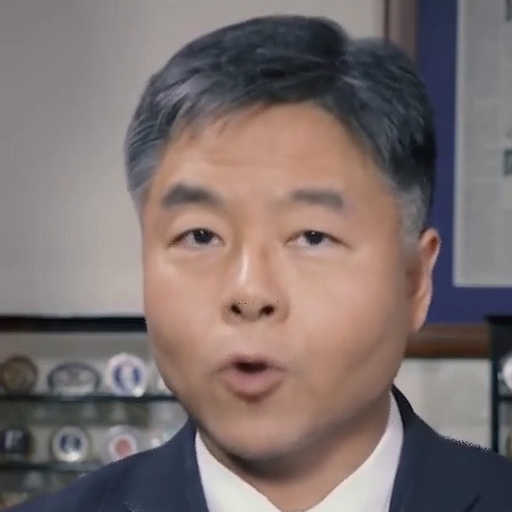}\hfill &
\includegraphics[width=\linewidth]{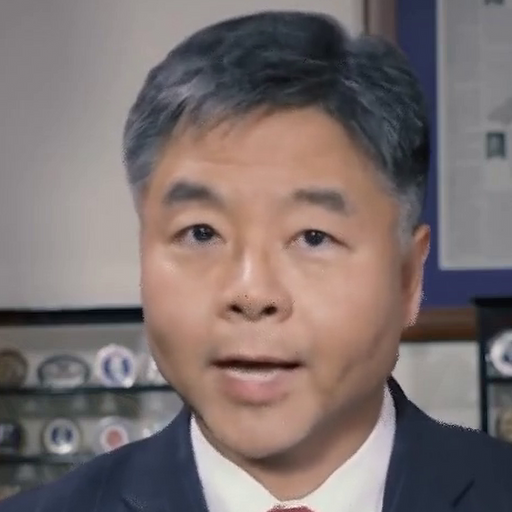}\hfill &
\includegraphics[width=\linewidth]{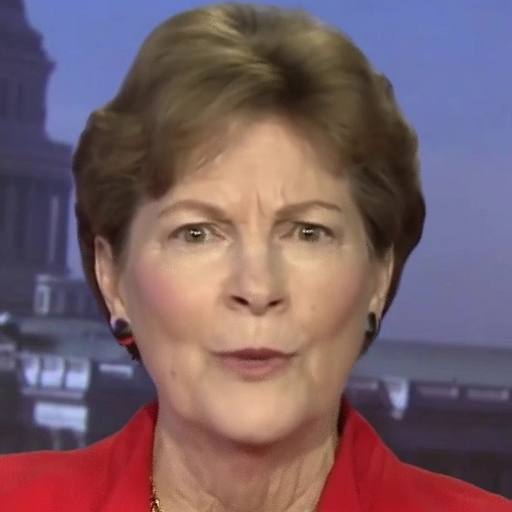}\hfill &
\includegraphics[width=\linewidth]{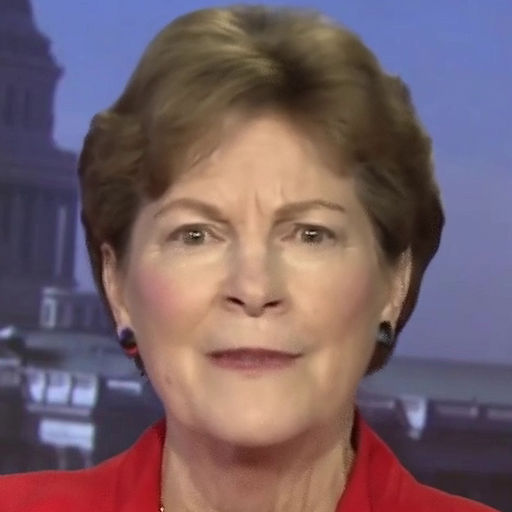}\hfill &
\includegraphics[width=\linewidth]{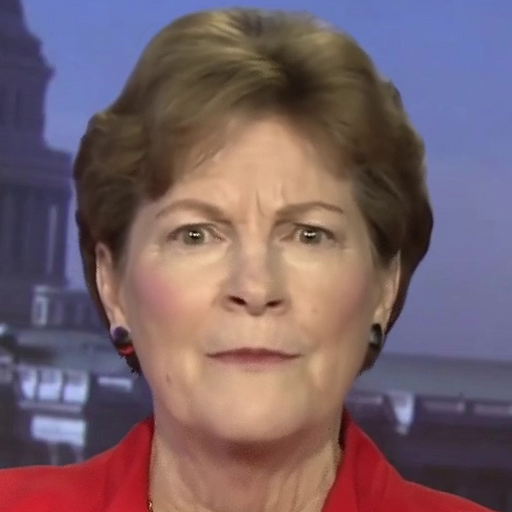}\hfill \\

\textbf{Talk3D (Ours)} &
\includegraphics[width=\linewidth]{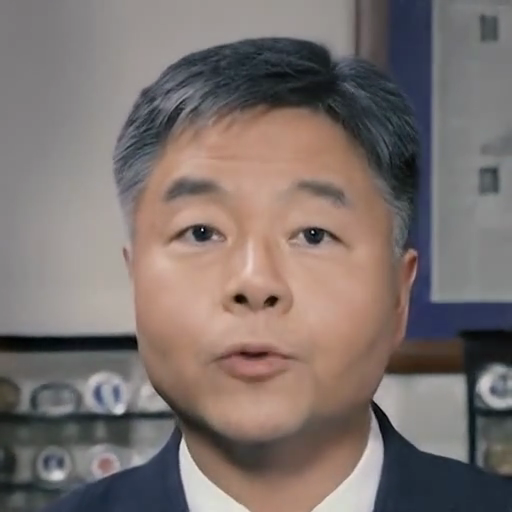}\hfill &
\includegraphics[width=\linewidth]{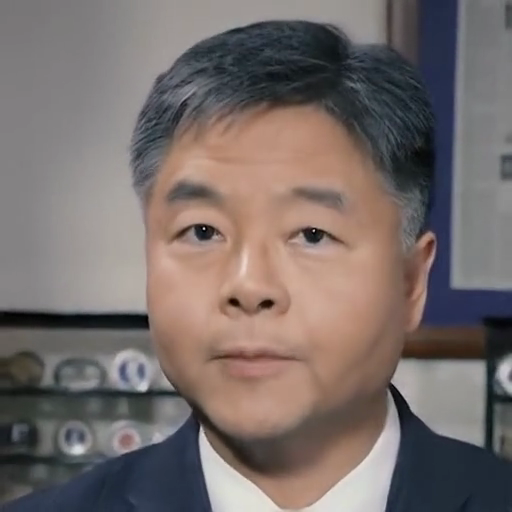}\hfill &
\includegraphics[width=\linewidth]{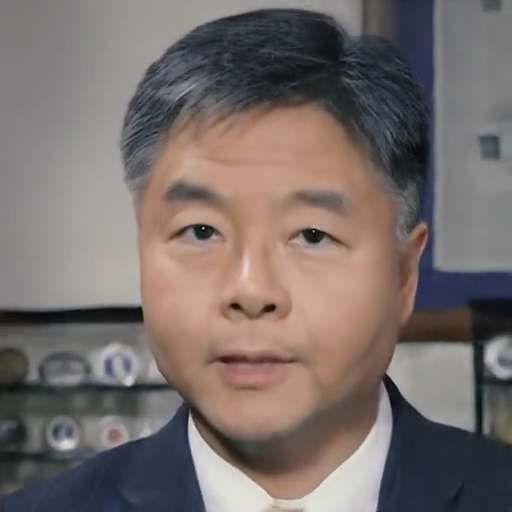}\hfill &
\includegraphics[width=\linewidth]{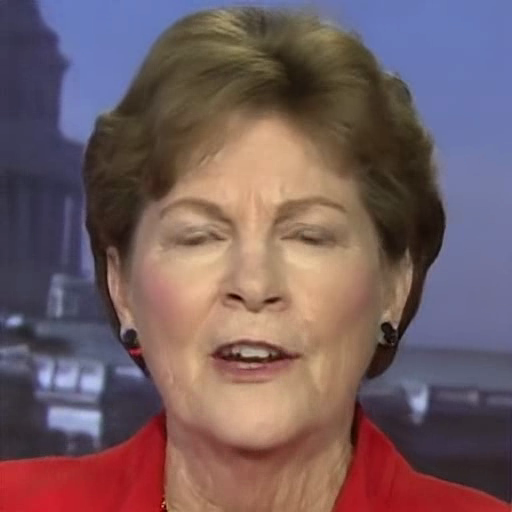}\hfill &
\includegraphics[width=\linewidth]{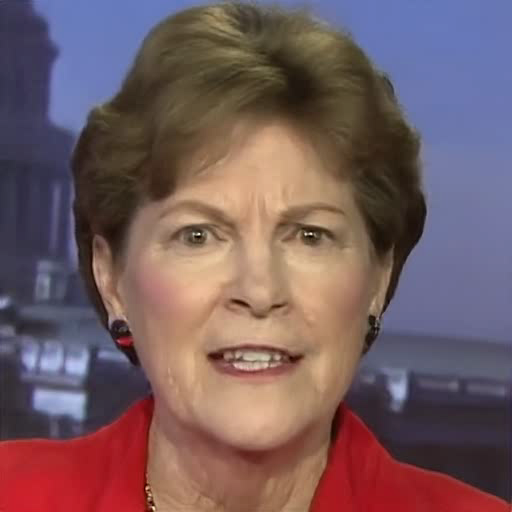}\hfill &
\includegraphics[width=\linewidth]{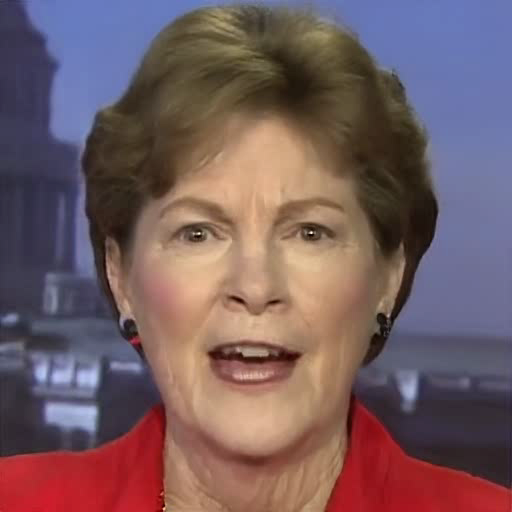}\hfill \\

\end{tabular}}
\vspace{-5pt}
\caption{\textbf{The comparison of the key frames and details of generated portraits.} We show visualizations of our method and GeneFace under the self-driven setting (left side) and the cross-driven setting (right side).}
\label{supfig:geneface}
\vspace{-10pt}
\end{figure*}

\section{Further Analysis}
\label{supsec:further_analysis}
\subsection{Analysis of attention}

\begin{figure}[!h]
\newcolumntype{M}[1]{>{\centering\arraybackslash}m{#1}}
\setlength{\tabcolsep}{0.5pt}
\renewcommand{\arraystretch}{0.25}
\centering
\scriptsize
\resizebox{1.\linewidth}{!}{
\begin{tabular}{M{0.19\linewidth} @{\hskip 0.015\linewidth} M{0.19\linewidth} M{0.19\linewidth}M{0.19\linewidth}M{0.19\linewidth}}

Result & Audio & Eye & Angles & Landmarks \\ \\

\includegraphics[width=\linewidth]{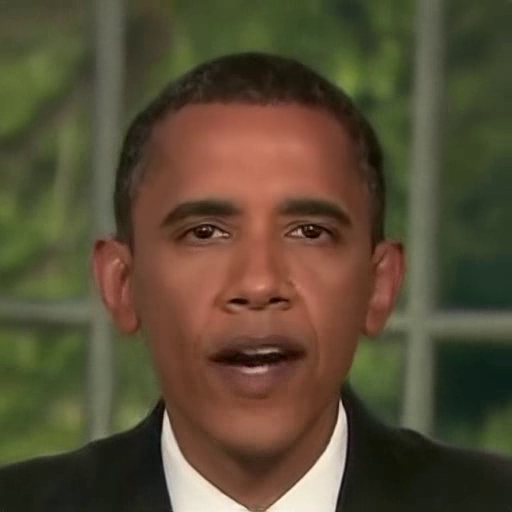}\hfill &
\includegraphics[width=\linewidth]{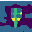}\hfill &
\includegraphics[width=\linewidth]{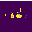}\hfill &
\includegraphics[width=\linewidth]{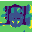}\hfill &
\includegraphics[width=\linewidth]{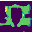}\hfill \\

\includegraphics[width=\linewidth]{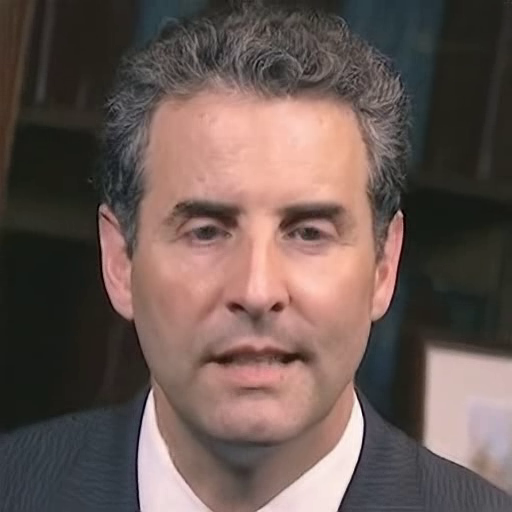}\hfill &
\includegraphics[width=\linewidth]{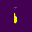}\hfill &
\includegraphics[width=\linewidth]{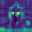}\hfill &
\includegraphics[width=\linewidth]{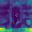}\hfill &
\includegraphics[width=\linewidth]{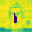}\hfill \\

\end{tabular}} \vspace{-5pt}
\caption{\textbf{Visualizations of attention maps.} Our region attention module successfully captures the relation between diverse conditioning tokens and spatial regions.}
\label{fig:ablation_token}
\end{figure}

In \figref{fig:ablation_token}, we visualize the attention map to demonstrate the efficacy of our attention-based network. The first column is the generated image result, while the rest of the columns show the attention map of the low-resolution xy-plane (the plane that is orthogonal to canonical direction) captured by each of the specific conditioning tokens. The result shows that each of the conditioning tokens successfully disentangles the local movements of the low-resolution feature map. 
Despite the close relationship between angles and landmarks, they capture different attention maps, since the head rotation angles are closely related to torso movement, while facial landmarks are suitable for capturing the background motion.

\subsection{Analysis of triplane}
\begin{figure}[!ht]
\newcolumntype{M}[1]{>{\centering\arraybackslash}m{#1}}
\setlength{\tabcolsep}{0.5pt}
\renewcommand{\arraystretch}{0.25}
\centering
\scriptsize
\resizebox{1.\linewidth}{!}{
\begin{tabular}{M{0.115\linewidth} M{0.14\linewidth} M{0.14\linewidth} M{0.14\linewidth} 
@{\hskip 0.015\linewidth} M{0.14\linewidth}M{0.14\linewidth}M{0.14\linewidth}}

Results
&
\includegraphics[width=\linewidth]{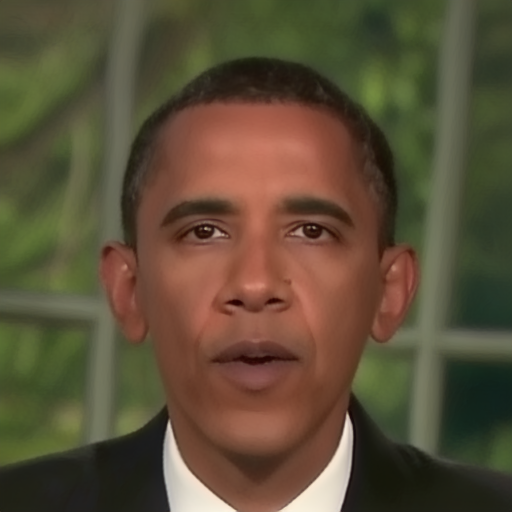}\hfill &
 &
 &
\includegraphics[width=\linewidth]{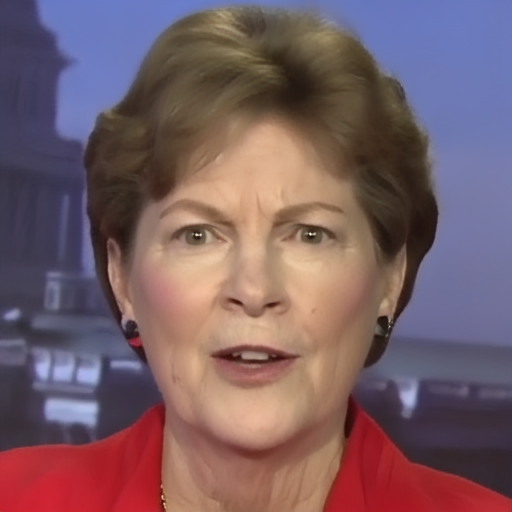}\hfill &
 &
 \\

Identity plane
&
\includegraphics[width=\linewidth]{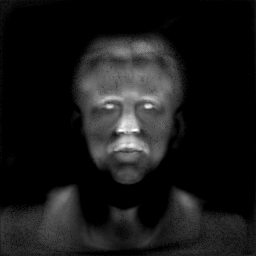}\hfill &
\includegraphics[width=\linewidth]{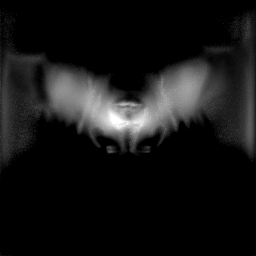}\hfill &
\includegraphics[width=\linewidth]{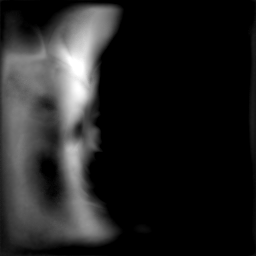}\hfill &
\includegraphics[width=\linewidth]{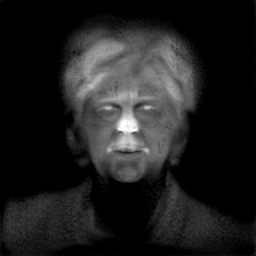}\hfill &
\includegraphics[width=\linewidth]{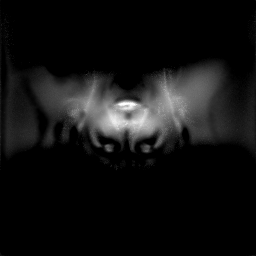}\hfill &
\includegraphics[width=\linewidth]{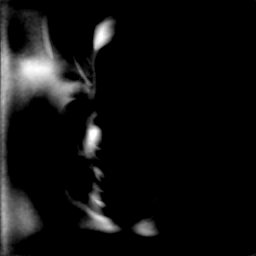}\hfill \\

Delta plane
&
\includegraphics[width=\linewidth]{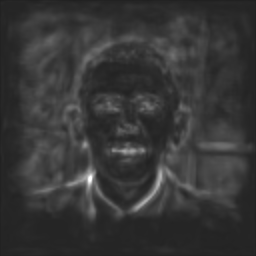}\hfill &
\includegraphics[width=\linewidth]{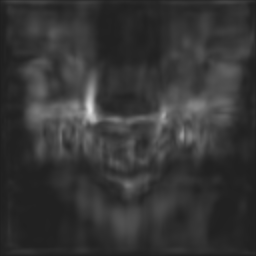}\hfill &
\includegraphics[width=\linewidth]{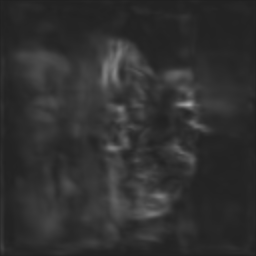}\hfill &
\includegraphics[width=\linewidth]{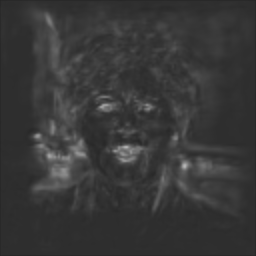}\hfill &
\includegraphics[width=\linewidth]{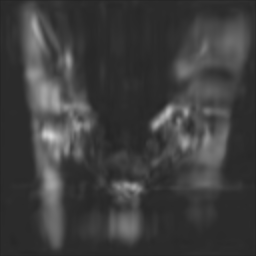}\hfill &
\includegraphics[width=\linewidth]{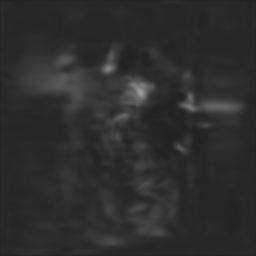}\hfill \\

\end{tabular}} \vspace{-5pt}
\caption{\textbf{Visualizations of triplanes.} We visualize generated image results and their corresponding triplanes. Each set of three columns depicts the orthogonal planes of the triplane representation.}
\label{fig:plane_vis}
\end{figure}

\figref{fig:plane_vis} visualizes the generated image alongside its two corresponding triplanes: the identity plane and the deltaplane. Each column shows the three orthogonal planes. Especially xy-plane (orthogonal to canonical direction) in the 1st and 4th columns highlights the facial structure representation within the identity plane. Also, the deltaplane visualization confirms our method's ability to precisely manipulate specific regions such as the lips, eyes, torso, and background.

\vspace{30pt}
\subsection{Ablation studies}

\subsubsection{Importance of each token selection. }
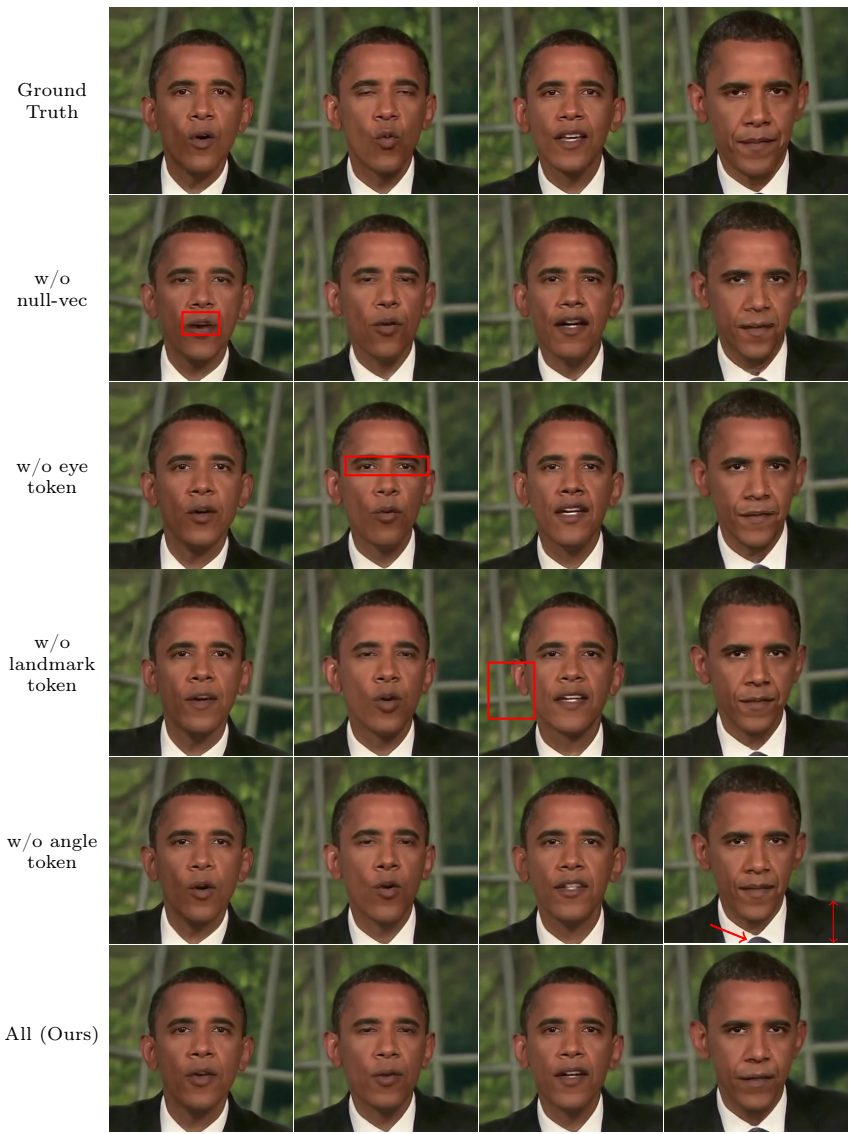
\begin{figure*}[!p]
\newcommand{\rom}[1]{\uppercase\expandafter{\romannumeral #1\relax}}
\newcommand{\RomanNumeralCaps}[1]{\MakeUppercase{\romannumeral #1}}
\newcolumntype{M}[1]{>{\centering\arraybackslash}m{#1}}
\setlength{\tabcolsep}{0.5pt}
\renewcommand{\arraystretch}{0.25}
\centering
\scriptsize

\begin{tabular}{ M{0.12\linewidth} M{0.20\linewidth}  M{0.20\linewidth} M{0.20\linewidth}M{0.20\linewidth}   }

Ground Truth &
\includegraphics[width=\linewidth]{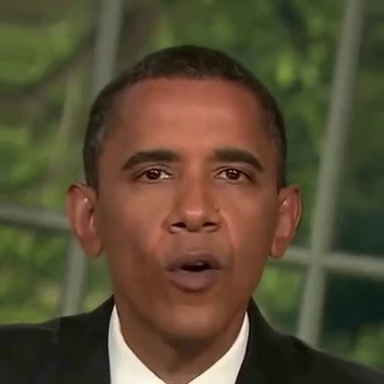}\hfill &
\includegraphics[width=\linewidth]{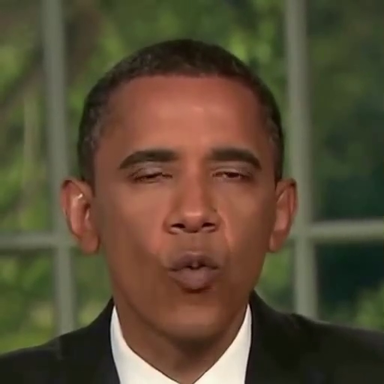}\hfill &
\includegraphics[width=\linewidth]{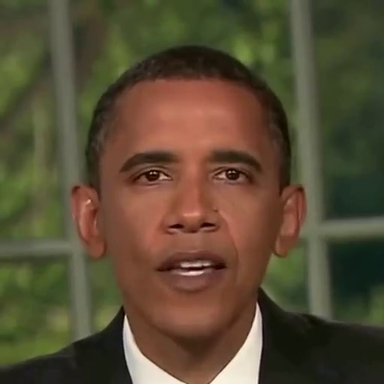}\hfill &
\includegraphics[width=\linewidth]{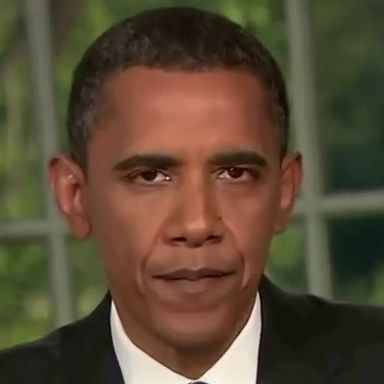}\hfill \\

w/o null-vec &
\begin{tikzpicture}
    \node[anchor=south west,inner sep=0] (image) at (0,0) {\includegraphics[width=\linewidth]{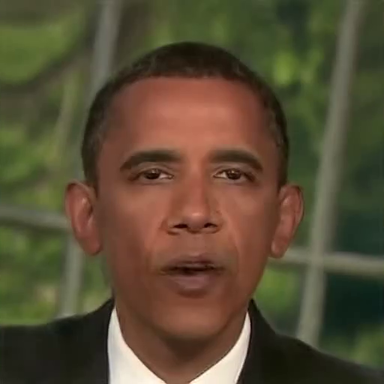}};
    \begin{scope}[x={(image.south east)},y={(image.north west)}]
        \draw[red,thick] (0.40,0.25) rectangle (0.60,0.37);
    \end{scope}
\end{tikzpicture} &
\includegraphics[width=\linewidth]{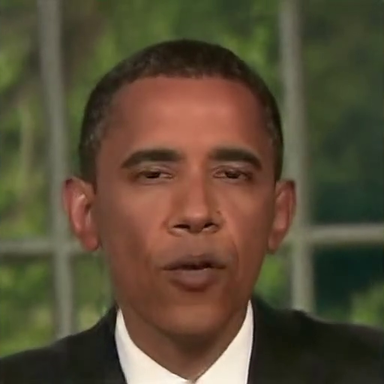}\hfill &
\includegraphics[width=\linewidth]{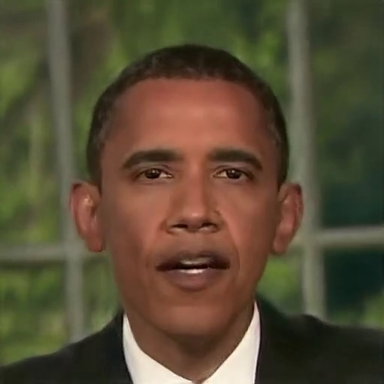}\hfill &
\includegraphics[width=\linewidth]{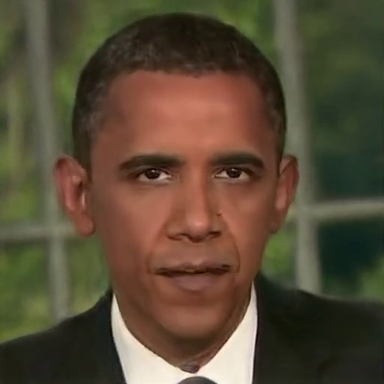}\hfill \\

w/o eye token &
\includegraphics[width=\linewidth]{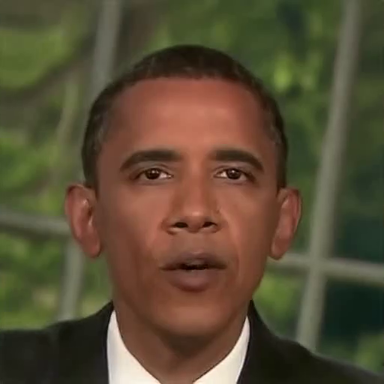}\hfill &
\begin{tikzpicture}
    \node[anchor=south west,inner sep=0] (image) at (0,0) {\includegraphics[width=\linewidth]{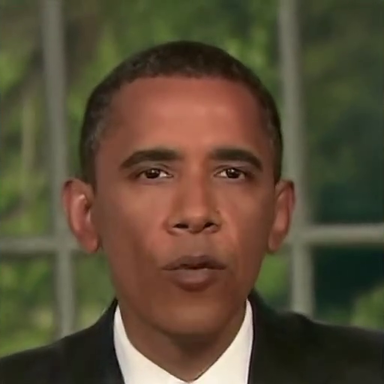}};
    \begin{scope}[x={(image.south east)},y={(image.north west)}]
        \draw[red,thick] (0.28,0.5) rectangle (0.73,0.6);
    \end{scope}
\end{tikzpicture} &
\includegraphics[width=\linewidth]{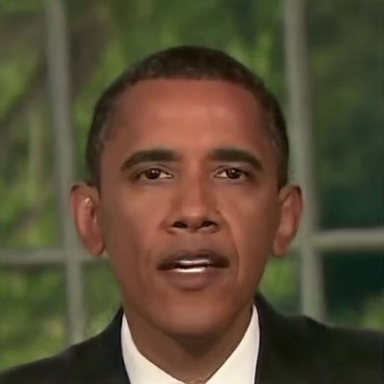}\hfill &
\includegraphics[width=\linewidth]{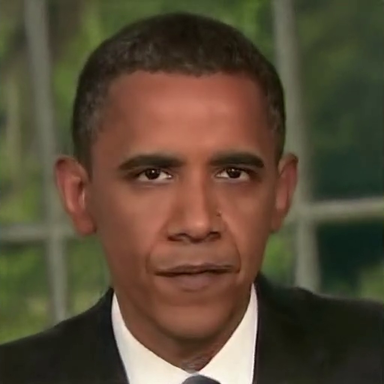}\hfill \\

w/o landmark token &
\includegraphics[width=\linewidth]{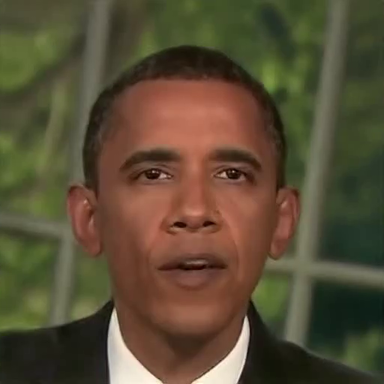}\hfill &
\includegraphics[width=\linewidth]{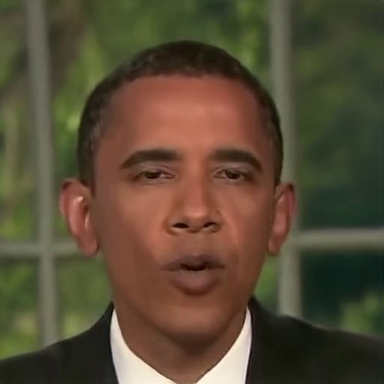}\hfill &
\begin{tikzpicture}
    \node[anchor=south west,inner sep=0] (image) at (0,0) {\includegraphics[width=\linewidth]{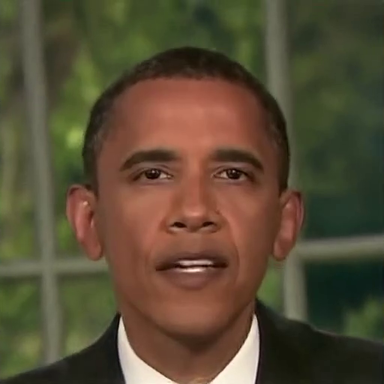}};
    \begin{scope}[x={(image.south east)},y={(image.north west)}]
        \draw[red,thick] (0.05,0.2) rectangle (0.3,0.5);
    \end{scope}
\end{tikzpicture} &
\includegraphics[width=\linewidth]{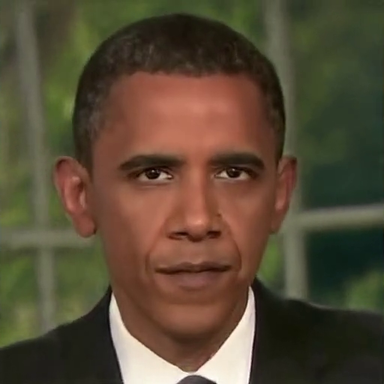}\hfill \\

w/o angle token &
\includegraphics[width=\linewidth]{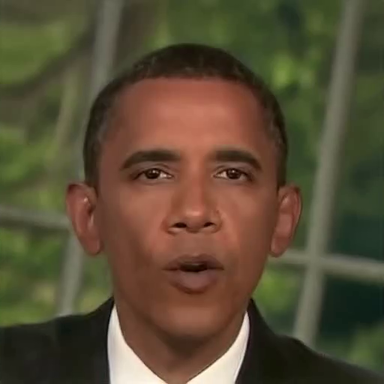}\hfill &
\includegraphics[width=\linewidth]{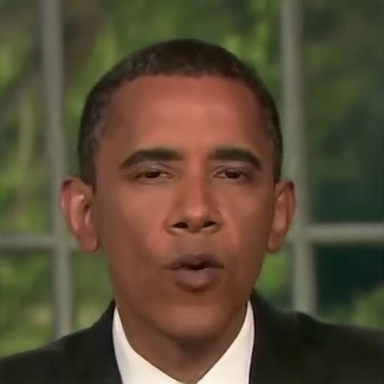}\hfill &
\includegraphics[width=\linewidth]{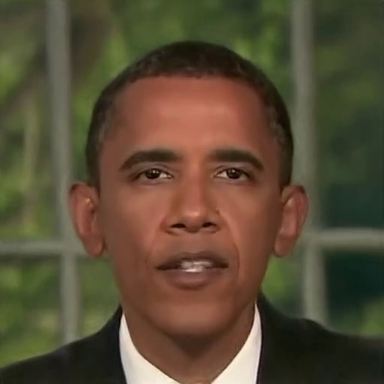}\hfill &
\begin{tikzpicture}
    \node[anchor=south west,inner sep=0] (image) at (0,0) {\includegraphics[width=\linewidth]{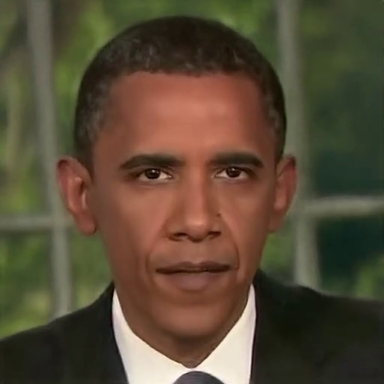}};
    \begin{scope}[x={(image.south east)},y={(image.north west)}]
        \draw[->, red, thick] (0.25, 0.1) -- (0.45, 0.02);
        \draw[<->, red] (0.92, 0.0) -- (0.92, 0.23);
    \end{scope}
\end{tikzpicture} \\

All (Ours) &
\includegraphics[width=\linewidth]{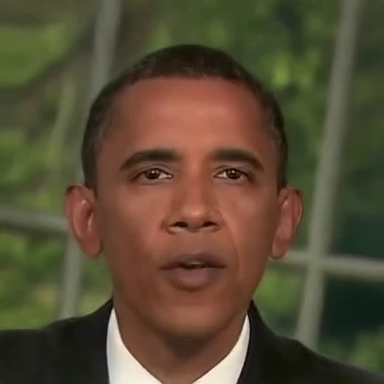}\hfill &
\includegraphics[width=\linewidth]{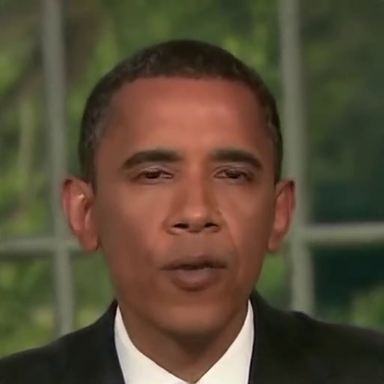}\hfill &
\includegraphics[width=\linewidth]{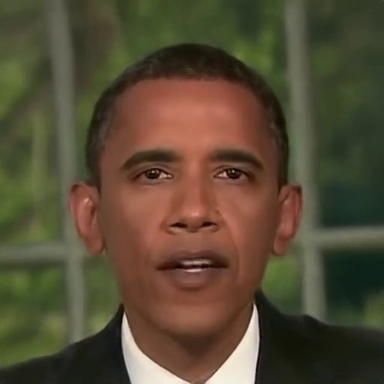}\hfill &
\includegraphics[width=\linewidth]{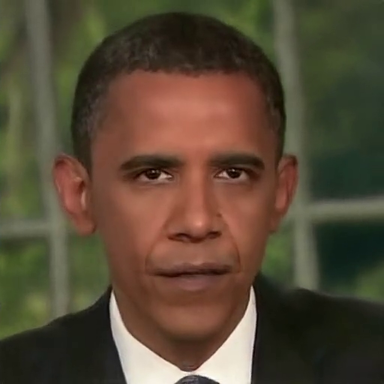}\hfill \\

\end{tabular}
\vspace{-5pt}
\caption{\textbf{Ablation study on the use of each feature token.} We assess the effectiveness of each feature token by alternatively turning them on and off.}
\label{supfig:ablation_features}
\vspace{-13pt}
\end{figure*}

\figref{supfig:ablation_features} provides a qualitative comparison from the ablation study of feature token selection (discussed in the main paper). The figure demonstrates how each feature token influences its corresponding region.
We highlight the specific regions with red boxes, that illustrate each of the image quality degradation. Eye features are crucial for accurate eye movement prediction, while angle and landmark tokens are essential in controlling torso and background movement, respectively. See the difference in the eye-blink, background movement, or neck-tie region. The results obtained without employing null vectors show artifacts, particularly in the mouth region.

\subsubsection{Deltaplane predictor. }
\begin{figure*}[!p]
\newcommand{\rom}[1]{\uppercase\expandafter{\romannumeral #1\relax}}
\newcommand{\RomanNumeralCaps}[1]{\MakeUppercase{\romannumeral #1}}
\newcolumntype{M}[1]{>{\centering\arraybackslash}m{#1}}
\setlength{\tabcolsep}{0.5pt}
\renewcommand{\arraystretch}{0.25}
\centering
\scriptsize

\resizebox{\linewidth}{!}{
\begin{tabular}{ M{0.12\linewidth} M{0.18\linewidth}  M{0.18\linewidth} M{0.18\linewidth}M{0.18\linewidth} M{0.18\linewidth}  }

Ground Truth &
\includegraphics[width=\linewidth]{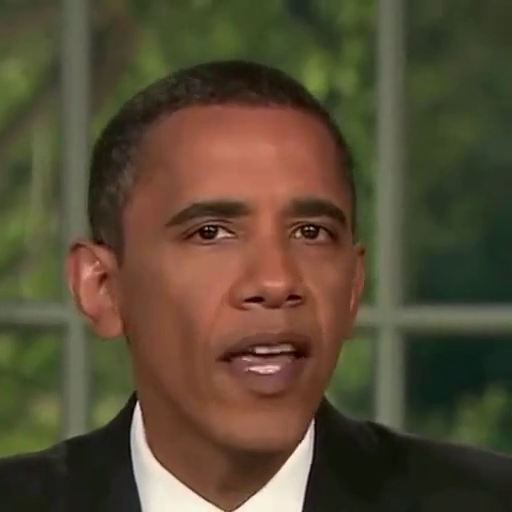}\hfill &
\includegraphics[width=\linewidth]{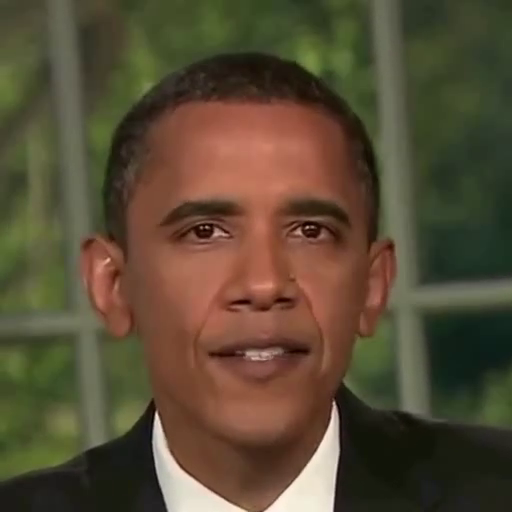}\hfill &
\includegraphics[width=\linewidth]{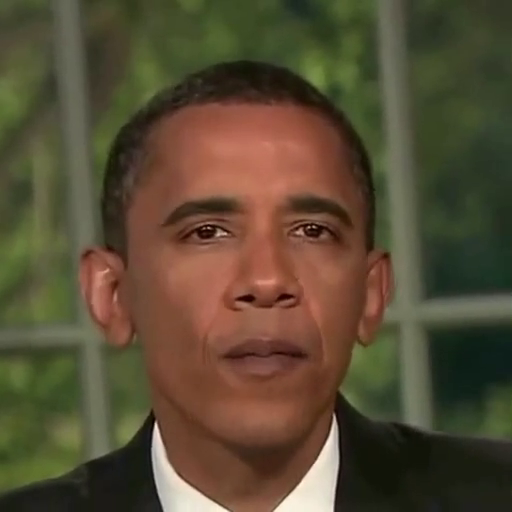}\hfill &
\includegraphics[width=\linewidth]{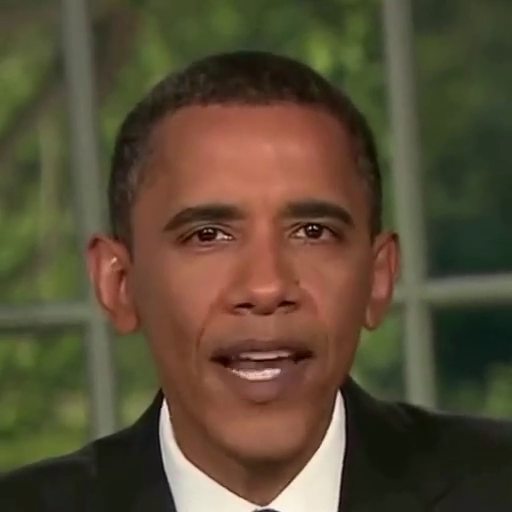}\hfill &
\includegraphics[width=\linewidth]{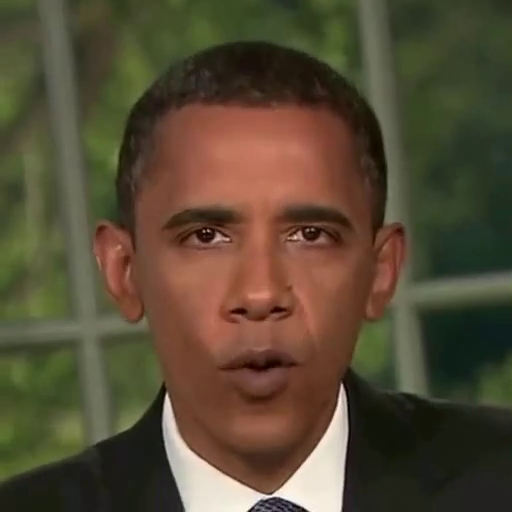}\hfill \\

w/o deltaplane &
\includegraphics[width=\linewidth]{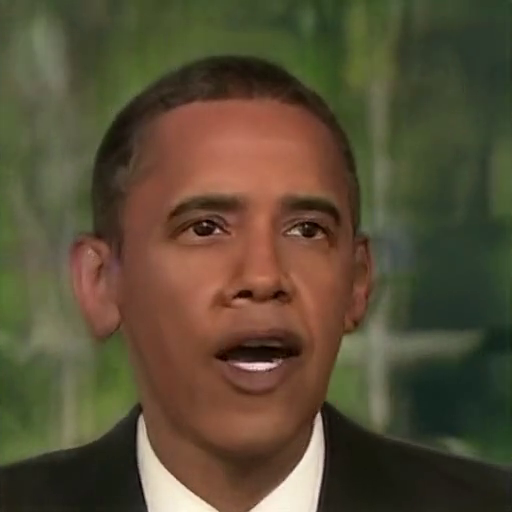}\hfill &
\includegraphics[width=\linewidth]{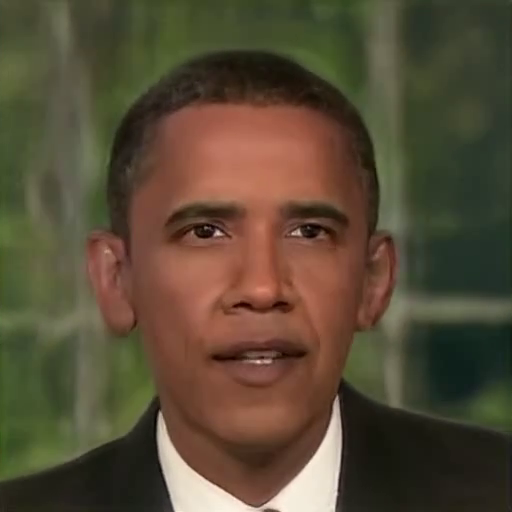}\hfill &
\includegraphics[width=\linewidth]{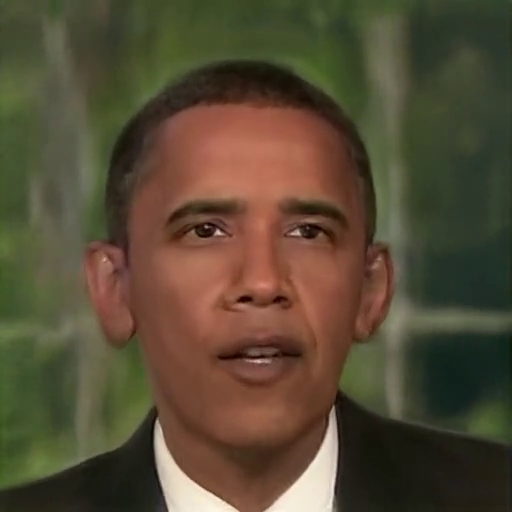}\hfill &
\includegraphics[width=\linewidth]{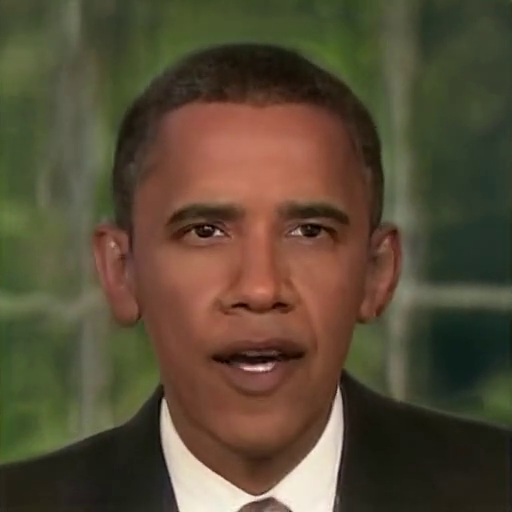}\hfill &
\includegraphics[width=\linewidth]{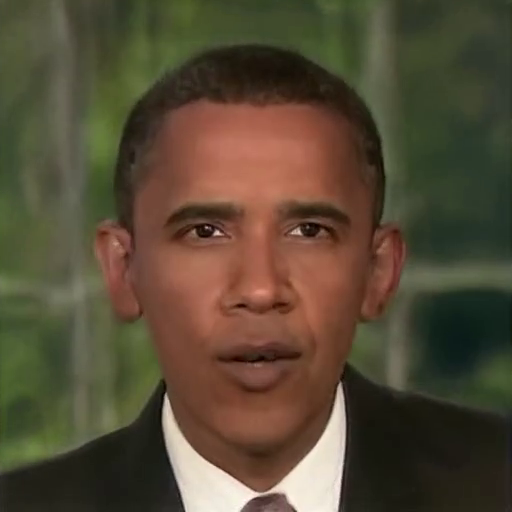}\hfill \\

w/o attention &
\includegraphics[width=\linewidth]{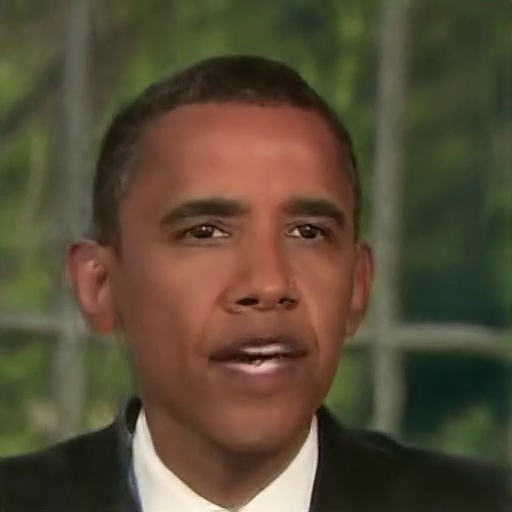}\hfill &
\includegraphics[width=\linewidth]{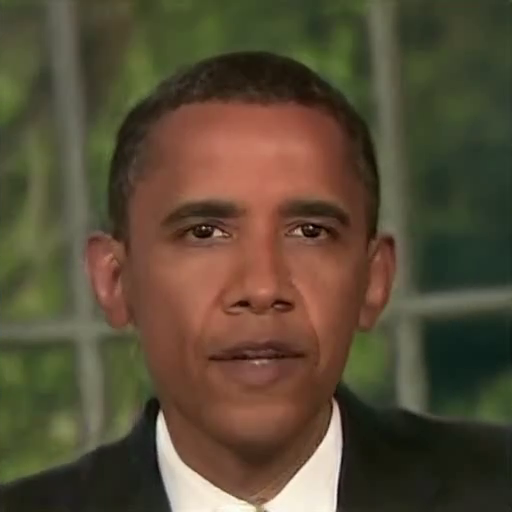}\hfill &
\includegraphics[width=\linewidth]{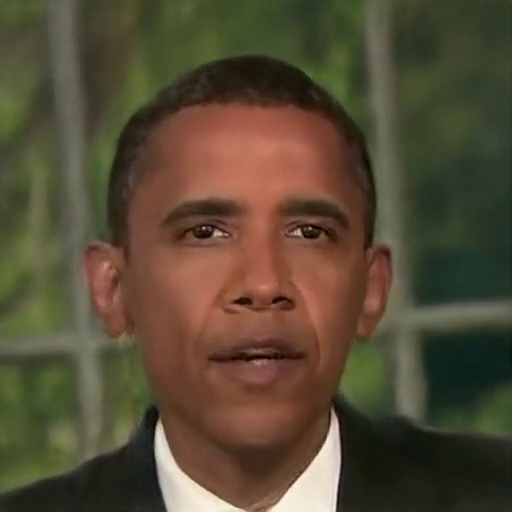}\hfill &
\includegraphics[width=\linewidth]{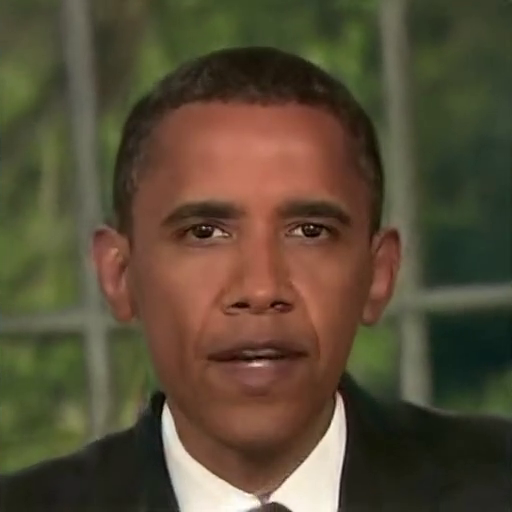}\hfill &
\includegraphics[width=\linewidth]{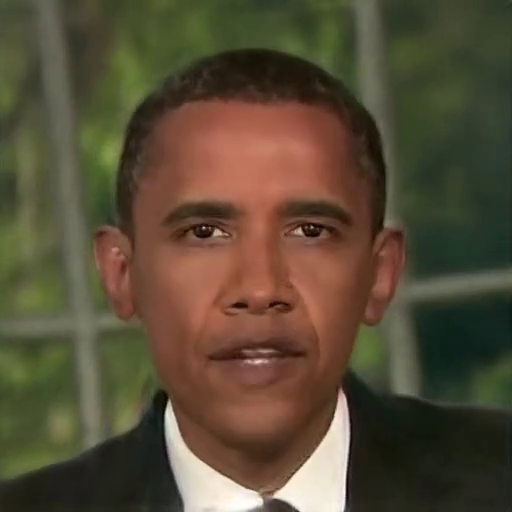}\hfill \\

w/o split &
\includegraphics[width=\linewidth]{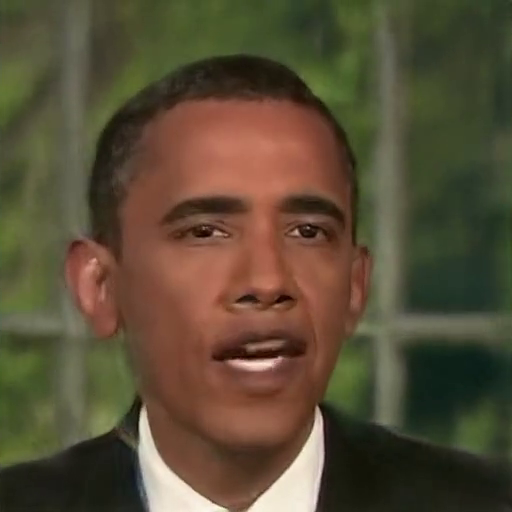}\hfill &
\includegraphics[width=\linewidth]{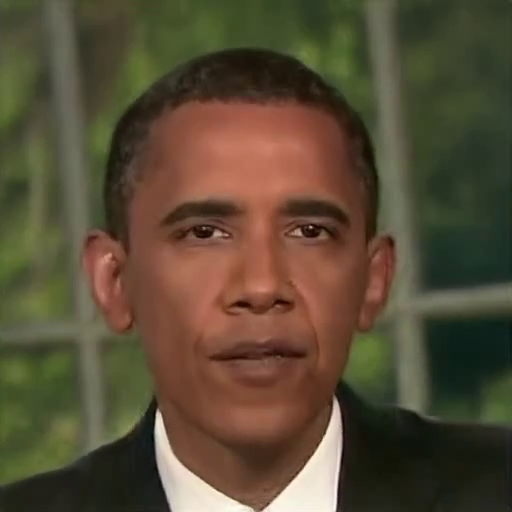}\hfill &
\includegraphics[width=\linewidth]{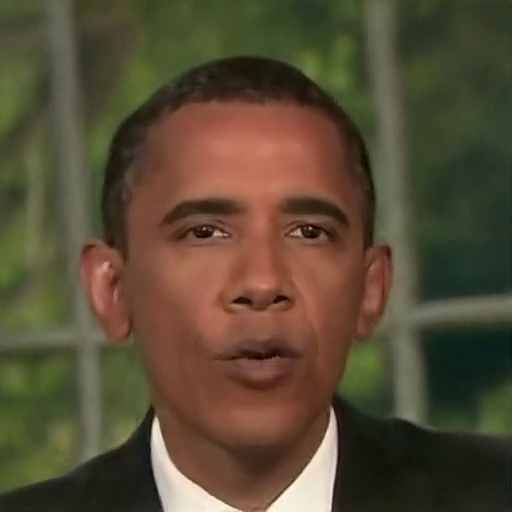}\hfill &
\includegraphics[width=\linewidth]{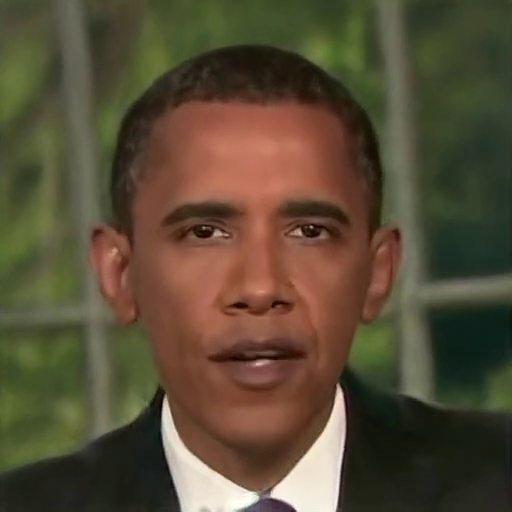}\hfill &
\includegraphics[width=\linewidth]{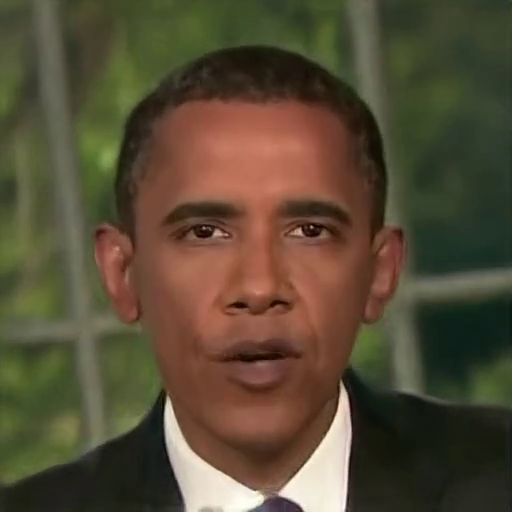}\hfill \\

w/o rollout &
\includegraphics[width=\linewidth]{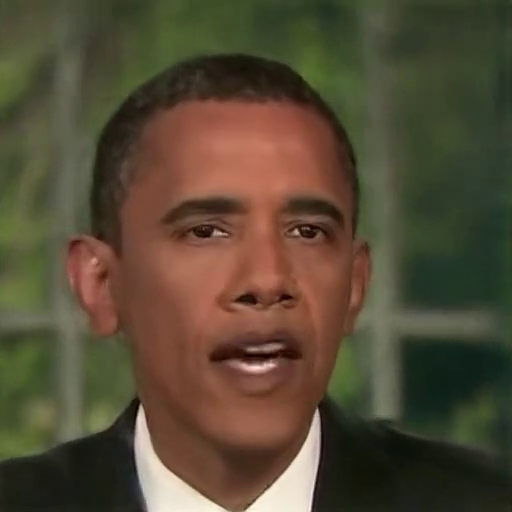}\hfill &
\includegraphics[width=\linewidth]{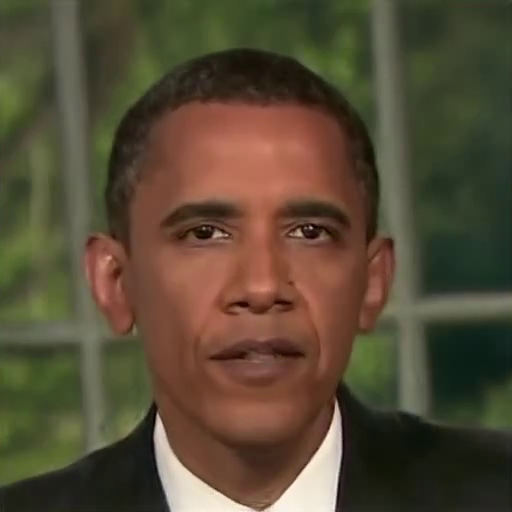}\hfill &
\includegraphics[width=\linewidth]{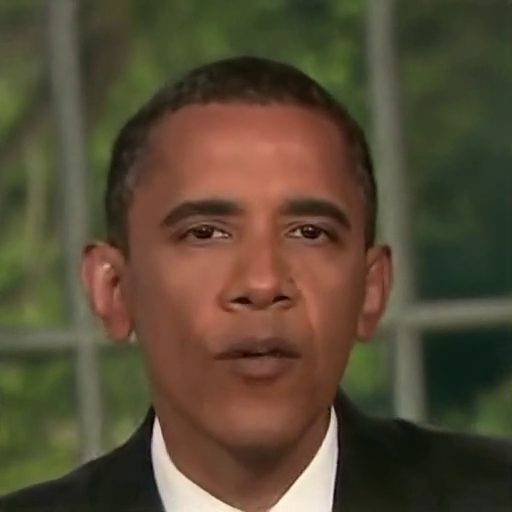}\hfill &
\includegraphics[width=\linewidth]{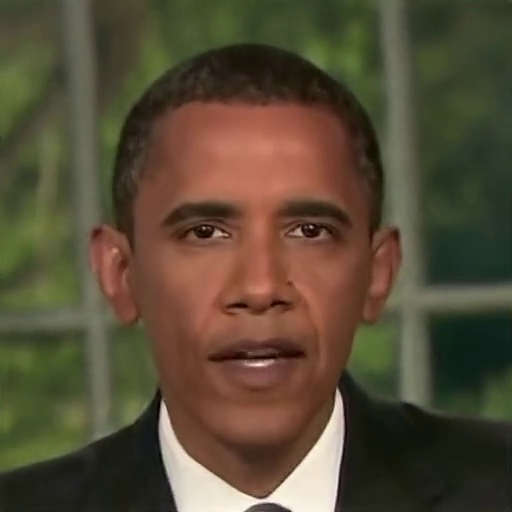}\hfill &
\includegraphics[width=\linewidth]{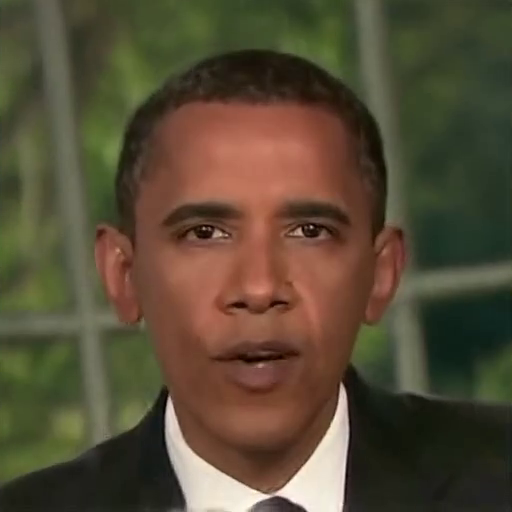}\hfill \\

All (Ours) &
\includegraphics[width=\linewidth]{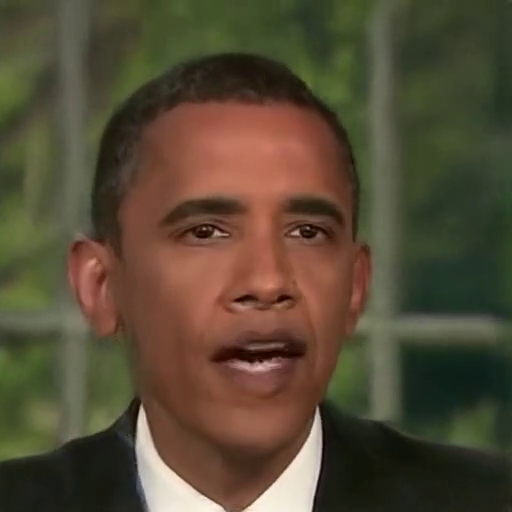}\hfill &
\includegraphics[width=\linewidth]{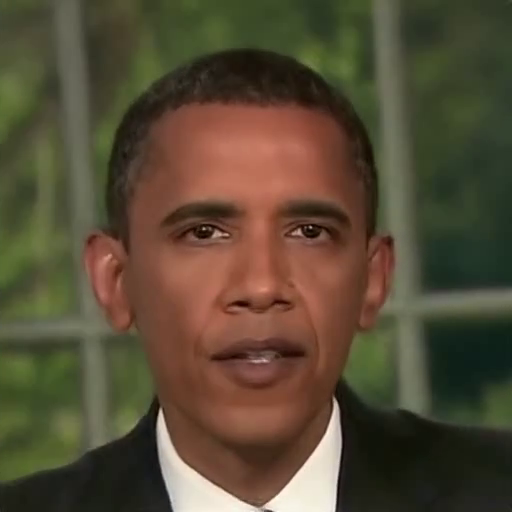}\hfill &
\includegraphics[width=\linewidth]{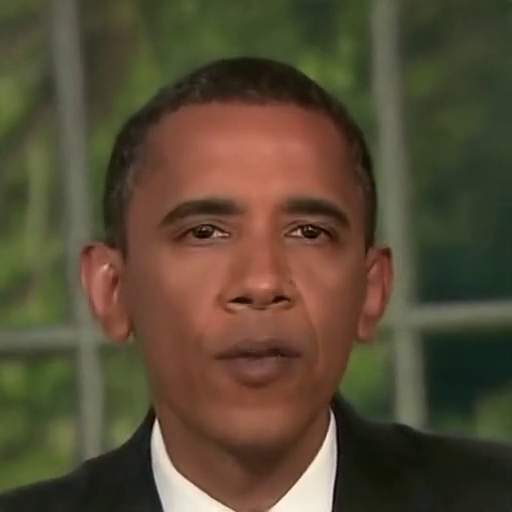}\hfill &
\includegraphics[width=\linewidth]{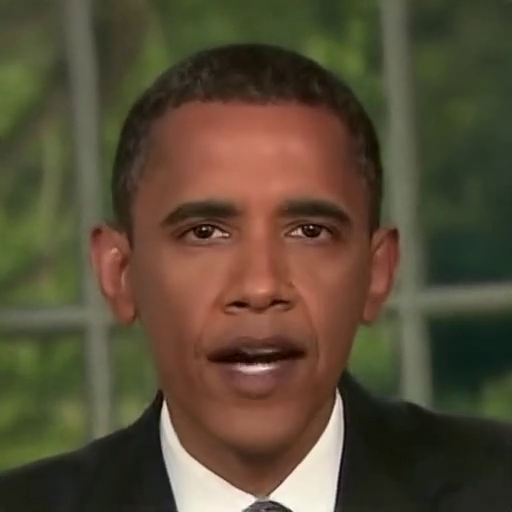}\hfill &
\includegraphics[width=\linewidth]{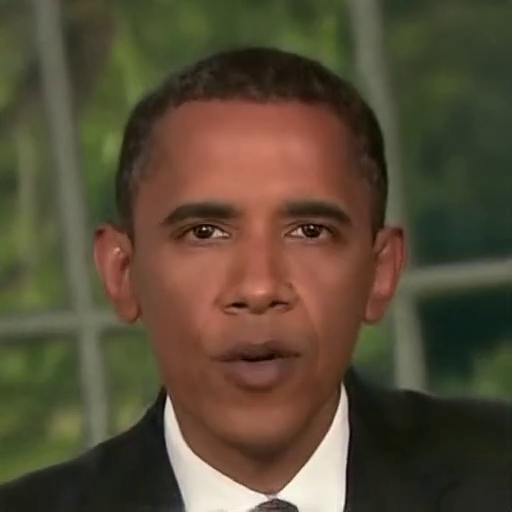}\hfill \\

\end{tabular}}
\vspace{-5pt}
\caption{\textbf{Ablation study on specific design selections of deltaplane predictor.} We perform an ablation study on the deltaplane predictor design choice to demonstrate the impact of our method on modeling accurate facial reconstruction. }
\label{supfig:ablation_deltaplane}
\vspace{-13pt}
\end{figure*}

\figref{supfig:ablation_deltaplane} showcases a qualitative evaluation of our deltaplane predictor's design choices discussed in the main paper. This evaluation highlights the critical role of each design element in optimizing both image quality and lip-synchronization accuracy. Our results establish the necessity of all design aspects in achieving these performance gains.

\subsubsection{Effect of the personalized generator. }
To demonstrate the effectiveness of incorporating the personalized generator, 
we compare our full model with an ablation setting that does not utilize the generator fine-tuning method of VIVE3D~\cite{fruhstuck2023vive3d}. The quantitative comparison in \tabref{tab:personalize_abl} illustrates the image quality enhancement achieved through our personalization method, as reflected in the higher metrics of our full model.
In \figref{abl:personalize}, we observed image quality degradation from the ablation result, especially for the noisy pixels around the nose or eye region.

\begin{table}[!t]
\centering

\caption{\textbf{Ablation study on the use of personalized generator.}}
\vspace{-5pt}
\newcommand{\RomanNumeralCaps}[1]{\MakeUppercase{\romannumeral #1}}
\newcolumntype{M}[1]{>{\centering\arraybackslash}m{#1}}
\setlength{\tabcolsep}{3.7mm}
\resizebox{1.\linewidth}{!}{
\begin{tabular}{ m{0.20\linewidth}  M{0.10\linewidth} M{0.10\linewidth} M{0.10\linewidth} M{0.10\linewidth} M{0.10\linewidth} M{0.10\linewidth} M{0.10\linewidth} }  
\toprule
Method  & PSNR $\uparrow$ & LPIPS $\downarrow$ & FID $\downarrow$ & Sync $\uparrow$  & LMD $\downarrow$ & AUE $\downarrow$   & IDSIM $\uparrow$  \\ \midrule
    Ground Truth  & -              & -            & -              & $8.605$  &  0  & 0 & 1  \\\arrayrulecolor{black!50}\specialrule{0.1ex}{0.2ex}{0.2ex}
w/o $\mathcal{G}_\mathrm{ID}$ & 26.031    &0.060 & 17.860 & 6.498 & 3.237 & 1.802 & 0.879  \\

w/ $\mathcal{G}_\mathrm{ID}$  & $26.799$  & $0.054$  & $8.627$  & $6.529$ &$3.227$  & $1.540$ &  $0.917$ \\\arrayrulecolor{black!100} \bottomrule
\end{tabular}
}

\label{tab:personalize_abl}
\end{table}

\begin{figure}[!htp]
\newcommand{\rom}[1]{\uppercase\expandafter{\romannumeral #1\relax}}
\newcommand{\RomanNumeralCaps}[1]{\MakeUppercase{\romannumeral #1}}
\newcolumntype{M}[1]{>{\centering\arraybackslash}m{#1}}
\setlength{\tabcolsep}{0.5pt}
\renewcommand{\arraystretch}{0.25}
\centering
\scriptsize

\resizebox{\linewidth}{!}{
\begin{tabular}{ M{0.1\linewidth} M{0.18\linewidth}  M{0.18\linewidth} M{0.18\linewidth}M{0.18\linewidth} M{0.18\linewidth}  }

Ground Truth &
\includegraphics[width=\linewidth]{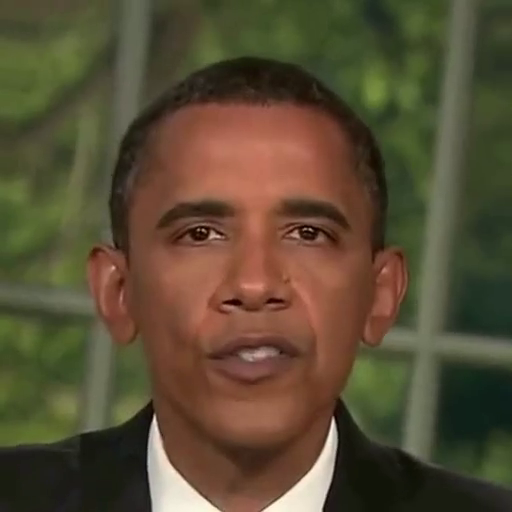}\hfill &
\includegraphics[width=\linewidth]{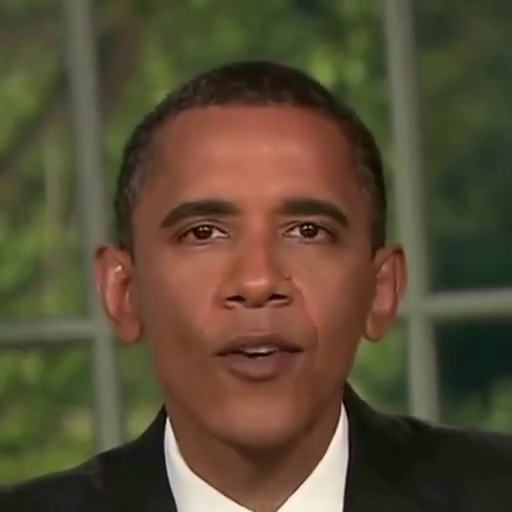}\hfill &
\includegraphics[width=\linewidth]{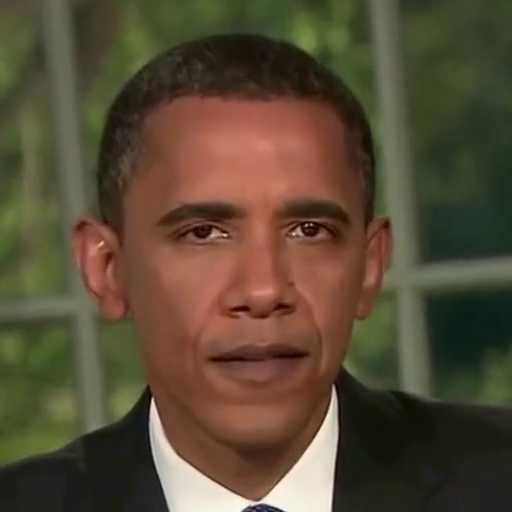}\hfill &
\includegraphics[width=\linewidth]{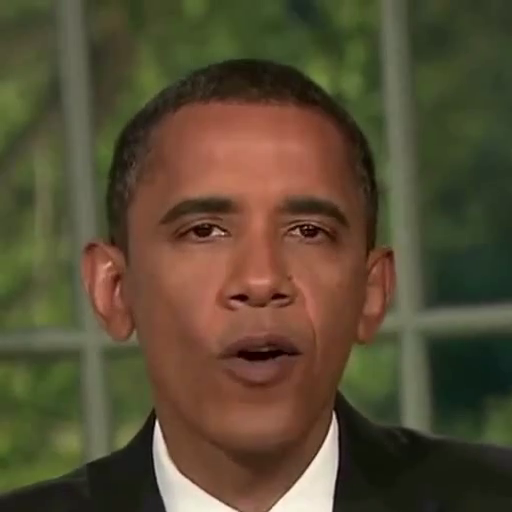}\hfill &
\includegraphics[width=\linewidth]{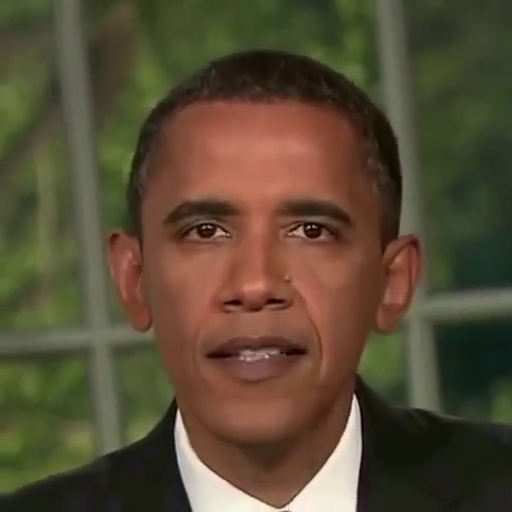}\hfill \\

w/o $\mathcal{G}_\mathrm{ID}$ &
\includegraphics[width=\linewidth]{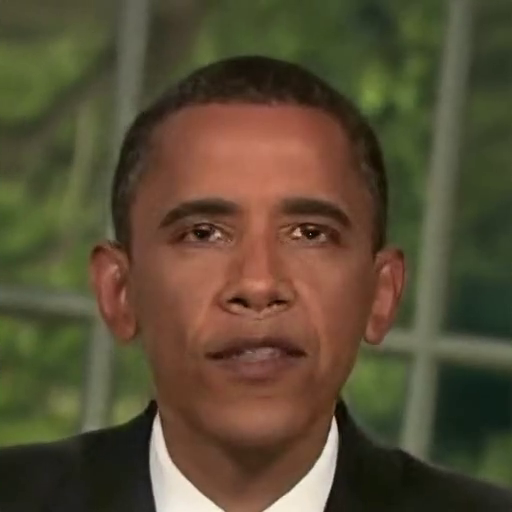}\hfill &
\includegraphics[width=\linewidth]{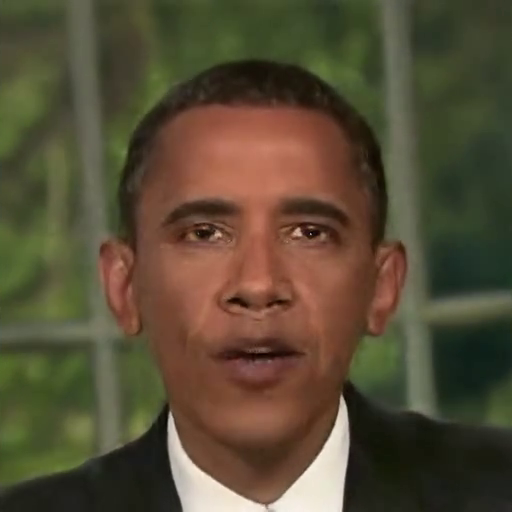}\hfill &
\includegraphics[width=\linewidth]{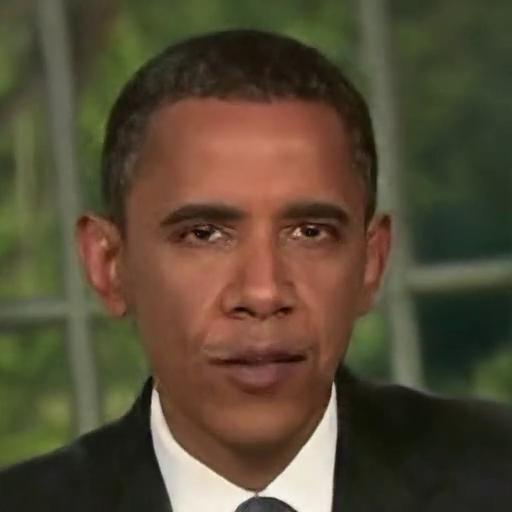}\hfill &
\includegraphics[width=\linewidth]{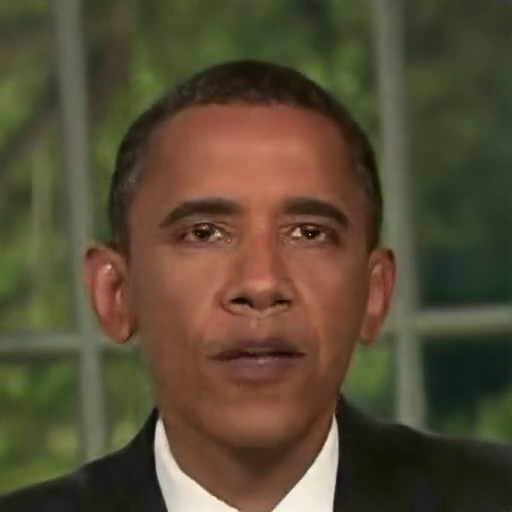}\hfill &
\includegraphics[width=\linewidth]{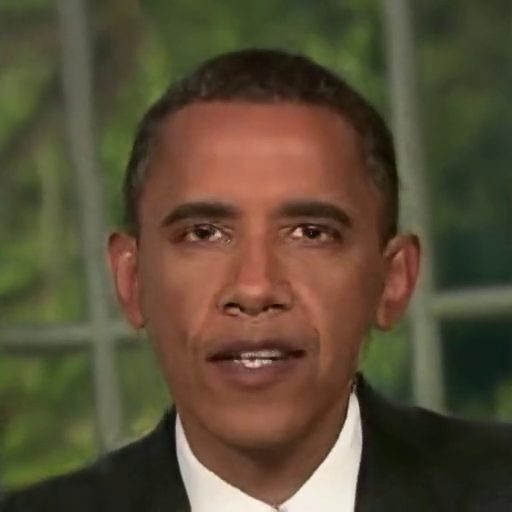}\hfill \\

w/ $\mathcal{G}_\mathrm{ID}$ &
\includegraphics[width=\linewidth]{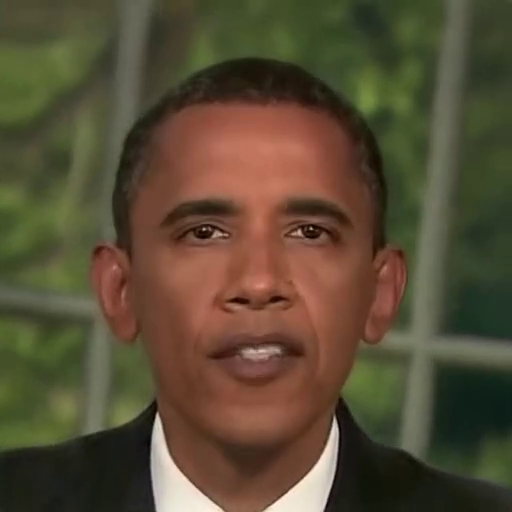}\hfill &
\includegraphics[width=\linewidth]{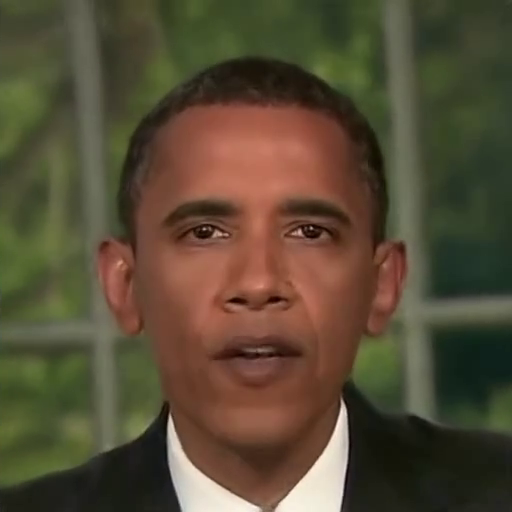}\hfill &
\includegraphics[width=\linewidth]{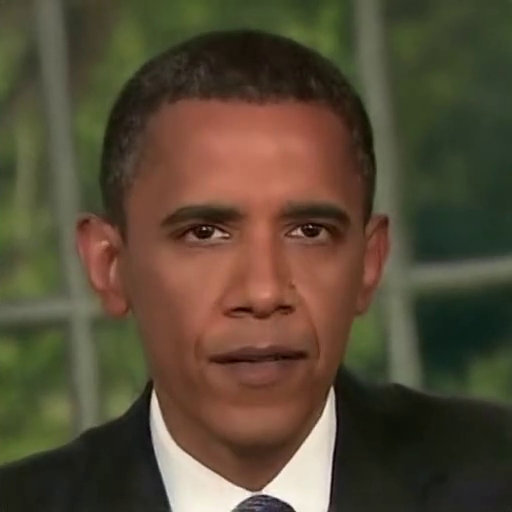}\hfill &
\includegraphics[width=\linewidth]{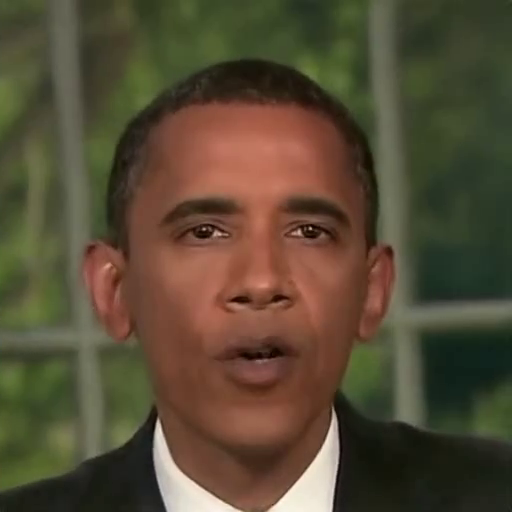}\hfill &
\includegraphics[width=\linewidth]{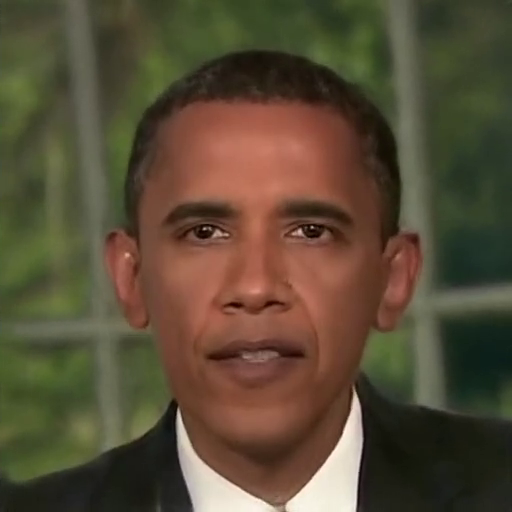}\hfill \\

\end{tabular}} \vspace{-5pt}
\caption{\textbf{Visualization of the effect of personalized generator.}}
\label{abl:personalize}
\end{figure}

\section{Facial Editing}
\label{supsec:editing}
In this section, we introduce an additional feature of our model that distinguishes it from other NeRF-based methods: facial attribute manipulation. 
Talk3D is built on pre-trained EG3D~\cite{chan2022efficient} and thus inherits the rich and diverse latent space of the generative models. The latent space of EG3D enables semantic editing by adding pre-defined style vectors to the input latent code. We exploit InterFaceGAN~\cite{shen2020interfacegan, shen2020interfacegan2} to find several style vectors $\mathbf{w}_\mathrm{edit}$ which represent the semantic editing directions within the EG3D latent space. 
However, naively applying InterFaceGAN to our methodology is not feasible, since our approach directly predicts the triplane representation instead of a latent code. So we slightly alter the methodology of InterFaceGAN by simply replacing the identity triplane with the edited triplane $\mathbf{P}_\mathrm{edit}$. Specifically, for given personalized generator $\mathcal{G}_\mathrm{ID}$ and identity latent code $\mathbf{w}_\mathrm{ID}$, we first construct the edited triplane $\mathbf{P}_\mathrm{edit}$ as:
\begin{equation}
    \mathbf{P}_\mathrm{edit} = \mathcal{G}_\mathrm{ID}(\mathbf{w}_\mathrm{ID}+\mathbf{w}_\mathrm{edit};\theta^*_\mathcal{G}).
\end{equation}
Then we replace the identity triplane to generate edited image $I^\mathrm{edit}_n$ as:
\begin{equation}
    I^\mathrm{edit}_n = \mathcal{R}(\mathbf{P}_\mathrm{edit} + \Delta \mathbf{P}_n, \pi_n;\theta^*_\mathcal{R}).
\end{equation}

\begin{figure*}[!p]
\newcommand{\rom}[1]{\uppercase\expandafter{\romannumeral #1\relax}}
\newcommand{\RomanNumeralCaps}[1]{\MakeUppercase{\romannumeral #1}}
\newcolumntype{M}[1]{>{\centering\arraybackslash}m{#1}}
\setlength{\tabcolsep}{0.5pt}
\renewcommand{\arraystretch}{0.25}
\centering
\scriptsize

\resizebox{\linewidth}{!}{
\begin{tabular}{ M{0.1\linewidth} M{0.18\linewidth}  M{0.18\linewidth} M{0.18\linewidth}M{0.18\linewidth} M{0.18\linewidth}  }

Original &
\includegraphics[width=\linewidth]{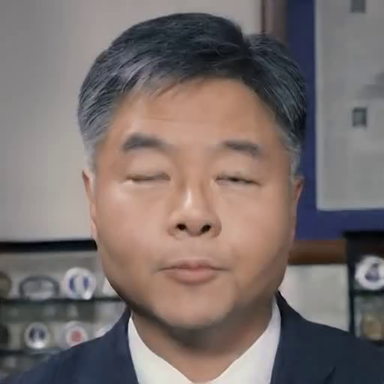}\hfill &
\includegraphics[width=\linewidth]{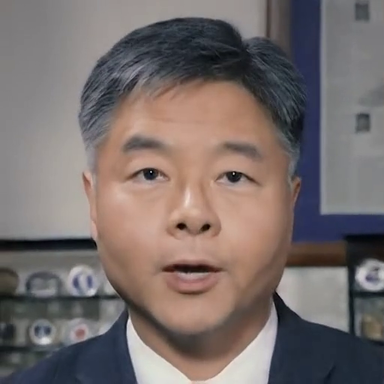}\hfill &
\includegraphics[width=\linewidth]{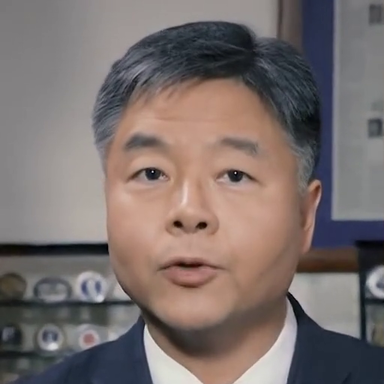}\hfill &
\includegraphics[width=\linewidth]{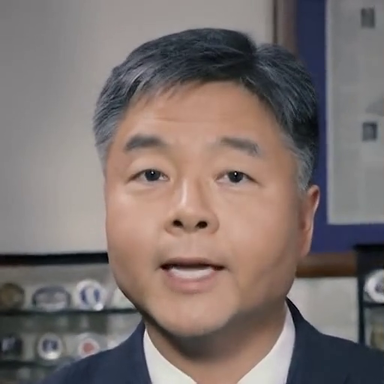}\hfill &
\includegraphics[width=\linewidth]{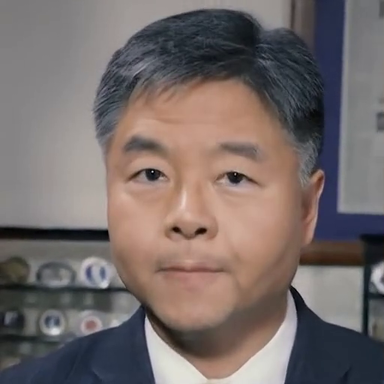}\hfill \\

+ Age &
\includegraphics[width=\linewidth]{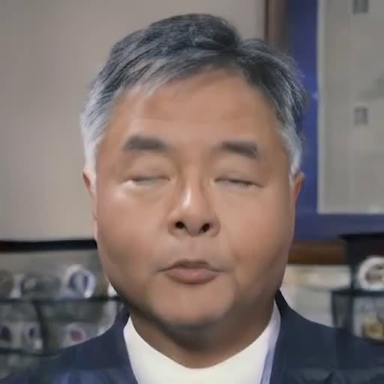}\hfill &
\includegraphics[width=\linewidth]{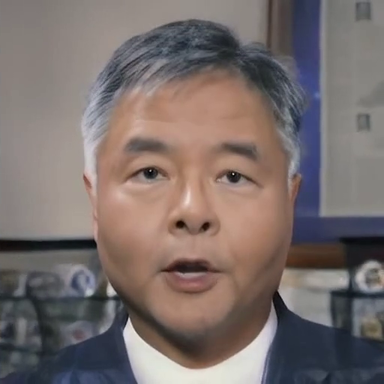}\hfill &
\includegraphics[width=\linewidth]{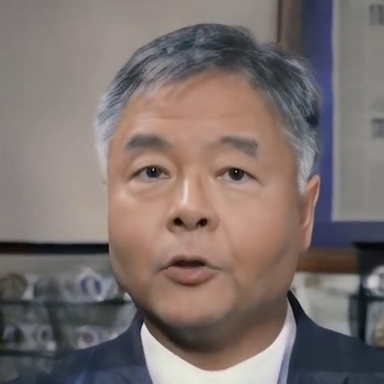}\hfill &
\includegraphics[width=\linewidth]{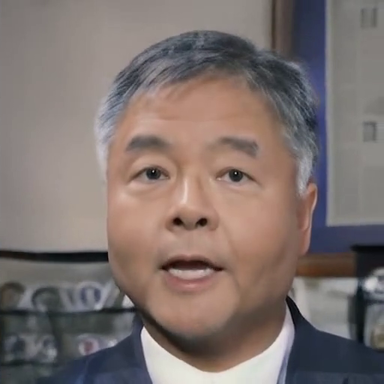}\hfill &
\includegraphics[width=\linewidth]{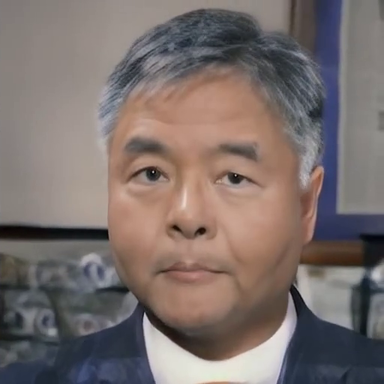}\hfill \\ \\

Original &
\includegraphics[width=\linewidth]{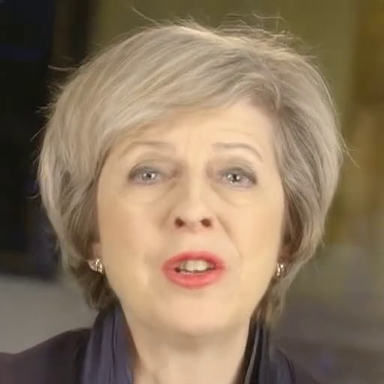}\hfill &
\includegraphics[width=\linewidth]{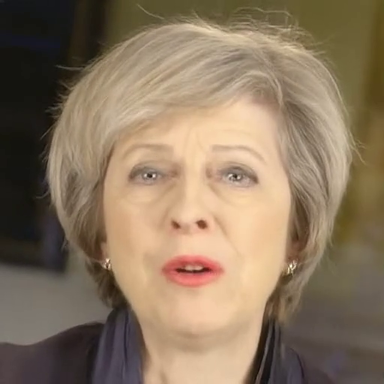}\hfill &
\includegraphics[width=\linewidth]{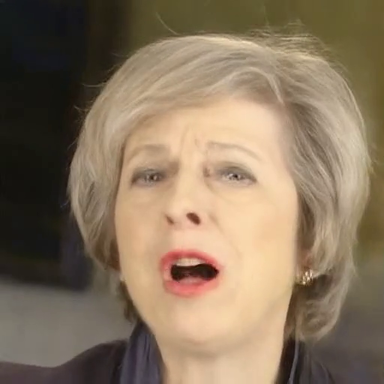}\hfill &
\includegraphics[width=\linewidth]{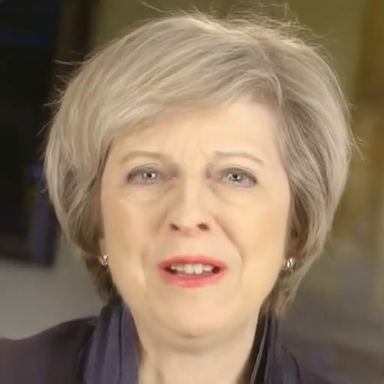}\hfill &
\includegraphics[width=\linewidth]{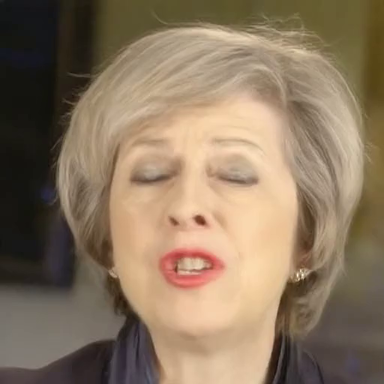}\hfill \\

+ Age &
\includegraphics[width=\linewidth]{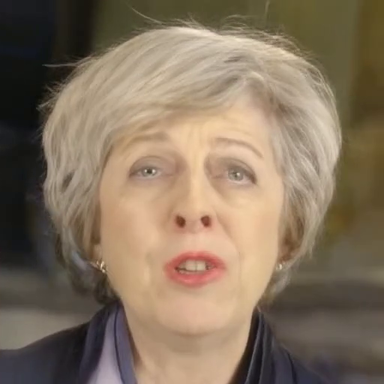}\hfill &
\includegraphics[width=\linewidth]{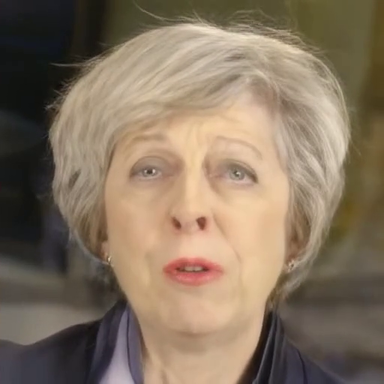}\hfill &
\includegraphics[width=\linewidth]{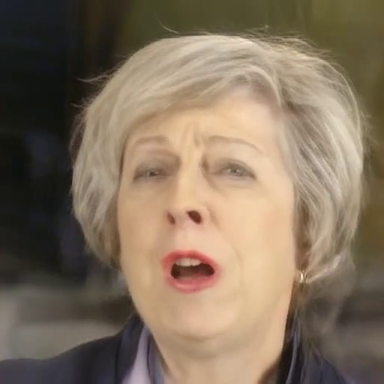}\hfill &
\includegraphics[width=\linewidth]{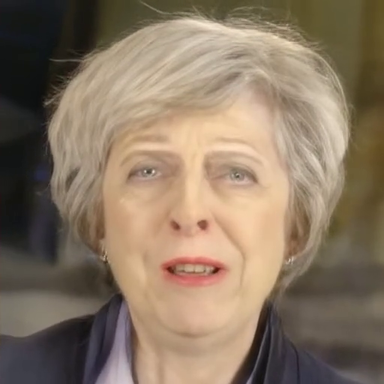}\hfill &
\includegraphics[width=\linewidth]{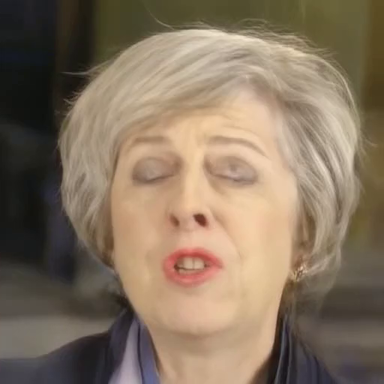}\hfill \\ \\

Original &
\includegraphics[width=\linewidth]{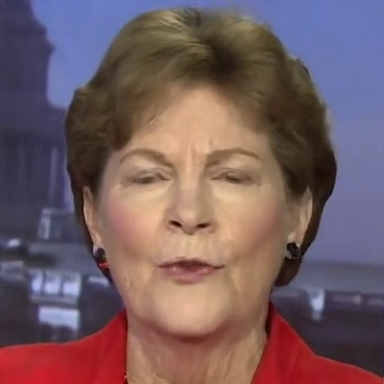}\hfill &
\includegraphics[width=\linewidth]{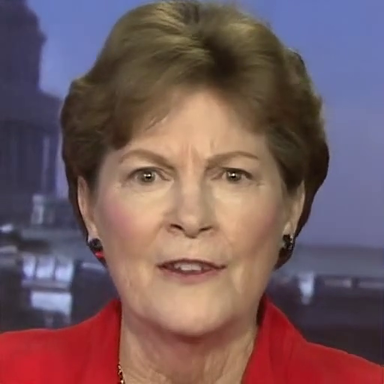}\hfill &
\includegraphics[width=\linewidth]{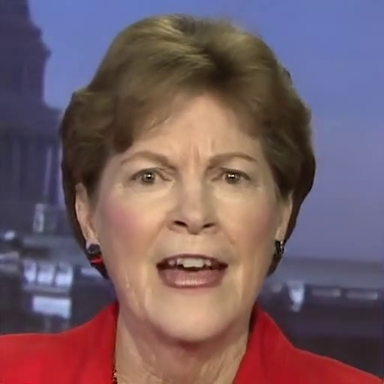}\hfill &
\includegraphics[width=\linewidth]{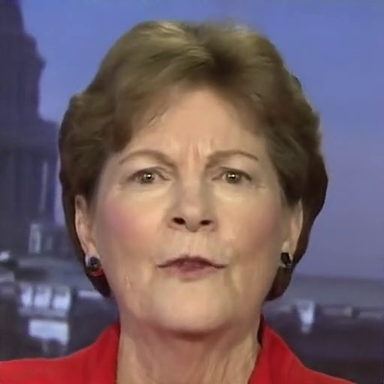}\hfill &
\includegraphics[width=\linewidth]{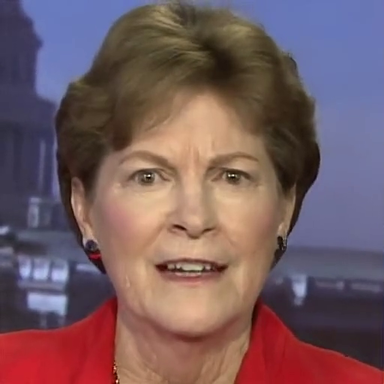}\hfill \\

+ Long Hair &
\includegraphics[width=\linewidth]{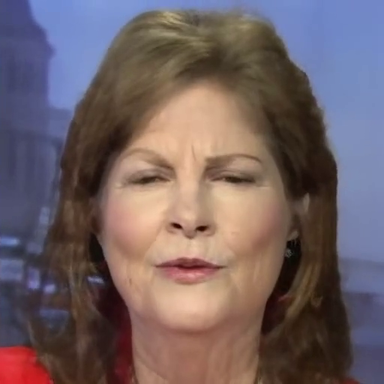}\hfill &
\includegraphics[width=\linewidth]{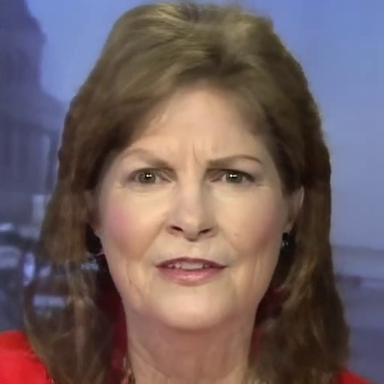}\hfill &
\includegraphics[width=\linewidth]{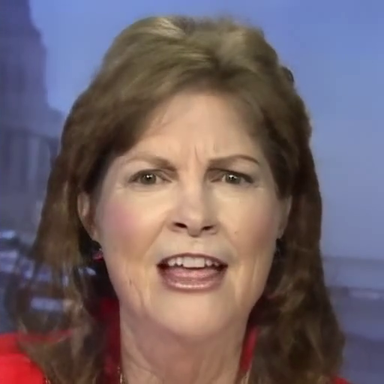}\hfill &
\includegraphics[width=\linewidth]{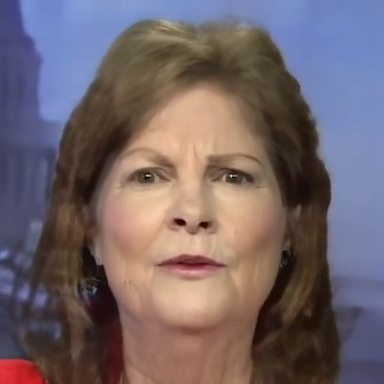}\hfill &
\includegraphics[width=\linewidth]{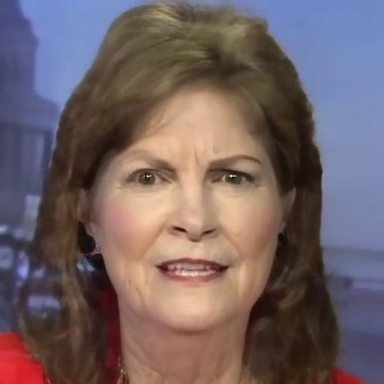}\hfill \\

\end{tabular}}
\vspace{-5pt}
\caption{\textbf{Facial attribute manipulation results.} }
\label{supfig:facial_edit}
\vspace{-13pt}
\end{figure*}

In \figref{supfig:facial_edit}, we visualize the results of editing several attributes, including age and hair length.  The process demonstrates consistent manipulation across attributes like age and hair length, without disrupting lip synchronization.

\section{Supplementary Video}
\label{supsec:video}
To comprehensively visualize the efficacy of our proposed method in the domain of talking facial video synthesis, we have compiled a supplementary video that encapsulates each result shown in the main paper and the supplementary materials. 
This video showcases not only the facial animations generated by our method but also includes detailed comparisons with other relevant techniques. The video also features diverse outcomes, highlighting our approach's versatility across different languages and robustness of rendering performance at extreme viewpoints.
Additionally, the video incorporates an ablation study that dissects the contributions of individual components within our methodology.

\section{Broader Impact}
\label{supsec:broader_impact}
\subsection{Ethical considerations}

In developing Talk3D, we hope to advance applications in digital humans, video production, and human-computer interaction assistance by generating highly realistic talking portraits with accurate lip-audio synchronization. However, we recognize the ethical considerations surrounding the misuse of such technology for malicious purposes. The photorealistic nature of the generated portrait videos makes it challenging for individuals to distinguish between authentic and synthetic content. To address this concern, we emphasize the importance of informing users about the authenticity of videos and recognize the ongoing challenges in discriminating synthesized high-fidelity portraits from recent NeRF-based methods. To contribute to the responsible development and deployment of talking portrait synthesis, we commit to sharing our generated results with deepfake detection communities and supporting the enhancement of detection mechanisms. Additionally, we advocate for protective measures, such as incorporating digital watermarks in real portrait speech videos, to mitigate potential misuse. Furthermore, we highlight the need to consider regulatory frameworks that govern the use of deepfake techniques to prevent unintended negative consequences when synthetic content is shared on social media platforms. Both policymakers and the public must be informed about the potential risks associated with deepfakes, fostering a cautious and responsible approach to their creation and utilization.

\subsection{Limitations and future work}
Compared to earlier NeRF-based works, our Talk3D excels in high-fidelity talking portrait synthesis, especially when rendered from the extreme viewpoint, thanks to the rich generative prior of EG3D~\cite{chan2022efficient}. However, we also inherit its shortcomings, as our method does not generalize well outside of photorealistic images of human faces, compared to other talking face synthesis works such as MakeItTalk~\cite{zhou2020makelttalk} and SadTalker~\cite{zhang2023sadtalker}, which can handle cartoon characters or stylized caricatures. Also, our model's current reliance on GAN inversion introduces technical complexities that impact data preparation. Specifically, it demands precise alignment and cropping of video frames, extending preprocessing time. Additionally, incomplete coverage of essential facial regions can lead to visual artifacts in the form of blurriness or distortion. Future development will focus on overcoming these limitations to increase our method's adaptability and ensure consistent performance across a wider range of training data.
\bibliographystyle{splncs04}
\bibliography{egbib}
\end{document}